\def\mode{0}

\if 0\mode
	\documentclass{article}
	\usepackage[utf8]{inputenc}
	\usepackage[margin=1in]{geometry}

	\usepackage{acronym}
\usepackage{graphicx}
\usepackage{mathtools, amscd, mathrsfs}
\usepackage{multirow}
\usepackage{url}
\usepackage{tikz}
\usepackage{tikz-cd}

\usetikzlibrary{shapes.geometric}
\usepackage{amsmath}
\usepackage{amsfonts}
\usepackage{amsthm}
\usepackage{multirow}
\usepackage{colortbl}
\usepackage{bm}
\usepackage{array}

\def\eqref#1{equation~\ref{#1}}

\def\1{\bm{1}}

\def\eps{{\epsilon}}

\def\veta{{\bm{\eta}}}

\def\vtheta{{\bm{\theta}}}

\def\vb{{\bm{b}}}

\def\vg{{\bm{g}}}

\def\vw{{\bm{w}}}
\def\vx{{\bm{x}}}
\def\vy{{\bm{y}}}
\def\vz{{\bm{z}}}

\def\mW{{\bm{W}}}

\DeclareMathAlphabet{\mathsfit}{\encodingdefault}{\sfdefault}{m}{sl}
\SetMathAlphabet{\mathsfit}{bold}{\encodingdefault}{\sfdefault}{bx}{n}

\def\sB{{\mathbb{B}}}

\def\sR{{\mathbb{R}}}
\def\sS{{\mathbb{S}}}

\def\sZ{{\mathbb{Z}}}

\newcommand{\Ls}{\mathcal{L}}
\newcommand{\R}{\mathbb{R}}

\DeclareMathOperator*{\argmax}{arg\,max}
\DeclareMathOperator*{\argmin}{arg\,min}

\DeclareMathOperator{\sign}{sign}

\newcommand{\specialcell}[2][c]{%
  \begin{tabular}[#1]{@{}c@{}}#2\end{tabular}}

\newcommand{\etal}{et al.}
\newcommand{\etc}{etc.}
\newcommand{\ie}{i.e.}
\newcommand{\eg}{e.g.}
\newcommand{\wrt}{w.r.t}
\newcommand{\eqq}{eq.}
\newcommand{\stt}{s.t.}

\if 0\mode

\else
	
\fi

\newcommand{\newpara}
    {
    \vskip 0.2in
    }

\allowdisplaybreaks

\if 0\mode
    \usepackage{newclude}
    \usepackage{dirtree}
    \usepackage{booktabs}
    \usetikzlibrary{backgrounds}

    \usepackage{fixltx2e}
    \usepackage{caption}
    \usepackage{subfig}
    \usepackage{lscape}
    \usepackage{longtable}

\else
    \usepackage{caption}
    \usepackage{subcaption}
    \usepackage{dirtree}
    \usepackage{fixltx2e}
    \usepackage{lscape}
    \usepackage{afterpage}
\fi

\begin{document}

	\title{Adversarial Examples~-~A Complete Characterisation of the Phenomenon}

	\author{Alexandru Constantin Serban\thanks{a.serban@cs.ru.nl}\\ Radboud University \\ Software Improvement Group \\ The Netherlands \and Erik Poll \\ Radboud University \\ The Netherlands \and Joost Visser\\ Radboud University \\ Software Improvement Group \\ The Netherlands}

    \date{}
    \maketitle

    \begin{abstract}
    \if 0\mode 
We provide a complete characterisation of the phenomenon of adversarial examples~-~inputs intentionally crafted to fool machine learning models.
We aim to cover all the important concerns in this field of study:
(1) the conjectures on the existence of adversarial examples, (2) the security, safety and robustness implications, (3) the methods used to generate and (4) protect against adversarial examples and (5) the ability of adversarial examples to transfer between different machine learning models.
We provide ample background information in an effort to make this document self-contained.
Therefore, this document can be used as survey, tutorial or as a catalog of attacks and defences using adversarial examples.

\else 

Recent publications have demonstrated that neural networks are vulnerable to inputs which are almost indistinguishable from natural data, yet cause misclassifications.
These inputs, called \emph{adversarial examples}, are intentionally crafted by humans in order to fool neural networks.
Besides revealing how fragile neural networks are, they can be exploited by attackers in security contexts.
The discovery of adversarial examples triggered a wide body of, sometimes controversial, publications.
In this paper, we present a survey of the state-of-the-art techniques used to generate or protect against adversarial examples.
We discuss the implications of adversarial examples on safety, security and robustness of neural networks and present a taxonomy able to characterise all attacks and defences using adversarial examples.
Moreover, we provide an in-depth analysis of the hypotheses on the existence of adversarial examples and discuss their property to transfer between different machine learning models.
We conclude shaping future research directions.

\fi
    \end{abstract}

    \acrodef{ML}[ML]{Machine Learning}
\acrodef{DL}[DL]{Deep Learning}
\acrodef{DNN}[DNN]{Deep Neural Networks}
\acrodef{SVM}[SVM]{Support Vector Machine}
\acrodef{RL}[RL]{Reinforcement learning}

\acrodef{wrt}[\emph{w.r.t}]{with respect to}
\acrodef{st}[\emph{s.t.}]{such that}

\acrodef{fgsm}[FGS]{Fast Gradient Sign}
\acrodef{bim}[BI]{Basic Iterative}
\acrodef{illcm}[ILC]{Iterative Least-likely Class}
\acrodef{jsma}[JSMA]{Jacobian-based Saliency Map Attack}
\acrodef{uap}[UAP]{Universal Adversarial Perturbations}
\acrodef{opa}[OPA]{One Pixel Attack}
\acrodef{pgd}[PGD]{projected gradient descent}
\acrodef{rssa}[RSSA]{Randomised Single Step Attack}
\acrodef{eat}[EAT]{Ensemble Adversarial Training}
\acrodef{gaa}[GAA]{Generative Adversarial Attacks}
\acrodef{gan}[GAN]{Generative Adversarial Networks}
\acrodef{nae}[NAE]{Natural Adversarial Examples}
\acrodef{atn}[ATN]{Adversarial Transformation Networks}
\acrodef{vae}[VAE]{Variational Auto-Encoders}
\acrodef{cfoa}[CFOA]{Complete First Order Adversary}
\acrodef{iid}[i.i.d]{independent and identically distributed}
\acrodef{bpda}[BPDA]{Backward Pass Differentiable Approximation}
\acrodef{alp}[ALP]{Adversarial Logit Pairing}
\acrodef{fbgan}[FB-GAN]{Featurized Bidirectional Generative Adversarial Networks}
\acrodef{sap}[SAP]{Stochastic Activation Pruning}
\acrodef{mat}[MAT]{Multi-strength Adversarial Training}
\acrodef{dam}[DAM]{Dense Associative Memory}
\acrodef{zoo}[ZOO]{Zeroth Order optimisation}
\acrodef{sa}[STA]{Strong Adversary}
\acrodef{lm}[LM]{Linear Models}
\acrodef{dt}[DT]{Decision Trees}
\acrodef{knn}[KNN]{K-nearest Neighbour}

    \section{Introduction}
\label{sec:intro}

There is no doubt that \ac{ML} and, in particular, \ac{DL} algorithms achieve impressive results on tasks where it is impossible to specify a procedural rule-set.
Some examples are object recognition~\cite{simonyan2014very,szegedy2015going, he2015delving, he2016deep}, machine translation~\cite{chen2018best, sutskever2014sequence} or speech recognition~\cite{xiong2016achieving, zhang2018deep, lee2018accelerating, vaswani2017attention}.
Fuelled by the ease of adoption (through the increase of cheap computational power and development of high level APIs), \ac{DL} algorithms are explored in a variety of new tasks and commercial applications.
One of the most promising application, with a meaningful impact in the transportation industry, is the development of autonomous vehicles~-~systems that heavily rely on \ac{DL} algorithms to perceive the environment and reason about it~\cite{bojarski2016end}.

Facing commercial deployment and the possibility of use in safety-critical systems, new properties of \ac{DL} algorithms become important.
In particular, their ability to maintain performance whenever faced with data coming from slightly different distributions than trained with.
This property is defined as the algorithm's power to \emph{generalise} outside the training dataset or, consistent with the optimisation literature, the algorithm's \emph{robustness}.

In optimisation, a robust solution is one that has the ability to perform well under a certain level of uncertainty~\cite{ben2009robust}.
Recent discoveries in \ac{DL} research~\cite{szegedy2013intriguing} showed that neural networks exhibit low robustness and triggered an impressive wave of (sometimes controversial) publications.
Notably, \ac{DNN} can be highly sensitive to small, \emph{intentional}, distribution drifts~-~inputs which substantially decrease their performance, while being in close resemblance to training data.
The term \emph{adversarial examples} was first used in~\cite{szegedy2013intriguing} to describe such inputs.

Since an \emph{intention} is required, a substantial number of publications claim security consequences, \eg~\cite{nguyen2015deep, kurakin2016aadversarial, guo2017countering, papernot2017practical, su2017one, hein17formal, sinha2018certifying, fawzi2018adversarial}.
The same body of publications hypothesise that commercial deployment is hindered by low robustness.
In contrast, recent publications~\cite{gilmer2018motivating} show these claims are sometimes exaggerated and demand clear security requirements for a correct evaluation.
In between, a considerable number of publications investigate the existence of adversarial examples from a theoretical perspective and try to shed light on this peculiar behaviour of \ac{DNN}.
Overall, there are two important reasons why adversarial examples are important: (1) because an attacker might exploit them and (2) because they suggest \ac{DNN} are not robust.

Another phenomenon revealed by the existence of adversarial examples is the power to use them against very different \ac{ML} techniques.
This means an input designed to confuse \ac{DNN} can trigger the same behaviour for kernel methods~\cite{papernot2016transferability, hearst1998support}.
From a security perspective, this property suggests an attacker does not need any information about the model under attack.
Moreover, from a learning theory perspective, it suggests that (1) algorithms extrapolate similar discriminants, in spite of the difference in method and (2) high sensitivity to similar distribution drifts is an universal phenomenon, independent of method.

\if 0\mode
The goal of this report is to provide a comprehensive and complete overview of this research field.
\else 
The goal of this paper is to provide a comprehensive and complete survey of this research field. Moreover, we present a taxonomy for the attacks and defences against adversarial examples.
\fi
We aim to characterise the phenomenon from its inception by discussing its causes, position it in the security context with relevant threat models, introduce methods to construct and defend against adversarial examples and explore the property of adversarial examples to transfer between \ac{ML} models.
Moreover, we discuss various definitions of robustness from the literature and present \if 0\mode a large body of distilled knowledge that helps defining\else\fi future research.
We consider related others surveys and uncover several weaknesses, as presented in Section~\ref{subsec:related_work}.
We aim to fill these gaps, but also to provide enough \if 0\mode background\else\fi information so this document becomes a self-contained introduction to this field.
\if 0\mode
Therefore, this report can be used in tutorials or as a catalog of attacks and defences against adversarial examples.
\else
\fi

Although adversarial examples can be found for a variety of tasks, we restrict our presentation to the task of object recognition because (1) most publications target this task and (2) examples from this field are easier to grasp and understand.
The rest of the paper is organised as follows. 
In Section~\ref{sec:background} we present related work and \if 0\mode sufficient background information to comprehend the phenomenon\else give a formal definition of adversarial examples\fi.
In Section~\ref{sec:attack_models} we discuss the security implications of adversarial examples.
Section~\ref{sec:robustness_eval} discusses safety and robustness implications, followed by a discussion about the existence of adversarial example in Section~\ref{sec:causes}.
In Sections~\ref{sec:attacks} and~\ref{sec:defences} we present and characterise the methods used to generate and protect against adversarial examples.
A discussion about the phenomenon of transferability follows in Section~\ref{sec:transferability}.
\if 0\mode We conclude with a body of distilled knowledge in Section~\ref{sec:distilled_knowledge}.
\else 
We conclude with a short discussion, a summary of our contributions and future research in Section~\ref{sec:conclusions}.
\fi

    \section{Background}
\label{sec:background}

\if 0\mode
In this section we cover several topics.
At first, we introduce related work presenting surveys or tutorials on adversarial examples.
Secondly, we formally define machine learning, deep learning and the task of object recognition.
Thirdly, we give more information about the underlying assumptions of machine learning models and some properties needed to understand adversarial examples.
Lastly, we formally introduce adversarial examples.

\else 

\fi

\subsection{Related Work}
\label{subsec:related_work}

Although the term \emph{adversarial examples} was first coined around 2014~\cite{szegedy2013intriguing} as a result of an investigation on \ac{DNN}, adversarial machine learning was established before \ac{DNN} made a come-back.
Unfortunately, recent publications concerning \ac{DNN} omit past publications on adversarial machine learning and loose important perspectives in this field.

The first publication concerning adversarial \ac{ML} was published in 2004, when Dalvi \etal~\cite{dalvi2004adversarial}, followed by Lowd and Meek~\cite{lowd2005adversarial}, managed to fool linear classifiers for spam detection adding intentional changes to an e-mail's body~\cite{biggio2017wild}.
Barreno \etal~\cite{barreno2006can} introduced a taxonomy for attacks and defences in adversarial settings; which was later refined in~\cite{barreno2010security}.
This early taxonomy clearly defines threat models and is comprehensive enough to include adversarial examples~-~a particular technique of adversarial \ac{ML}.
Alas, late literature on adversarial examples often oversees these developments~\cite{gilmer2018adversarial}.

Thereafter, a large body of publications have developed adversarial attacks against \ac{ML} models at both \emph{training} time~\cite{rubinstein2009antidote, biggio2012poisoning, kloft2010online, xiao2015feature}, \emph{test/inference} time~\cite{dalvi2004adversarial, lowd2005adversarial, globerson2006nightmare} or various defence mechanisms able to overcome these threats~\cite{jordaney2017transcend, kolcz2009feature, bruckner2012static}.
In parallel, several publications have proposed methods to \emph{evaluate} the security of \ac{ML} models against adversarial attacks~\cite{biggio2014security, barreno2010security}.
Biggio and Roli~\cite{biggio2017wild} discuss a parallel between the evolution of adversarial \ac{ML} and the advent of \ac{DL}.

Within this framework, adversarial examples represent attacks against machine learning models at \emph{test} (\emph{inference}) time~-~also called \emph{evasion} attacks.
However, as opposed to early literature on this topic, adversarial examples exhibit a specific trait: the perturbation applied in order to fool a classifier is desired to be minimal, or as small as possible.
Take the case of a spam detector.
Early literature on adversarial machine learning can go as far as modifying all characters in the adversarial e-mail, while the literature on adversarial examples would restrict the changes to (maybe) a few characters.
This requirement enforced in the adversarial examples literature often leads to misconceptions regarding their security implications.
Because adversarial examples can not be differentiated from normal inputs by human observers, a consistent body of publications claim security consequences.
However, as we will see in Section~\ref{sec:attack_models}, a machine can not distinguish between inputs with small or large perturbations, therefore their impact on security is equivalent.

The terms \emph{adversarial classification}~\cite{dalvi2004adversarial}, \emph{evasion attack}~\cite{biggio2013evasion} or \emph{adversarial examples}~\cite{szegedy2013intriguing} have, therefore, equivalent meaning.
In this \if 0\mode report \else paper \fi we are concerned with recent literature, triggered by~\cite{szegedy2013intriguing} and the widely adopted definition of adversarial examples.
This body of work focuses, in particular, on \ac{DNN} and was triggered by the surprisingly small perturbations needed to fool such algorithms.
However, this does not mean we disregard prior work on adversarial classification.
On the contrary, we try to bind the two by discussing both security, safety and robustness implications of adversarial examples.

\if 0\mode \bigskip \else  \newpara  \fi
Two recent publications attempted to survey the field of adversarial examples.
At first, Liu \etal~\cite{liu2018survey} investigate security threats at both training and testing time.
Their work, together with~\cite{biggio2017wild} represents a bridge between the two positions mentioned earlier.
However, because it considers both training and testing time, it fails to cover important details concerning adversarial examples.
At first, the list of attacks and defences is neither complete or exhaustive and fails to illustrate the full range of attack techniques.
It does not explicitly cover attacks based on geometric transformations, generative models or any black-box attacks.
Secondly, the survey fails to provide a clear intuition about state-of-the-art attacks~-~such as the Carlini attacks or the complete first order adversary attack (Section~\ref{subsec:optimisation_attacks})~-~and state-of-the-art defences.
Thirdly, the authors do not mention the real intriguing property revealed by adversarial examples~-~that \ac{DNN} exhibit very low \emph{robustness}.
If this property is exploited in security, safety or other context is independent of its existence.
Nevertheless, the work is important because it maps the phenomenon of adversarial examples to the initial taxonomy of adversarial attacks~\cite{barreno2010security}.

Akhtar and Mian~\cite{akhtar2018threat} focus only on adversarial examples (although the title of their work can be misleading).
The authors present an improved list of attacks and defences.
However, as opposed to Liu \etal~\cite{liu2018survey}, they omit to characterise adversarial threats from a security standpoint.
Although the list of attacks and defences features more entries, it is far from complete.
Once again, attacks based on first order adversary or provable and certified defences are completely missing.
The authors succeed, however, to illustrate in detail the hypothesis on the existence of adversarial examples and the  use of adversarial examples in real-life scenarios.

We aim to complement these two surveys by discussing the different positions regarding the property of robustness, introduce an exhaustive list of attacks and defences and discuss in details the property of transferability.
Moreover, we provide a taxonomy of attacks and defences that is specific to the adversarial examples field.
This taxonomy is meant to properly structure the methods in this field of study and help future researchers to position their work in comparison to other publications.

\if 0\mode

    \subsection{Formal Definition of Machine and Deep Learning}
\label{subsec:machine_learning}

\ac{ML} algorithms are computer programs conceived to learn from data. 
Instead of manually specifying decision rules about a scenario, the algorithms automatically learn them by examining descriptive data.
Mitchell defines a computer program to learn from \emph{experience} $E$ \wrt~some class of \emph{tasks} $T$ and \emph{performance measure} $P$, if its performance at tasks in $T$ as measured by $P$ \emph{improves} with experience $E$~\cite{mitchell1997machine, goodfellow2016deep}.
Following this definition, the three ingredients of \ac{ML} algorithms are: (1) a task, (2) data about the task and (3) a measure of performance or improvement on the task.

As mentioned in the introduction, we will focus on the task of object detection (or classification) because (1) the majority of papers targeting adversarial examples focus on this task and (2) it is easier to provide visual examples.
Formally, a classification algorithm identifies a mapping from an input space (experiences) to an output space that represents a class (label) the input belongs to:

\begin{equation}
	 f : \sR^{n} \rightarrow \{1,2, \dots, k \}. 	
\end{equation}
\noindent
For object recognition, the input is a vector containing the image pixels -~$\vx$~- and the output is a number corresponding to a class (label)~-~$\hat{y}$.
If we group the parameters of the classification function in a vector~-~$\vtheta$~-~we can define a machine learning model as $\hat{y} = f(\vtheta, \vx)$.
Instead of manually specifying the parameters~-~$\vtheta$~-~the goal of \ac{ML} is to automatically learn them by looking at multiple examples.

In order to evaluate the performance of a \ac{ML} model we need to define a \emph{cost function} which computes the \emph{error rate} \ie~the proportion of examples for which the model produces incorrect labels.
If we know the true label of an input before designing the model, the task is called \emph{supervised} learning.
In this case we can define a cost function that takes the correct label and the one returned by our model and outputs an error score:
\begin{equation}
	J(\vtheta) = g(f(\vx, \vtheta), y).
\end{equation}
\noindent
Some examples of functions $g$ are mean square error or cross entropy.

We say that, in order to build a machine learning algorithm, one needs to design an algorithm that will automatically adjust its parameters when seeing new data, in order to increase its performance for a task.
The range of possible parameters describe a hypothesis space; from which a \ac{ML} algorithm selects the best hypothesis for the given task.
The \ac{ML} approach to this problem is to minimise the cost function, $J(\vtheta)$, \wrt~parameters $\vtheta$, on each sample in the dataset, through a process called \emph{training}. 
In order to evaluate the performance of an algorithm in a neutral manner, it is common to separate the available data into a \emph{training dataset} - used only during the training phase - and \emph{testing dataset} - a smaller portion of the data that is only used for testing.

The choice for a cost function may seem straightforward and objective, but in practice it is difficult to choose a function that corresponds well to the desired behaviour of a system.
In some cases, this is a consequence of the fact that it is difficult to decide what should be measured~\cite{goodfellow2016deep}.
Moreover, the dependence of $J$ on $f(\vx, \vtheta)$ impacts both the algorithm design and the optimisation strategy. 
It is well known that non-convex functions can have multiple local optimum points and deciding if a local optimum is a global one or wether the problem has any solutions is computationally expensive. 

%
%

\bigskip
In complex settings, such as object recognition, the function $f(\vx, \vtheta)$ is highly non-linear and hard to engineer.
Moreover, features relevant to the task are hard to extract due to high dimensionality, computational costs or lack of expert knowledge.
A solution to this problem is to compose $f(\vx, \vtheta)$ from several functions and use raw, unprocessed data as input.

\ac{DNN} are \ac{ML} models loosely inspired by neuroscience, in which a model is built through composition of linear and non-linear functions.
The model is analogous to a directed acyclic graph describing how the functions are composed~\cite{goodfellow2016deep}. 
For example, a model can be composed of three functions $f^{(1)}, f^{(2)}, f^{(3)}$, connected in a chain to form $f(\vx) = f^{(3)}(f^{(2)}(f^{(1)}(\vx)))$. 
In this case $f^{(1)}$ is called the \emph{first layer}, $f^{(2)}$ is called the \emph{second layer}, \etc~
Because an algorithm must learn the parameters of these functions, they are called \emph{hidden layers}.
The final layer of a neural network is also called the \emph{output layer}.
Deep Learning is a term attributed to models with many hidden layers (Figure~\ref{fig:fwdprop}).

In practice, each hidden layer applies a linear and a non-linear (activation) transformation to its input. 
The goal of deep learning is to find the parameters of each function in the hidden layers that obtain maximum performance on the training dataset.
The activation function is typically chosen to be a function applied element-wise (\eg~ReLU~\cite{nair2010rectified}).
Formally, we can define the output of one hidden layer as:

\begin{equation}
	h_i = g(\mW_{:,i}^{T} \vx  + \vb_i),
\end{equation}
\noindent
where $g$ is a the non-linear, element-wise, function, $\mW$ is the weight vector for the linear transformation and $\vb_i$ is a vector of bias terms.

The parameters are learned by minimising a cost function (taking a step in the opposite direction of its gradient, at each iteration through the dataset).
Because we have a composition of functions, the gradient is computed through the hidden layers using a chain-rule based algorithm, called \emph{back-propagation}~\cite{chauvin2013backpropagation}.
One of the necessary conditions for back-propagations is to have differentiable activation functions.

\begin{figure}[h]
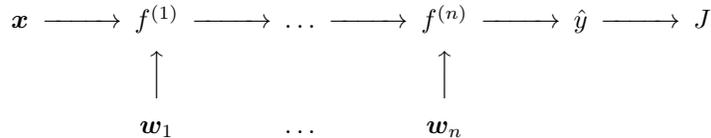

	\begin{equation*}
	  \begin{CD}
    	\vx @>>> f^{(1)} @>>> \dots @>>> f^{(n)} @>>> \hat{y} @>>> J \\
		@. @AAA @. @AAA \\
    	@. \vw_{1} @. \dots @. \vw_{n}
	  \end{CD}
	\end{equation*}	
	\caption{Forward propagation of error (increases the cost).}	
	\label{fig:fwdprop}

\end{figure}

\begin{figure}[h]
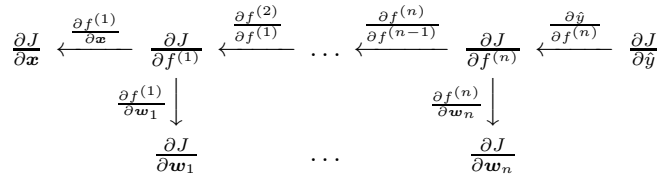

	\begin{equation*}
	  \begin{CD}
    	\frac{\partial J}{\partial \vx} @<\frac{\partial f^{(1)}}{\partial \vx}<< \frac{\partial J}{\partial f^{(1)}} @<\frac{\partial f^{(2)}}{\partial f^{(1)}}<< \dots @<\frac{\partial f^{(n)}}{\partial f^{(n-1)}}<< \frac{\partial J}{\partial f^{(n)}} @<\frac{\partial \hat{y}}{\partial f^{(n)}}<< \frac{\partial J}{\partial \hat{y}} \\    	
		@. @V\frac{\partial f^{(1)}}{\partial \vw_{1}}VV @. @V\frac{\partial f^{(n)}}{\partial \vw_{n}}VV  \\
		@. \frac{\partial J}{\partial \vw_{1}} @.\dots @. \frac{\partial J}{\partial \vw_{n}}
	  \end{CD}
	\end{equation*}
	\caption{Backward propagation of error (decreases the cost).}	
	\label{fig:backprop}
\end{figure}

One may think of a neural network as increasing the error function from left to right through forward-propagation, as showcased in Figure~\ref{fig:fwdprop} and decreasing it from right to left through back-propagation, as showcased in Figure~\ref{fig:backprop}.
By learning the right parameters for the hidden layers, \ac{DNN} allow to trace discriminants between classes in high dimensional spaces (hyper-spaces), where the discriminants are often non-linear.

    \subsection{The Manifold Assumption}
\label{subsec:manifold}

An important concept underlying many ideas in \ac{ML} is that of a manifold~\cite{goodfellow2016deep}.
Intuitively, a manifold is an n-dimensional surface.
Accurately, a manifold is a set of connected points associated with their neighbourhoods and transformations that allow transitions from one point to another.
In \ac{ML}, the term manifold is used to describe a set of connected points that can be well approximated considering only a small numbers of dimensions.
Moreover, in \ac{ML}, the dimensionality of the manifold can vary from one point to another~-~when the manifold intersects itself.

Because approximations across all of $\sR^n$ are impossible, \ac{ML} algorithms assume that most of $\sR^n$ consists of invalid inputs and that interesting inputs lie in a collection of manifolds.
Moreover, it is assumed that variations of the learned function occur only in the directions that lie on the manifold or when moving from one manifold to another.
For example, one can imagine in $\sR^n$ several manifolds describing 10 classes, corresponding to digits from $0-9$.
By moving on a manifold which describes one class, we can identify variations of the same input (such as rotations, translations, \etc).
When moving across manifolds, however, the input can denote a change of class.
This behaviour is called \emph{the manifold assumption}.
Learning the structure of the manifold where the data lies is usually easier because the manifold is described by less dimensions than $\sR^n$.

Although the concept of manifold is mostly used in un-supervised and semi-supervised learning, it is of importance when talking about adversarial examples.

    \subsection{Independent and Identically Distributed Random Variables}
\label{subsec:iid}

\ac{ML} algorithms assume training and test datasets are drawn from the same probability distribution.
Moreover, the examples in each dataset are assumed to be \emph{independent}, \ie~they convey no information about each other and, as a consequence, knowing any information about one does not change the probability distribution of the others.
These assumptions are called the \ac{iid} assumptions and allow the data generation process to described with a probability distribution over a single sample.
The same distribution can later be used to generate all training and testing examples.

The \ac{iid} assumptions are a fundamental tool to study the relationship between training and testing errors.
A key requirement of \ac{ML} algorithms, that distinguishes them from other optimisation solutions, is the capacity to obtain low error on both training and testing datasets \ie~\emph{generalise} to unseen data, such as the testing set.
Because data in the training set is limited, all \ac{ML} algorithms can only approximate the true data generating distribution, \ie~they can only \emph{estimate} it.
Whenever this estimate or the true distribution shifts,  the accuracy of \ac{ML} models drops.

A popular example of distribution shift is \emph{covariate shift} - a change in the distribution of the independent variables that should not impact the output of a \ac{ML} model.
For example, a change in the brightness of all images in the test set should not cause any misclassifications.
Covariate shift is a consequence of a change in the state of latent variables from the distribution.
Adversarial examples have been interpreted in some contexts as an instantiation of covariate shift~\cite{song2017pixeldefend}.
However, common techniques to alleviate the impact of covariate shift in \ac{DNN}~\cite{ioffe2015batch} do not help with adversarial examples.

    \subsection{Norms and Norm Ball}
\label{subsec:norm_balls}

In order to measure the size of a vector or the distance between two vectors, we use a function called a \emph{norm}.
Formally, the $L_p$ norm, $\| \vx \|_p$, is defined as:

\begin{equation}
	\| \vx \|_p = \left(\sum\limits_{i=1}^n \lvert \vx_i \rvert ^p\right)^{1\over p}.
\end{equation}
\noindent
Norms are functions that map vectors to non-negative scalars.
In order to measure the distance between two vectors, we can take the norm of their difference: $\| x' - x \|_p$, which will return a positive scalar.

Three $p$ values are used in the context of adversarial examples, which lead to the following distance measures:

\begin{enumerate}
	\item $L_0$ (based on the absolute norm) - which measures the number of coordinates $i$ such that $\vx_i \ne \vx'_i$. It corresponds to the number of pixels that have been altered in an image.\footnote{In RGB images, there are
			three channels that each can change. We count the number of \emph{pixels} that
			are different, where two pixels are considered different if \emph{any} of the three colours are different.}

	\item $L_2$ (based on the Euclidian norm) - which measures the Euclidean distance between $\vx$ and $\vx'$. The $L_2$ distance can remain small when there are many small changes to many pixels. This distance metric was used in the initial adversarial example work~\cite{szegedy2013intriguing}.

	\item $L_{\infty}$ (based on the maximum norm) - which measures the maximum change to any coordinate:
		\begin{equation*}
			\|\vx'-\vx\|_\infty = \max (|\vx'_1-\vx_1|,\dots,|\vx'_n-\vx_n|).
		\end{equation*}
		For images, one can imagine a maximum limit which bounds the change in each pixel, without restricting the number of pixels that are modified.
\end{enumerate}

\bigskip
In some circumstances, we would like to measure the set of points that lie at a certain norm distance from a chosen point.
This set is called a \emph{norm ball} and is formally defined as:
\begin{equation}
	\sB(\vx_c, r) = \{\vx \mid \| \vx - \vx_c \|_p \leq r\},
\end{equation}
where $\vx_c$ is the point chosen as the centre of the ball and $r$ is the maximum distance from the centre aka radius.

    \subsection{Lipschitz Continuity}
\label{subsec:lipschitz}

In order to limit how fast a function changes, given a change in inputs, one can restrict to functions that are either Lipschitz continuous or have Lipschitz continuous derivatives.
A Lipschitz continuous function is a function whose rate if change is bounded by a Lipschitz constant $\Ls$:
\begin{equation}
	| f(\vx) - f(\vy) | \leq \Ls \| \vx - \vy \|_2.
\end{equation}
\noindent
This property allows us to quantify the assumption that a small change in the input will have a small impact in the output of a function.
Enforcing this constraint is fairly easy, and often used to prove a bound to robustness of \ac{DNN}. 
Szegedy \etal~\cite{szegedy2013intriguing} showed that deep  neural  networks with  half-rectified  layers  (\ie~convolutional  or  fully  connected layers with ReLU activation functions), max pooling and  contrast-normalisation  layers  are  Lipschitz  continuous.
Later it has been proved that the softmax output layer and the sigmoid and hyperbolic tangent activation functions also satisfy Lipschitz continuity~\cite{ruan2018reachability}.

    \subsection{Formal Definition of Adversarial Examples}
\label{subsec:formal_adversarial}

We \if 0\mode are now ready to \else\fi provide a formal definition of adversarial examples, as first stated in the paper that coined the term~\cite{szegedy2013intriguing}.
Given a classification function $f$ and a clean sample $\vx$, which gets correctly classified by $f$ with the label $l$, we construct an adversarial example $\vx'$ by applying the minimal \emph{perturbation} $\veta$ to input $\vx$ such that it gets classified by the model with a different label, $l'$:
\begin{equation}
	\begin{aligned}
	\label{eq:adversarial_generic}
		&\min_{\vx'} & & \|\vx' - \vx\|_p, \\
		&s.t. & & f(\vx') = l', \\
		& & & f(\vx) = l, \\
		& & & l \neq l', \\
		& & & \vx' \in [0,1]^m,				
	\end{aligned}
\end{equation}
\noindent
where $ || \cdot \|_p$ is the distance between the samples (usually the p-norm) and $\vx' - \vx = \veta$ is called a \emph{perturbation}.
Searching the minimal perturbation is "not trivial"~\cite{papernot2017practical} because different properties of \ac{DNN} make the search non-linear and non-convex~\cite{larochelle2009exploring}.
However, many approximation methods have been proposed (Section~\ref{sec:attacks}).
\if 0\mode
The condition that a perturbation is \emph{minimal} does not always hold because, in contrast to humans, a \ac{ML} model can not distinguish between large or small perturbations.
However, it is enforced in most methods used to generate adversarial examples.
\else \fi

We illustrate the adversarial generation process and give some examples of adversarial inputs in Figure~\ref{fig:adv_examples}.
Determining the minimal perturbation is equivalent to moving the data-point corresponding to an input from the region of a correct class to another, as illustrated in Figure~\ref{adv_space}.
If the perturbation is small, the result can not be perceived by humans as malicious, as illustrated in Figure~\ref{adv_exam}.

\if 0\mode
\begin{figure}[h]%
    \centering
    \subfloat[The adversarial generation process~\cite{tramer2017space}.]{\label{adv_space}{\includegraphics[height=4cm, keepaspectratio]{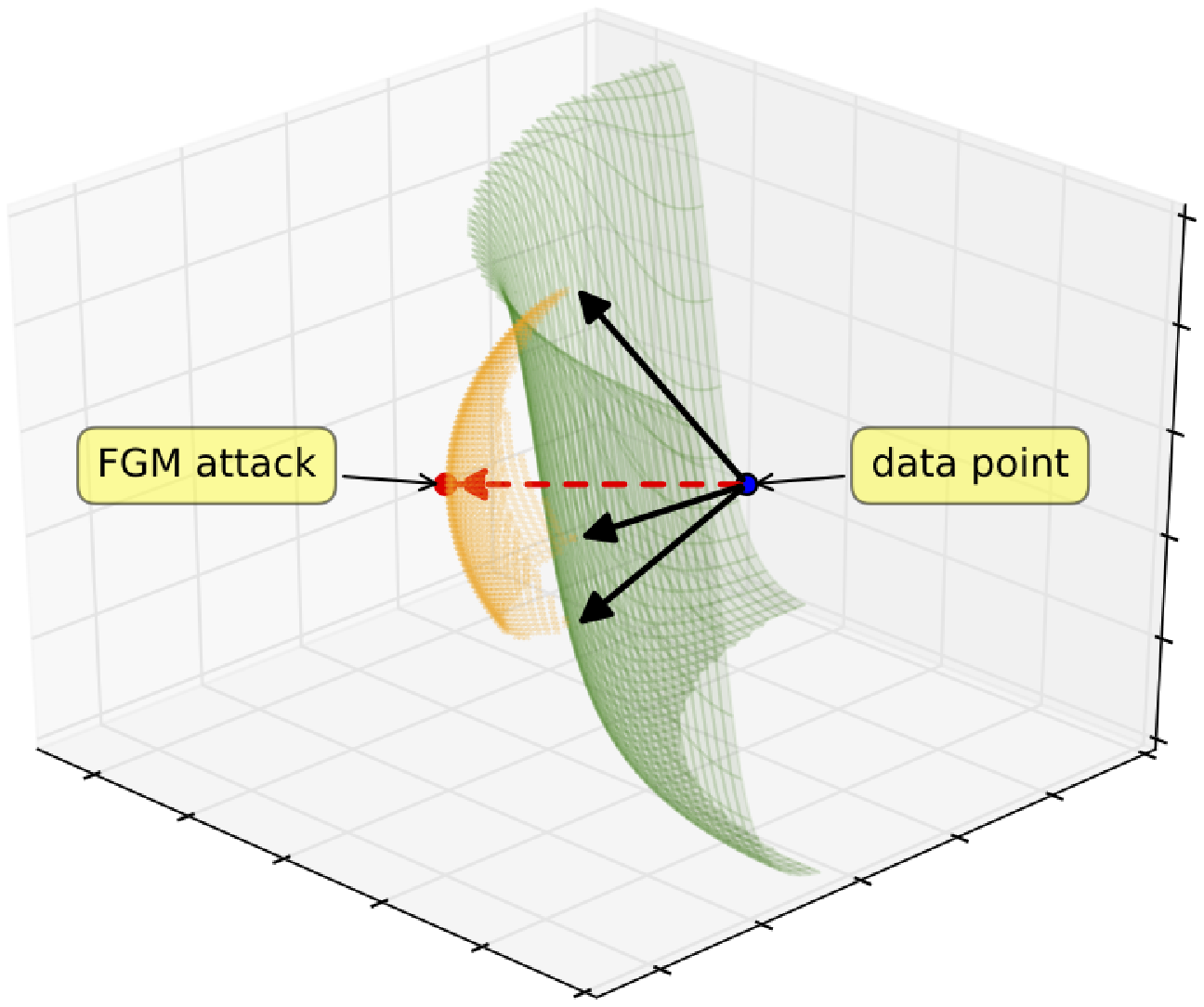} }}%
    \qquad \qquad
    \subfloat[Adversarial inputs~\cite{szegedy2013intriguing}.]{\label{adv_exam}{\includegraphics[height=4cm, width=5cm, keepaspectratio]{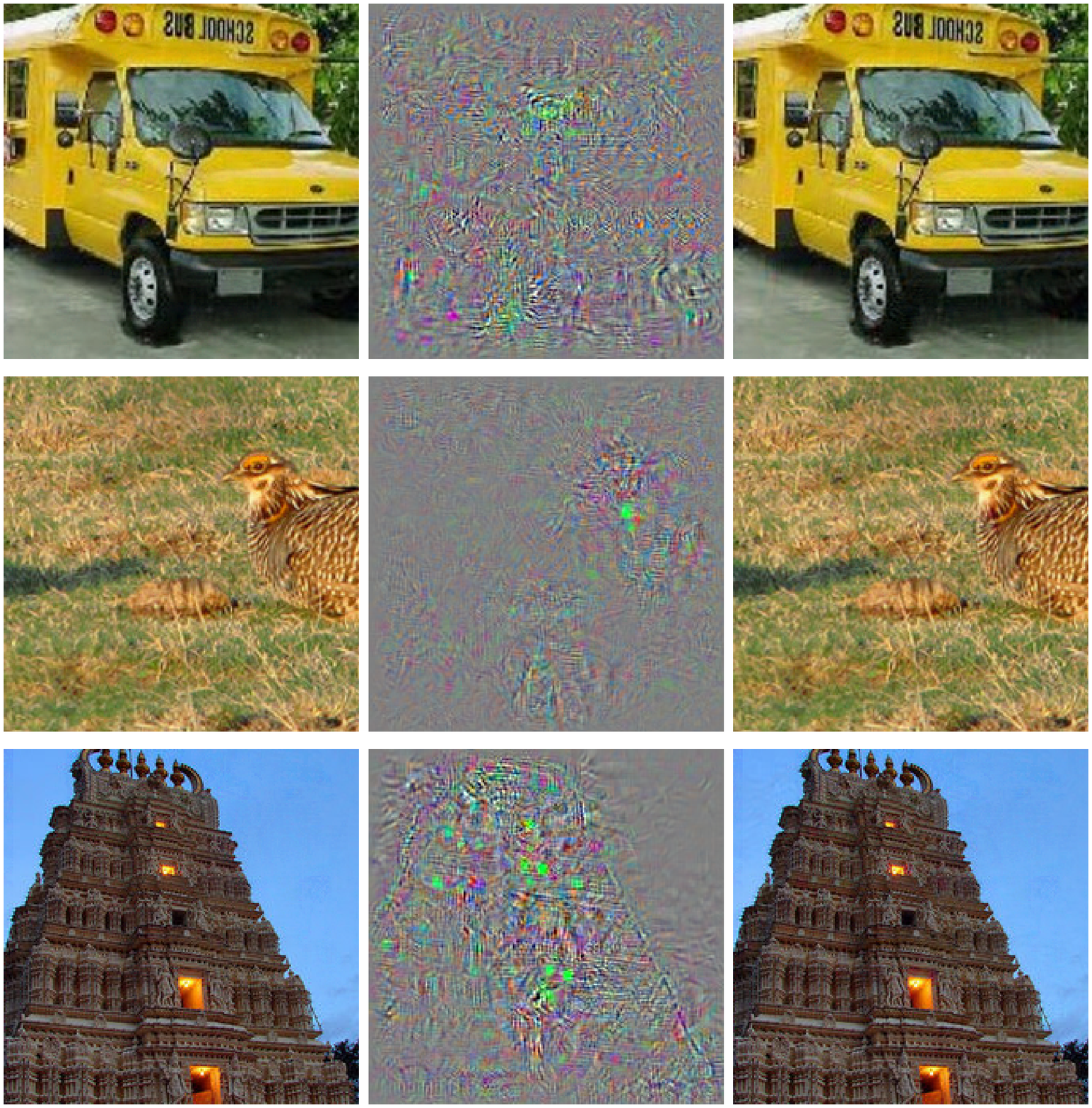} }}%
    \caption{An illustration of crafting an adversarial examples (a)~-~equivalent to moving the data-point corresponding to an input from the region of a correct class to an incorrect class and (b) an illustration of adversarial examples with very low perturbations~-~the pictures in the first column are correctly classified inputs, while the pictures in the last column are adversarial examples, crafted to cause a misclassification. In between, the perturbation used to cause the misclassification is illustrated.}%
    \label{fig:adv_examples}%
\end{figure}
\else 

 \begin{figure}[h]	
 		\centering
		\begin{subfigure}[]{0.49\textwidth}
			\centering
			\includegraphics[height=4cm, keepaspectratio]{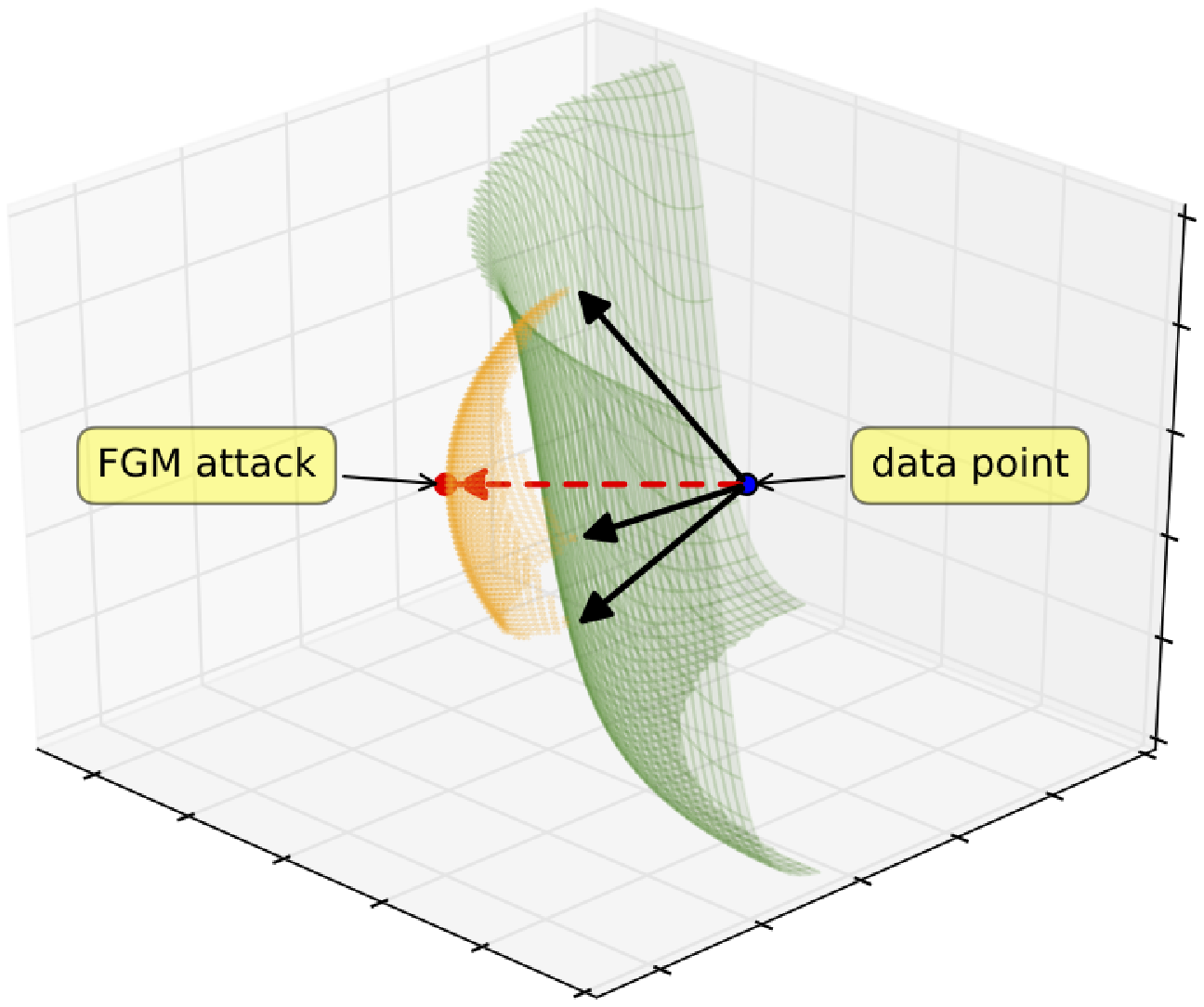}
            \subcaption{The adversarial generation process~\cite{tramer2017space}.}
			\label{adv_space}						
		\end{subfigure}
		\begin{subfigure}{0.49\textwidth}
        	\centering
            \includegraphics[height=4cm, width=5cm, keepaspectratio]{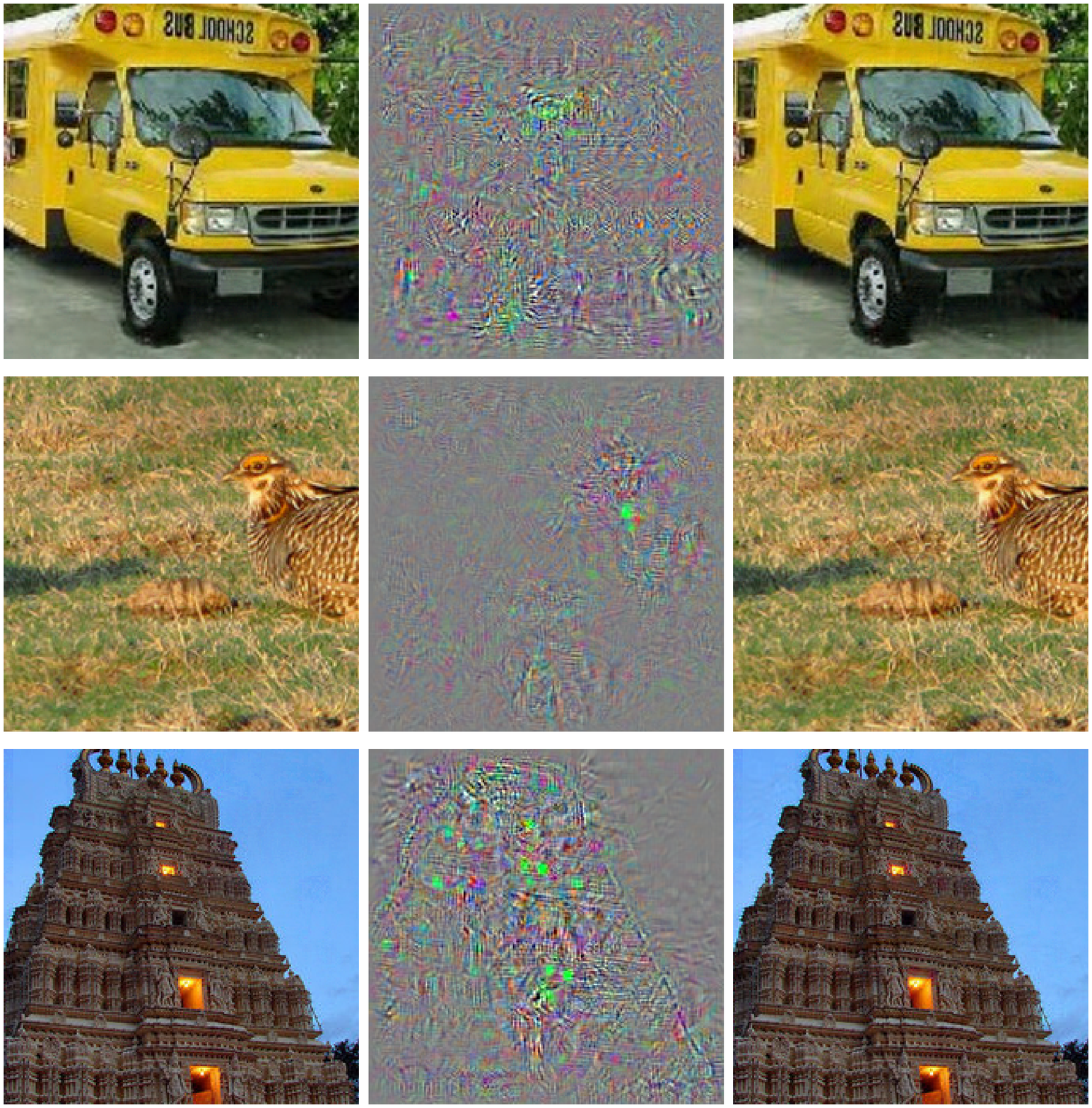}
            \subcaption{Adversarial examples~\cite{szegedy2013intriguing}.}
            \label{adv_exam}			
		\end{subfigure}
	\caption{An illustration of crafting an adversarial examples (a)~-~equivalent to moving the data-point corresponding to an input from the region of a correct class to an incorrect class and (b) an illustration of adversarial examples with very low perturbations~-~the pictures in the first column are correctly classified inputs, while the pictures in the last column are adversarial examples, crafted to cause a misclassification. In between, the perturbation used to cause the misclassification is illustrated.}%
    \label{fig:adv_examples}%
\end{figure}  
\fi

Because the perturbation is not perceivable by human beings, many publications claimed protecting against adversarial examples is important from a security perspective.
Therefore, before presenting the algorithms used to craft adversarial examples, we first discuss the threat models and position the field of adversarial examples in a security context.

\else 

\fi
    \section{Threat Models}
\label{sec:attack_models}

While most publications neglect the threat model when discussing attacks using adversarial examples, we chose to introduce the attacker's goals and capabilities before discussing the attacks because this information \if 0\mode can shape \else helps shaping \fi an initial taxonomy.
Moreover, on the same basis, we introduce the defender's \if 0\mode goals and \else \fi capabilities.
Towards the end of this section, we discuss the need for clear threat models whenever security consequences of adversarial examples are claimed.


\subsection*{Attacker Goals}

Threats are defined in regard to an objective that must be defended.
In our case, "the integrity of the classification is of paramount importance"~\cite{papernot2016limitations}, but not the only property to be defended~\cite{papernot2016towards}.
Some attackers target the availability or the confidentiality of a model.
In the case of adversarial examples, an attacker targets the integrity of a classifier at \emph{inference} time, supplying an input that causes an incorrect output.
In practice, an attacker can cause:
\begin{enumerate}
	\item Confidence reduction. In this case an attacker can reduce the output confidence score of a classifier, thus introducing class ambiguity,
	\item Random misclassification. In this case an attacker modifies an input in order to output \emph{any class} different than the correct one.
	\item Targeted misclassification. In this case an attacker modifies an input in order to output a \emph{specific, target, class}.
\end{enumerate}
\noindent
Although confidence reduction is useful in active learning contexts, it is not explored in the adversarial example literature.

Adversarial examples consist in \emph{modifying} - or crafting - an input before sending it to a classifier.
When the input is known, an attacker strives to modify it as little as possible.
Thus, the input has common elements with a real object.
Other methods from literature generate examples that are not recognisable by human observers and can still lead to a misclassification~\cite{nguyen2015deep}.
The focus of this \if 0\mode report \else paper \fi is on adversarial examples that resemble real life objects because they are the main driver of this research field. 
However, we do not neglect the case of un-recognisable images.
 
\subsection*{Attacker Capabilities}
As mentioned in the past section, an attacker can craft an adversarial example by modifying a normal input or by generating it from noise.
Although crafting an adversarial example by applying a very small perturbation to a normal input will make it not recognisable by human observers, in most cases no humans will supervise a \ac{ML} algorithm.

Attackers can be characterised using the information they can use and the actions they can take.
In the case of \ac{DNN}, the information available to an attacker is related to the training data or the neural network's architecture and parameters.
Since the development of \ac{DNN} models is tightly coupled with powerful hardware, the capability of an attacker can be limited by her inability to train a large model.
However, we exclude this restriction and only consider attackers that are not bounded by hardware constraints.
Therefore, the information available for an attacker, that limits her capabilities is:
\begin{itemize}
	\item Training data, \ac{DNN} architecture and hyper-parameters (\emph{white-box} scenario). In this case an attacker knows everything about the target \ac{DNN}~(architecture, hyper-parameters, weights, \etc), has access to the training dataset and knows about any defence mechanisms employed (\eg~adversarial detection systems). Therefore, an attacker has the ability to \emph{completely} replicate the model under attack.
	\item Knowledge of the training data, the \ac{DNN} architecture, or \emph{some} knowledge about the defences employed (\emph{grey-box} scenario). In this case the attacker can collect some information about the network's architecture (\eg~she knows a certain model uses an open-source architecture), she knows the model under attack was trained using a certain dataset (\eg~the ImageNet~\cite{krizhevsky2012imagenet} dataset is very common for object recognition) or she has information about some defence mechanisms. In any of these cases, the information is neither complete or certain and provides the attacker an ability to \emph{partially} simulate the model under attack.
	\item No knowledge at all (\emph{black-box} scenario). In this case an attacker does not know anything about the model under attack, however, she has the ability to use the model as an \emph{oracle}. Therefore, an attacker can supply \emph{limited} or \emph{unlimited} inputs and collect output information.
\end{itemize}
\noindent
An attacker might also have access to pairs of inputs and outputs collected from a classifier, but no ability to modify the inputs or test them with an oracle.
This scenario can help to extract some information about the model, by reverse-engineering the data, however it is not explored in the field of adversarial examples.

\subsection*{Defender Capabilities}
\label{subsec:defender_goals}
While some publications take into consideration the attacker's capabilities, \eg~\cite{huang2017adversarial, kurakin2016adversarial, papernot2016limitations},  almost none discuss the defender's.
An incipient taxonomy is presented in~\cite{papernot2016towards}, however, it is far from complete.
We consider equally important to define the capabilities of a defender, whenever designing a defence.
As in the case of attackers, defenders aim to protect the model at \emph{inference} time.
However, a defender can choose different processing stages to work with:
\begin{itemize}	
	\item Input pre-processing. In this case, the defender aims to apply some techniques before the inputs reach the \ac{ML} model. Given that a defender knows the model in details, she can train a classifier to either spot adversarial examples early in the pipeline or apply different pre-processing techniques (\eg~geometrical transformations) that can alleviate the impact of adversarial examples.
	\item Hide (obscure) relevant information for attackers. In some cases the defender aims to mitigate weak points of the model that can be exploited by attackers. A common method to generate adversarial examples is to exploit sensitive features, estimated using gradients taken \wrt~to an input. It follows that a natural class of defences seeks to reduce these sensitivities and minimise the gradients during the learning phase. These defences only minimise the impact of adversarial examples and can be defeated by adaptive attacks (as discussed in Section~\ref{subsec:overall_defences}).
	\item Model hardening against small perturbations. A large body of publications focuses on defences against small, unrecognisable, perturbations. Although impossible to detect by human observers, such perturbations can easily fool \ac{DNN}. In this case, the defender can alter the model in any way, aiming to improve its robustness against small perturbations.
	\item Native defence against all types of adversarial examples. This category is, at the moment, hypothetical. It is not yet clear what the existence of adversarial examples tells us about state-of-the-art \ac{DNN} and either it can help us to develop robust \ac{ML} models (or models that will better resemble our perception system). This category leaves room for models that not only aim to defend against small perturbations, but against any kind of perturbations. 

\end{itemize}


\subsection*{The security of adversarial examples in real world scenarios}
Although it was shown that adversarial examples persist during the image acquisition process~\cite{kurakin2016adversarial} and can  be deployed in real-life scenarios~\cite{eykholt2018robust}, the economics of using adversarial examples are often mistaken.
In particular, it is not clear if an attacker would prefer an attack through adversarial examples over other methods (some with no machine learning components)~\cite{gilmer2018motivating}.
For example, in~\cite{eykholt2018robust, lu2017standard} the attacker makes small changes to a stop sign in order to fool autonomous vehicles.
While generating adversarial examples is one way of performing this attack, obscuring the stop sign or going as far as removing it are somehow easier choices. 

Moreover, the motivation for minimal or very small perturbation is often over-emphasised from a security \if 0\mode perspective \else standpoint\fi.
Indeed, human observers can spot big perturbations and act accordingly, however, if machines can not, why enforce this requirement?
Until motivated by real-world scenarios, this constraint is confusing and adds clutter to the field.
In some cases, the adversary will benefit from a less distinguishable attack - for example the attacker could use the same attack for a longer time, before being detected - however this adds as a requirement for the flexibility of an attack~\cite{gilmer2018motivating} and is not always a strong requirement.

Nevertheless, minimal perturbations become important when the property of \emph{robustness} is discussed.
There are real-life scenarios such as sensor wear or small changes in distributions (\eg~sales in some region), which should not drastically impact the behaviour of a model.
We argue these use cases belong to the field of \emph{robustness}, sometimes mistakenly called \emph{safety}, and not security.
Therefore, in the next section we present the safety perspective of adversarial examples and introduce the property of robustness, as discussed in literature.

For a complete taxonomy and overview of real-life scenarios where adversarial examples can be used, see~\cite{gilmer2018motivating}.
In order to claim security consequences for adversarial examples, it is important to specify a complete description of the scenarios and threat models involved.
Moreover, it is important to discuss the economics of using adversarial examples, compared to other methods of achieving the same objective.
The goal of this paper is to characterise the complete research field; therefore, we do not omit defence mechanisms that are poorly motivated or, sometimes, impossible to find in real life scenarios.
However, we signal these drawbacks whenever possible. 
    \section{On Safety, Robustness and Evaluation Methods}
\label{sec:robustness_eval}


From an engineering perspective, safety defines the ability of a system to protect its users from harmful or non-desirable outcomes.
Some methods in literature aim to improve or validate the safety property of \ac{DNN}~\cite{huang2017safety, lu2017safetynet, gehr2018ai}.
However, safety is an inherent property of a system and not of an algorithm solely.
In order to guarantee safety, one should make sure that possible errors are contained and do not impact the overall system.
Take the example of a banking \ac{DL} algorithm that decides if an user should be awarded a financial credit or not.
If the output of such a system is double checked by a human employee, the algorithm can be sensitive to adversarial examples and still be safe to operate.
Safety breaks down to assessing and mitigating the impact of an error and often resembles a risk management framework.
In some cases it is of crucial importance to detect and contain an error as early as possible, in others it is not.
These criteria must drive the discussion about safety of \ac{DNN} and their adoption.

In most publications, when talking about safety, the authors invoke the property of \emph{robustness} \ie~the \emph{insensitivity} of an algorithm to small deviations from the underlying assumptions~\cite{huber2011robust}.
Two general definitions of robustness are valid in this case: (1) \emph{distributional robustness}, defined as insensitivity to \emph{slight deviations} of the underlying distribution from the assumed model~\cite{huber2011robust} and (2) \emph{optimisation robustness}, defined as an algorithm's ability to perform well under a certain level of uncertainty~\cite{ben2009robust}.
This means the uncertainty margins are known beforehand.
In real world applications of optimisation, small uncertainty in the data can heavily affect the quality of the nominal solutions.
Therefore, instead of deploying uncertain solutions, an associated \emph{robust counterpart} is used.
\if 0\mode
At this point we can draw similarities between robustness and factors of safety - described as the capability of a system to perform well beyond maximum loads. 
For example, we would like an autonomous vehicle to perform well in a temperature interval of $\pm 40\deg$. 
However, if the temperature exceeds the upper bound by $1\deg$, the system should perform reasonably well and not instantly crash. 
Safety factors, however, assume some bounds are known and the behaviour within these bounds is proven to be safe.
Therefore, some methods to guarantee safety are still needed.
We advise caution when using the concept of safety as a verifiable property of an algorithm.
\else\fi

When judging adversarial examples through the lenses of the two definitions above, we observe that (1) according to some publications (discussed in \if 0\mode Section~\ref{subsubsec:detection} \else Section~\ref{subsec:reactive}\fi) adversarial examples violate the \ac{iid} assumptions or the first definition of robustness because they are drawn from a different distribution than the training set and (2) the proof of robustness requires to know the uncertainty margins up-front. 
As discussed in \if 0\mode Section~\ref{subsec:minmax}\else Section~\ref{subsec:proactive}\fi, some publications define uncertainty bounds and use them to prove some limits of robustness.
The problem is, however, not trivial and still not solved in a scalable fashion.
In this context, it is important to decide if we want to guarantee performance for inputs drawn from a different distribution or to develop the robust counterpart of a model.
The problem is sometimes context dependent and a scalable, universal, solution is lacking, therefore many interesting research directions can spark from it. 

\if 0\mode \bigskip \else  \newpara  \fi
Several definitions and methods to prove robustness have been proposed in the adversarial example literature, sometimes causing misleading evaluation methods.
We reiterate the defender's goal \wrt~small perturbations: one can provide a lower-bound guarantee which searches for the smallest norm ball around an input where no adversarial examples can be found or an upper-bound guarantee that perturbations smaller than a threshold can not cause misclassifications. 

In the adversarial examples inception paper, Szegedy \etal~\cite{szegedy2013intriguing} introduced a method to measure the stability of a network using spectral analysis of each layer.
Under the assumption that all layers are Lipschitz continuous, one can inspect the upper Lipschitz constant for each layer.
It follows that a lower bound stability measure can be derived for a \ac{DNN}, by multiplying the Lipschitz upper bounds of each layer.
However, this global Lipschitz constant often gives a very loose bound \cite{weng2018evaluating}.

Fawzi \etal~\cite{fawzi2018analysis} propose to average over the minimal perturbations required to cause a misclasification, for each example in the dataset:
\if 0\mode
\begin{equation}
	\rho(f, \vx) = \mathbb{E}_{f(\vx) \sim p_{data}} [\| \eta \|_{p}],
\end{equation}
\noindent \else $	\rho(f, \vx) = \mathbb{E}_{f(\vx) \sim p_{data}} [\| \eta \|_{p}],$
and provide a theoretical upper bound guarantee for linear and quadratic classifiers.
However, this boundary can also be affected by distribution drifts.\fi

Bastani \etal~\cite{bastani2016measuring} provide a formalism for lower bound robustness to adversarial examples, independent of the Lipschitz constant. 
At first, they formalise the notion of robustness at point $\vx$~-~called \emph{point-wise robustness}~-~the minimal threshold for which there exists at least one adversarial example.
Given $f(\vx)$~-~a classification function, $f$ is considered \emph{$(\vx,\alpha)$-robust} if for every input $\vx$ and every adversarial examples generated from it $\vx'$ such that $\|~\vx'~-~\vx~\|_{p}~\leq~\alpha$, $f(\vx')~=~f(\vx)$.
It follows that point-wise robustness is the minimal $\alpha$ for which $f$ is not \emph{$(\vx',\alpha)$-robust}:
\if 0\mode
\begin{equation}
	\rho(f, \vx) = \inf \{ \alpha \geq 0 \mid \|\vx'-\vx\|_{p} \leq \alpha \text{  and } f(\vx') \neq f(\vx) \}.
\end{equation}
\noindent
\else $\rho(f, \vx) = \inf \{ \alpha \geq 0 \mid \|\vx'-\vx\|_{p} \leq \alpha \text{  and } f(\vx') \neq f(\vx) \}.$ \fi
Point-wise robustness can help us reason about a lower bound specific only to \emph{one} adversarial example.
In order to reason about the data generation distribution, $p_{data}$, Bastani \etal~\cite{bastani2016measuring} introduce \emph{adversarial frequency}, defined as the probability mass function when $f$ is not \emph{$(\vx,\alpha)$-robust}, upper bounded by a parameter $\epsilon$.
Formally, adversarial frequency is defined as:
\if 0\mode
\begin{equation}
	\phi(f, \epsilon) = P_{\vx \sim p_{data}} [\rho(f, \vx) \le \epsilon].
\end{equation}
\noindent
\else $\phi(f, \epsilon) = P_{\vx \sim p_{data}} [\rho(f, \vx) \le \epsilon].$ \fi
Through $\epsilon$ we can control the size of the perturbation, which can impact the subtlety of the adversarial example. 
For $\epsilon$ small enough, the perturbation is not detectable by a human observer.

Lastly, Bastani \etal~\cite{bastani2016measuring} define \emph{adversarial severity} as the average minimal space where $f$ fails to be robust, conditioned by an upper bound $\epsilon$:
\if 0\mode
\begin{equation}
	\mu(f, \epsilon) = \mathbb{E}_{f(\vx) \sim p_{data}} [\rho(f, \vx) \mid \rho(f,\vx) \le \epsilon].
\end{equation}
\noindent
\else $\mu(f, \epsilon) = \mathbb{E}_{f(\vx) \sim p_{data}} [\rho(f, \vx) \mid \rho(f,\vx) \le \epsilon].$ \fi
\if 0\mode
\emph{Smaller} values for $\mu(f,\epsilon)$ correspond to \emph{worse} adversarial severity, because $f$ is more susceptible to misclassifications when the average distance to the nearest adversarial example is small.
Frequency and severity capture different robustness behaviours.
A neural network may have high adversarial frequency, but low adversarial severity; indicating that most adversarial examples are about $\epsilon$ distance away from the original point $\vx$.
Analogously, a neural network may have low adversarial frequency but high adversarial severity, indicating that it is typically robust, but occasionally severely fails to be robust.
Frequency is, therefore, more important, because a neural network with low adversarial frequency is robust most of the time.
\else\fi
The generalisation for point-wise robustness extension is, however, still limited by an upper bound on the perturbation.
On contrast, Weng \etal~\cite{weng2018evaluating} developed CLEVER, an \emph{attack agnostic} metric to measure lower bound robustness, based on Lipshitz continuity. 
CLEVER generalises a metric introduced by Hein and Andriushchenko~\cite{hein17formal} for kernel methods and neural networks with only one layer.
Consider a classifier $f(\vx)$  with continuously differentiable components $f_i$ and define the class which $f$ predicts for an input $\vx_0$ as $l = \argmax_{1\leq i \leq K} f_i(\bm{x_0})$, then the lower bound robustness of $f$ is defined as:
\if 0\mode 
\begin{equation}
	\label{eq:our_delta_bnd}
	\beta_L=\min_{l' \neq l} \frac{f_l(\vx_0)-f_l'(\vx_0)}{L_q^{l'}},
\end{equation}
\noindent
\else $	\beta_L=\min_{l' \neq l} \frac{f_l(\vx_0)-f_l'(\vx_0)}{L_q^{l'}},$ \fi
where $L_q^{l'}$ is the Lipschitz constant for the function $f_l(\vx)-f_{l'}(\vx)$ in $\ell_p$ norm. 
Weng \etal~propose to use extreme value theory in order to approximate $\beta_L$. 
However, Goodfellow~\cite{goodfellow2014explaining} show that CLEVER fails to correctly estimate lower bound robustness, even in theoretical settings.

%

The question of measuring robustness remains an open and non tractable problem.
Some publications, presented in \if 0\mode Section~\ref{subsubsec:provable}\else Section~\ref{subsec:proactive}\fi, search for methods to approximate the lower bound and, thus, certify that \ac{DNN} can operate normally within some bounds.
In practice, most publications use accuracy to argue that attacks or defences are effective or in order to evaluate robustness of \ac{DNN}.
This method, however provides no certainty that an algorithms is robust within some bounds and it is too simplistic for use in security contexts.
\if 0\mode 
Moreover, as shown in Appendix~\ref{sec:benchmarks} and discussed in Section~\ref{subsec:overall_defences}, most publications use different datasets and models (most of the times smaller in capacity than state-of-the-art) to evaluate accuracy against adversarial examples.
As a consequence, it is very hard to compare methods.
Valuable insights about future evaluation techniques are provided in Section~\ref{sec:distilled_knowledge}.
Meanwhile, we use this observation when classifying attacks and defences.
\else\fi


    \section{Hypotheses on the Existence of Adversarial Examples}
\label{sec:causes}

\paragraph{Initial hypothesis.} At first, adversarial examples were though to represent low-probability "pockets in the data manifold, hard to efficiently reach by sampling the input space around a given example"~\cite{szegedy2013intriguing}.
Searching for a solution to \eqq~(\ref{eq:adversarial_generic}) enables the input space to be efficiently spanned in search for adversarial examples.
State-of-the-art \ac{DNN} models already employ input transformation during training, in order to increase their robustness.
However, the transformed inputs are highly correlated and drawn from the same distribution across the training set.
Adversarial examples are thought to be neither correlated or identically distributed, thus leading to the theory that they lie in 'pockets' of the data manifold.

Gu \etal~\cite{gu2014towards} later investigated the size of such pockets and discovered they are relatively large in input space volume, and locally continuous.
The authors hypothesised that sensitivity to adversarial examples relates to "intrinsic deficiencies in the training procedure and objective function, rather than to model topology"~\cite{gu2014towards}.
Therefore, coming up with a training procedure that can efficiently output regions with low variance around training data can solve this issue.

\paragraph{The linearity hypothesis.} Goodfellow \etal~\cite{goodfellow2014explaining} refuted the hypothesis that adversarial examples lie in small regions of the data manifold and advanced the conjecture that adversarial examples span large and high-dimensional regions.
The authors argued adversarial examples exist because \ac{DNN} have, in fact, very linear behaviour despite non-linear transformations within hidden layers.
The choice for activation functions that are easy to optimise (\eg~ReLU) drive \ac{DNN} to behave more linear.
Therefore, summing small perturbations in all dimensions of a high dimensional input forces the entire sum in a direction that will likely cause a misclassification.
This hypothesis lead to the discovery of very efficient methods to generate adversarial examples, as discussed in Section~\ref{subsec:sensitivity}.
Empirical evidence for the linearity hypothesis was also brought by~\cite{tabacof2015exploring, tramer2017space, krotov2017dense}.
Lou \etal~\cite{luo2015foveation} proposed a variant of this conjecture in which \ac{DNN} operate linearly in certain regions of the input manifold, but non-linear in others.

\paragraph{Evolutionary stalling.} Rozsa \etal~\cite{rozsa2016towards} believe that the gradients of correctly classified inputs diminish during training and fail to create flat regions around the training data.
Therefore, most training data lies close to a decision boundary and small perturbations are able to push inputs over the boundary.
The authors hypothesise that coming up with a training algorithm that will avoid this phenomenon will mitigate the threat to adversarial examples. 
However, as discussed in Section~\ref{subsec:proactive} (architectural defences), imposing constraints on the gradients is not an efficient defence. 

\paragraph{The boundary tilting hypothesis.} Tanay and Griffin~\cite{tanay2016boundary} challenge the linear hypothesis as not 'convincing'.
At first, because small perturbations are taken relatively to the activations, which increase linearly to the problem.
Therefore, the ratio between inputs and perturbations remains constant.
Secondly, the authors argue that linear behaviour is not sufficient to explain the adversarial examples phenomenon and demonstrate the possibility to build linear models that are not sensitive to adversarial examples.

In contrast, the authors propose the \emph{boundary tilting perspective}, based on the assumption that a learned class boundary lies close to the data manifold, but the boundary is tilted with respect to it. 
Adversarial examples can then be found by perturbing points from the data manifold towards the classification boundary until the perturbed input crosses the boundary. 
If the boundary is only slightly tilted, the distance required by the perturbation to cross the decision-boundary is very small, leading to strong adversarial examples that are visually almost imperceptibly close to the data.
The authors argue that adversarial examples are likely to occur along directions of low variance in the data and thus speculate that adversarial examples can be considered an effect of an overfitting phenomenon that could be alleviated through regularisation.

\paragraph{Relation to decision boundaries.} Moosavi-Dezfooli \etal~\cite{moosavi2017analysis} showed that it is possible to generate an universal perturbation, that can be applied to all inputs.
When investigating the phenomenon, the authors hypothesised that adversarial examples exploit "geometric correlations between different parts of the decision boundaries"~\cite{moosavi2017analysis}.
Precisely, the authors suggest the existence of a low dimensional sub-space which contains the vectors normal to the decision boundaries around an input.
However, the frequency of these sub-spaces is not analysed.

Fawzi \etal~\cite{fawzi2016robustness} examined sensitivity to adversarial examples in relation to the curvature of decision boundaries.
Their results show that a small curvature in the decision boundary increases the classifier's robustness to adversarial examples.
Thus, it is assumed that limiting the curvature of decision boundaries can increase sensitivity to adversarial examples
A similar hypothesis is proposed in~\cite{tramer2017ensemble} and more theoretical analysis presented in~\cite{moosavi-dezfooli2018robustness}.

\paragraph{Not i.i.d hypothesis.} A different hypothesis assumes that adversarial examples lie off the data manifold, and are sampled from a different distribution~\cite{song2017pixeldefend, meng2017magnet, ghosh2018resisting, lee2017generative}.
This hypothesis lead to the proposal of adversarial detection methods \if 0\mode (Section~\ref{subsubsec:detection}) \else (Section~\ref{subsec:reactive}) \fi and the attempt to learn this new distribution with generative models.
While interesting in nature, because the proof of this hypothesis means adversarial examples break the \ac{iid} assumption, more empirical data is needed.
Moreover, Carlini and Wagner~\cite{carlini2017adversarial} question this hypothesis by developing attacks that are can easily by-pass the detector.

\paragraph{The manifold geometry hypothesis.} Gilmer \etal~\cite{gilmer2018adversarial} hypothesise that adversarial examples are a result of the high-dimensional (and possibly intricate) geometry of the data manifold.
The authors use a synthetic dataset that can be better explored and find that whenever the classifier has the slightest test error, most data points in the input distribution which get correctly classified lie in the neighbourhood of a misclassified input.
Therefore, whenever training is performed on an approximation of the real distribution, the model is sensitive to adversarial examples.
This result raises the question if the sensitivity to adversarial examples could ever be removed.
Moreover, the authors deny the hypothesis that adversarial examples lie off the data manifold.
Nevertheless, the research is still incipient and only applied to small use-cases~-~the assumptions still have to find their proof on real-world tasks such as object or speech recognition.




\paragraph{Relation to training dataset.} In a very recent work, Schmidt \etal~\cite{schmidt2018adversarially} show that the sample complexity for robust adversarial learning can be significantly higher than for normal learning. 
Therefore, in order to train robust \ac{DNN} for complicated tasks, such as object recognition for over 1000 classes, the available datasets are not sufficient.
Culina, Bhagoji and Mittal~\cite{cullina2018pac}, however, show that the sample complexity of PAC-learning when the hypothesis class is the set of half-space classifiers does not increase in the presence of adversaries bounded by convex constraint sets.
Tsipras \etal~\cite{tsipras2018there} present a somehow orthogonal result that shows there is a trade-off between achieving robustness and accuracy. The authors suggest this trade-off will prevail, independent of the \ac{ML} model.

\if 0\mode \bigskip \else\fi
In summary, although Tanay and Griffin~\cite{tanay2016boundary} claim to refute the linearity hypothesis, they do not bring enough empirical evidence.
One can argue that linear transformations in high dimensional spaces can be sufficient to move a sample in the direction of a tilting boundary.
It is clear that the linearity hypothesis is not sufficient to explain the adversarial examples phenomenon, but it plays an important role as some attacks \if 0\mode (\eg~FGSM~-~Section~\ref{subsubsec:fgsm}, DeepFool~-Section~\ref{subsubsec:deepfool}) \else (\eg~FGSM~-~Section~\ref{subsec:sensitivity}, DeepFool~-~\ref{subsec:optimisation_attacks})\fi make extensive use of this hypothesis, with great success.

That adversarial examples are a consequence of the complicated geometry of the data manifold, which can lead to sub-spaces where adversarial examples lie is a conjecture that holds.
Mainly because there is no study to indicate that adversarial examples lie off the data manifold.
Some publications, which assumed this claim, developed adversarial detectors with some degree of accuracy.
However, none brings sufficient evidence to prove adversarial examples lie off the data manifold.

It is not yet settled if these sub-spaces are small or large. 
According to Gilmer \etal~\cite{gilmer2018adversarial} they are proportional with the testing error and the capacity of a model to correctly approximate the input distribution.
However, more evidence is needed to support this conjecture for models with high capacity.

Some publications suggest there are limits to adversarial robustness~\cite{gilmer2018adversarial, fawzi2015fundamental} and even that sensitivity to adversarial examples can not be removed~\cite{gilmer2018adversarial}.
This is an interesting hypothesis that sparks several questions regarding the research in adversarial examples.
At first, if theoretical bounds for robustness can be obtained, are they significant to claim security consequences?
Secondly, what are the incentives to achieve these bounds in practice?
And lastly, assuming such bounds will lead to adversarial examples similar to normal inputs, what  do this field tells us about constructing computer vision models that resemble the human perceptual system?
Which is the impact of adversarial examples, if any, to fundamental computer vision research?
We argue that more fundamental research, similar to~\cite{gilmer2018adversarial, schmidt2018adversarially, cullina2018pac}, is needed in order to explain both the causes and the effects of this peculiar behaviour of \ac{DNN}.

    \section{Attacks}
\label{sec:attacks}

\if 0\mode

In this section we aim to fulfil three objectives: (1) provide an overview of the methods used to generate adversarial examples,  (2) classify them according to common characteristics and (3) describe all important attacks in detail, such that we can later use this section as a catalog of attacks.

Because most characteristics of adversarial examples are orthogonal, some classes will overlap (and might lead to confusion).
At a high level, there are two classes of algorithms used to craft adversarial examples: (M) attacks which \emph{modify} a real input - \eg~an image - and (G) attacks which \emph{generate} adversarial inputs from noise. 
Following the hypothesis presented in Section~\ref{sec:causes}, of the first class we distinguish between (OP) attacks that efficiently search the input space for adversarial examples using \emph{optimisation} techniques (Section~\ref{subsec:optimisation_attacks}) and (SA) attacks that exploit \emph{sensitive features} (Section~\ref{subsec:sensitivity}).
In some cases, (GT), adversarial examples can be generated only using \emph{geometric transformations} (Section~\ref{subsec:geometric_attacks}).
\emph{Generative models} (GM) aim to learn the  adversarial or input data generation distribution and use it in order to convert inputs to adversarial examples or generate adversarial examples from noise (Section~\ref{subsec:gan_attacks}).

One level deeper, we distinguish between attacks that use (IT) \emph{iterative} or (SS) \emph{single-shot} methods~-~thus illustrating a tradeoff between precision and speed.
Some algorithms approximate the minimal perturbation, at the cost of speed, while others use very fast methods, trading precision.
Fast algorithms are used to enhance the training set with adversarial examples - a procedure called adversarial training and presented in Section~\ref{subsec:adv_training} - but are weaker and easier to protect against.
Iterative algorithms require more processing time, but are powerful and hard to protect against.

A dominant theme in the publications about adversarial examples is to reduce or minimise the size of the perturbation.
As discussed in Section~\ref{sec:attack_models}, this requirement needs better motivation and clear threat models.
Without information about the context and the economics of an attack, searching for minimal perturbations is futile.
In Section~\ref{subsubsec:unrecognisable} we present an attack that uses evolutionary algorithms to generate inputs that are indistinguishable for human observers, but still fool \ac{DNN}.

Further on, when classifying adversarial attacks we have to consider the adversarial goals and capabilities introduced in Section~\ref{sec:attack_models}.
At first, given the adversarial goals, one could aim to generate a \emph{non-targeted} (NTG), random, misclassification or a \emph{targeted} (TG) misclassification.
This goal often shapes the algorithms used to generate adversarial examples because simpler methods (such as single-shot sensitivity attacks) are fitted for non-targeted attacks, but harder to adapt to targeted attacks.
Similarly, iterative attacks are sometimes harder to implement and require more processing time, therefore they are more qualified for targeted attacks.
Once again, the economics of carrying an attack matters. 
Without clear security requirements, it is impossible to draw a line between the efficiency of any of the methods.

Secondly, given the knowledge available to an attacker, we distinguish between \emph{white-box} (WB), \emph{grey-box} (GB) and \emph{black-box} attacks (BB).
Most attacks assume full-knowledge of the model under attack and, therefore, fall in the white-box category.
Therefore, we will present grey and black-box attacks in a separate section (Section~\ref{subsec:black_box}), although they sometimes use optimisation methods.
The rather small number of black-box attacks, in contrast with white-box attacks, is another way of judging the security implications of adversarial examples.

We note again that most attack traits are orthogonal and the existence of one does not imply the absence of others.
For example, one may generate an iterative, targeted, white-box attack as well as an iterative, not-targeted, grey-box attack.
Selecting a method is, therefore, context dependent and requires a clear threat model.
Unfortunately, most publications do not present any threat model, therefore, we have to accept some overlap in the classification.

Lastly, we distinguish between \emph{specific} attacks (SP)~-~that aim to modify or generate an input against able to fool \emph{one} model~-~and \emph{universal} attacks (UN)~-~that aim to modify or generate an input that can work against \emph{all} models.

Before delving into the details of each attack, we present an overview in Table~\ref{tbl:attacks} where we enumerate all attacks in relation to their features.
We observe that most attacks are based on modifying a known input.
We also note that the most powerful attacks~-~Carlini and PGD~-~are based on optimisation methods.

In the following sections we introduce, in details, the most important attacks.
We try to extensively cover all attacks in the current literature, however, new attacks might be rolled out after writing this paper.
The classification follows the criteria introduced earlier.

\if 0\mode
\begin{landscape}
\begin{table}
	\centering
	\begin{tabular}{|l|l|l|l|l|l|l|}
		\toprule%
		Attack & \specialcell{Modify (M) or \\ Generate (G) \\ Input} &%
		  		 \specialcell{Optimisation (OP), \\ Sensitivity (SA), \\ Geometric \\ Transformations (GT) \\ Generative Models (GM)} &%
		  		 \specialcell{Targeted (TG), \\ Non-Targeted (NTG)} &%
		  		 \specialcell{Single-Shot (SS), \\ Iterative (IT)} &%
  				 \specialcell{White-box (WB),  \\ Grey-box (GB), \\ Black-box (BB)}&%
	 			 \specialcell{Specific (SP), \\ Universal (UN)} \\
		\midrule
	L-BFGS~\cite{szegedy2013intriguing} & M & OP & TG & IT & WB & SP  \\ \hline
	Deep Fool~\cite{moosavi2016deepfool} & M & OP & NTG & IT & WB & SP \\ \hline
	UAP~\cite{moosavi2017universal} & M & OP & NTG & IT & WB & UN \\ \hline
	Carlini~\cite{carlini2017towards} & M & OP & TG / NTG & IT & WB & SP  \\ \hline 
	CFOA (Madry / PG)~\cite{madry2017towards} & M & OP & TG / NTG & IT & WB & SP   \\ \hline 
	STA~\cite{huang2015learning} & M & OP & TG / NTG & IT & WB & SP  \\ \hline
	ZOO~\cite{chen2017zoo} & M & OP & TG / NTG & IT & BB & SP  \\ \hline
	IS~\cite{narodytska2017simple} & M & OP &  TG / NTG & IT & BB & SP   \\ 
\specialrule{1pt}{1pt}{1pt}
	FGS~\cite{goodfellow2014explaining} & M & SA & NTG & SS & WB & SP  \\ \hline
	JSMA~\cite{papernot2016limitations} & M & SA & TG & IT & WB & SP  \\ \hline
	RSSA~\cite{tramer2017ensemble} & M & SA & NTG & SS / IT & WB & SP  \\ \hline
	BPDA~\cite{athalye2018obfuscated} & M & SA & TG & IT & WB & SP  \\ \hline
	Elastic-Net~\cite{chen2017ead} & M & SA & TG & IT & WB & SP  \\ \hline 
	BI~\cite{kurakin2016adversarial} & M & SA & NTG & IT & WB & SP  \\ \hline
	ILC~\cite{kurakin2016adversarial} & M & SA & TG & IT & WB & SP  \\ \hline
	Momentum~\cite{dong2017boosting} & M & SA & NTG & IT & WB & SP  \\  	\hline	
	Substitute~\cite{papernot2017practical} & M & SA & TG & SS / IT & BB & SP \\ 
	\specialrule{1pt}{1pt}{1pt}
	\specialcell{Rotation Tr.}~\cite{engstrom2017rotation} & M & GT & NTG & SS / IT & WB / GB & SP  \\ \hline
	ManiFool~\cite{kanbak2017geometric} & M & GT & TG / NTG & IT & WB & SP  \\ \hline
	\specialcell{Spatial Tr.}~\cite{xiao2018spatially} & M & GT  & TG & IT & WB & SP  \\  
	\specialrule{1pt}{1pt}{1pt}
	ATN~\cite{baluja2017adversarial} & G & GM & TG / NTG & IT & WB & SP  \\ \hline
	NAE~\cite{zhao2017adversarial} & G & GM & TG & IT & WB & SP \\ 
	\bottomrule
	\end{tabular}	
	\caption{Catalog of Adversarial Attacks.}	
	\label{tbl:attacks}
\end{table}
\end{landscape}

\else 

\begin{landscape}

\begin{table}
	\centering
	\begin{tabular}{|l|l|l|l|l|l|l|}
		\toprule%
		Attack & \specialcell{Modify (M) or \\ Generate (G) \\ Input} &%
		  		 \specialcell{Optimisation (OP), \\ Sensitivity (SA), \\ Geometric \\ Transformations (GT) \\ Generative Models (GM)} &%
		  		 \specialcell{Targeted (TG), \\ Non-Targeted (NTG)} &%
		  		 \specialcell{Single-Shot (SS), \\ Iterative (IT)} &%
  				 \specialcell{White-box (WB),  \\ Grey-box (GB), \\ Black-box (BB)} \\
		\midrule
	L-BFGS~\cite{szegedy2013intriguing} & M & OP & TG & IT & WB   \\ \hline
	Deep Fool~\cite{moosavi2016deepfool} & M & OP & NTG & IT & WB  \\ \hline
	UAP~\cite{moosavi2017universal} & M & OP & NTG & IT & WB \\ \hline
	Carlini~\cite{carlini2017towards} & M & OP & TG / NTG & IT & WB   \\ \hline 
	CFOA (Madry / PG)~\cite{madry2017towards} & M & OP & TG / NTG & IT & WB    \\ \hline 
	STA~\cite{huang2015learning} & M & OP & TG / NTG & IT & WB   \\ \hline
	ZOO~\cite{chen2017zoo} & M & OP & TG / NTG & IT & BB \\ \hline
	IS~\cite{narodytska2017simple} & M & OP &  TG / NTG & IT & BB    \\ 
\specialrule{1pt}{1pt}{1pt}
	FGS~\cite{goodfellow2014explaining} & M & SA & NTG & SS & WB   \\ \hline
	JSMA~\cite{papernot2016limitations} & M & SA & TG & IT & WB   \\ \hline
	SV-UAP~\cite{khrulkov2017art} & M & SA & NTG & IT & WB \\ \hline
	RSSA~\cite{tramer2017ensemble} & M & SA & NTG & SS / IT & WB   \\ \hline
	BPDA~\cite{athalye2018obfuscated} & M & SA & TG & IT & WB   \\ \hline
	Elastic-Net~\cite{chen2017ead} & M & SA & TG & IT & WB   \\ \hline 
	BI~\cite{kurakin2016adversarial} & M & SA & NTG & IT & WB   \\ \hline
	ILC~\cite{kurakin2016adversarial} & M & SA & TG & IT & WB   \\ \hline
	Momentum~\cite{dong2017boosting} & M & SA & NTG & IT & WB   \\  	\hline	
	Substitute~\cite{papernot2017practical} & M & SA & TG & SS / IT & BB  \\ 
	\specialrule{1pt}{1pt}{1pt}
	\specialcell{Rotation Tr.}~\cite{engstrom2017rotation} & M & GT & NTG & SS / IT & WB / GB   \\ \hline
	ManiFool~\cite{kanbak2017geometric} & M & GT & TG / NTG & IT & WB   \\ \hline
	\specialcell{Spatial Tr.}~\cite{xiao2018spatially} & M & GT  & TG & IT & WB   \\  
	\specialrule{1pt}{1pt}{1pt}
	ATN~\cite{baluja2017adversarial} & G & GM & TG / NTG & IT & WB   \\ \hline
	NAE~\cite{zhao2017adversarial} & G & GM & TG & IT & WB \\ 
	\bottomrule
	\end{tabular}	
	\caption{Catalog of Adversarial Attacks.}	
	\label{tbl:attacks}
\end{table}

\end{landscape}
\fi


\subsection{Attacks based on Optimisation Methods (OP)}
\label{subsec:optimisation_attacks}

In this section we review attacks which use optimisation methods in order to search for adversarial examples.
We start with the initial paper, which revealed the sensitivity of \ac{DNN} to adversarial examples.
Later, we introduce a series of attacks that use optimisation techniques similar to sensitivity analysis - \eg~gradient descent - but with different objectives.

\subsubsection{The L-BFGS Attack}
\label{subsubsec:lbfgs}

Szegedy \etal~\cite{szegedy2013intriguing} were the firsts to discover the phenomenon and coin the term \emph{adversarial examples}.
The authors propose an equivalent formalism for \eqq~(\ref{eq:adversarial_generic}) and use limited memory box constrained optimisation (L-BFGS) in order to approximate the minimal perturbation needed to change a label.
They propose a targeted attack, formally defined as follows. Given $c > 0$ and a target label $l'$:
\begin{equation}
	\begin{aligned}
	\label{eq:lbfgs}
		&\min_{\veta} & & c \| \veta \|_2 + J(\vtheta, \vx + \veta, y), \\
		&s.t. & & \veta \in [0,1]^{m}, \\
		& & & f(\vx + \veta) = l'. \\
	\end{aligned}
\end{equation}
\noindent
The final adversarial example is defined as $x' = x + \veta$.
Experimental results show the minimal average distortion drops as low as $0.062$, which means adversarial examples are almost identical to normal inputs.
The authors also propose to augment the training set with adversarial examples in order to increase the robustness to adversarial examples.
However, L-BFGS is a slow procedure, making adversarial training almost impossible for big datasets, where one would have to find adversarial examples for all inputs.
\subsubsection{The Deep Fool Attack}
\label{subsubsec:deepfool}

The Deep Fool~\cite{moosavi2016deepfool} attack is equivalent to moving a sample towards a hyperplane that separates two classes. 
In this case, the shortest distance from a sample $\vx_0$ to the separation hyperplane is equivalent to the algorithm's robustness for sample $\vx_0$ (similar to point-wise robustness in Section \ref{sec:robustness_eval}).
If we assume that a classifier behaves in a linear fashion, this distance is equivalent to the orthogonal projection of a point on a plane (or separation line).
For a binary, linear, classifier we can define the minimal perturbation as:
\begin{equation}
	\begin{aligned}
		\eta_{\vx_0} = \frac{f(\vx_0)}{\| \vw \|_2^2}\vw,
	\end{aligned}
	\label{eq:deep_fool_linear_binary}
\end{equation}
\noindent
where $\vw$ is equivalent to the \emph{geometrical normal}. 

Further on, for a general differentiable function, we assume an implicit equation\footnote{An implicit equation is a function of form f(x\textsubscript{1}, x\textsubscript{2}, ..., x\textsubscript{n}) = 0} defining the surface that separates two classes such that the geometrical normal is equal to its gradient vector.
The Deep Fool algorithm then takes an iterative approach assuming that, at every iteration $i$, the classification function is linearised around the input and computes the minimal perturbation as:
\begin{equation}
	\begin{aligned}
		\eta_{\vx_i} = \frac{f(\vx_i)}{\| \nabla f(x_i) \|_2^2}\nabla f(x_i).
	\end{aligned}	
	\label{eq:deep_fool_binary}
\end{equation}
\noindent
The final perturbation is computed as the sum of all intermediate perturbations $\eta_i$.

The generalisation to linear multi-class classifiers follows from the example above.
If instead of a single separation surface there are many, one can imagine that the initial sample $\vx_0$ is mapped to a region equivalent to a polyhedron, $P$, whose faces are discriminants from other classes.
Following the same rationale, by computing the distance from $\vx_0$ to one of the faces of $P$ one can measure the classifier's robustness for input $\vx_0$.
In practice, the attack selects the closest hyperplane to the boundary of $P$ and projects $\vx_0$ on its surface.
Formally, the minimal perturbation for a linear multi-class classifier is:
\begin{equation}
	\eta_{\vx_0} = \frac{\left|f_{\hat{y}(\vx_0)}(\vx_0)-f_{y(\vx_0)}(\vx_0)\right|}{\| \vw_{\hat{y}(\vx_0)}-\vw_{y(\vx_0)}\|_2^2}(\vw_{\hat{y}(\vx_0)}-\vw_{y(\vx_0)}),
	\label{eq:deep_fool}
\end{equation}
\noindent	
where $\hat{y}(\vx_0)$ is the closest face of $P$ to $\vx_0$.
Further more, the generalisation to a differentiable multi-class classifier can be obtained using the same rationale as for \eqq~(\ref{eq:deep_fool_binary}) in \eqq~(\ref{eq:deep_fool}).
An important observation is that Deep Fool can converge on the discriminant hyperplane. 
In order to "push" the classifier over the border, the final perturbation is multiplied by a small constant.
Another important observation is that Deep Fool assumes that classifiers behave in a linear fashion, according to the linearity conjecture from Section~\ref{sec:causes}.
\subsubsection{Attacks using \ac{uap}}
\label{subsubsec:universal}

Moosavi-Dezfooli \etal~\cite{moosavi2017universal} demonstrate the existence of image and \ac{DNN} agnostic perturbations that cause misclassification with high probability.
The algorithm iteratively applies the Deep Fool attack (Section \ref{subsubsec:deepfool}) for all images in the training dataset, until one perturbation can fool a large part of the training set.
Formally, given the input distribution $\hat{p}_{data}$, the algorithms searches for an upper-bounded universal perturbation $\eta$ such that:
\begin{equation}
	\begin{aligned}
	\label{eq:universal}
		& & \| \eta \| &  \le \epsilon  \text{    and}, & \\
		& & \underset{\vx \sim \hat{p}_{data}}{\mathbb{P}} ( f(\vx+\eta) &  \neq f(\vx) ) \geq 1 - \delta, &
	\end{aligned}
\end{equation}
\noindent
where $\delta$ quantifies the desired fooling rate for all images sampled from the distribution $\hat{p}_{data}$.
At each iteration step, the perturbation is found using the Deep Fool attack:
\begin{equation}
	\begin{aligned}
		\eta' & \gets \arg\min_{\eta_{\vx_i}} \| \eta_{\vx_i} \|_{2} \text{ s.t. } f(\vx_i + \eta_{\vx_{i-1}} + \eta_{\vx_i}) \neq  f(\vx_i), 		
	\end{aligned}
\end{equation}
\noindent
where $\eta_{\vx_i}$ is the output of Deep Fool.
In order to ensure the upper bound constraint, the perturbation is projected on the $l_p$ ball of radius $\epsilon$ and centred at 0, leading to the following update rule:
\begin{equation}
	\begin{aligned}
		\eta & = \arg\min_{\eta'} \| \eta_{\vx_{i-1}} - \eta' \| _2\text{ subject to } \| \eta' \|_p \leq \epsilon.
	\end{aligned}
\end{equation}
\noindent
Several passes over the training dataset are performed in order to improve the quality of an universal perturbation.
The algorithm is able to find multiple universal perturbations that are different in nature and can fool \ac{DNN} with high accuracy.

\subsubsection{The Carlini $L_{p}$ Attacks}
\label{subsubsec:carlini}

Carlini and Wagner~\cite{carlini2017towards} propose three new attacks based on the $L_{p}$ distance metrics discussed in Section~\ref{subsec:norm_balls}.
All three attacks use upper-bound robustness and are a variance of \eqq~(\ref{eq:adversarial_generic}), where the classification function is replaced in the constraints set.
Because the classification constraint is highly non-linear, the authors propose to replace it with another function such that $f(\vx + \eta) = l'$ if and only if $f(\vx + \eta) \le 0$.
Thus, the optimisation objective becomes:
\begin{equation}
	\begin{aligned}
	\label{eq:adversarial_carlini}
		&\min_{\eta} & & \|\eta\|_p + c \cdot f(\vx+\eta),\\
		&s.t. & & \vx + \eta \in [0,1]^n,
	\end{aligned}
\end{equation}
\noindent
where $c$ is a constant determined empirically through binary search.
In order to ensure the optimisation result yields a valid image, $\eta$ is constrained as follows: $0 \le x_i+\eta_i \le 1$ for all $i$ (box-constraints).
However, in order to use optimisation algorithms that do not support box constraints, the authors optimise $\eta$ through a change of variable, $w$, such that:
\begin{equation}
	\label{eq:distorsion_carlini}
	\eta= \frac12 (\tanh(w) + 1) - \vx.
\end{equation}
\noindent
Since $-1 \le \tanh(w) \le 1$, it follows that $0 \le \vx+ \eta \le 1$, so the solution will automatically be valid.

The authors then substitute \eqq~(\ref{eq:distorsion_carlini}) in \eqq~(\ref{eq:adversarial_carlini}) and optimise for $L_{0}, L_{2} \text{ and } L{\infty}$ norms.
They are able to craft performant adversarial examples that remain state-of-the-art today.
\subsubsection{\ac{cfoa}}
\label{subsubsec:madry}

Madry \etal~\cite{madry2017towards} study the adversarial robustness of \ac{DNN} through the lens of robust optimisation.
The authors propose an attack and a defence mechanism formulated in a min-max fashion.
Because the attack can be independently used, we discuss it here and outline the defence in Section~\ref{subsubsec:madry_learning}.

Assuming for each data point in the training set $\vx_i$ a set of allowed perturbations $ \mathcal{U} \subseteq \R^d $ is known, the attack searches for perturbations that maximise the loss function across the training dataset. 
For the image classification task, $\mathcal{U}$ is chosen to be the $L_\infty$-ball around the normal input $\vx$, limited by a constant $\epsilon$.
The attack is formalised for all training set as:
\begin{equation}
\label{eq:minmax}
	\rho(\theta) = \mathbb{E}_{(\vx,y)\sim p_{data} }\left[\max_{\eta \in \mathcal{U}} J(\vtheta, x + \eta, y) \right].
\end{equation}
\noindent
The main contribution of this attack is the use of \ac{pgd} for the maximisation problem, which suggests the problem is tractable.
In order to explore a large part of the loss landscape, \ac{pgd} is restarted from many points in the $L_{\infty}$ ball around an input.
Surprisingly, although there are many local maxima spread widely apart within $\vx_i + \mathcal{U}$, they tend to have very well concentrated loss values.
This suggests that an adversarial example found by this method is representative for \emph{all} adversarial examples generated with first order methods.
Because of this, the attack was named \emph{complete first order adversary}.

The general aspect of \ac{cfoa} suggests that training against complete first order examples can yield robustness against all possible examples generated with first order methods.
The training results are discussed in Section~\ref{subsubsec:madry_learning}.

%
%


\subsubsection{\ac{sa}}
\label{subsubsec:strong_adversary}

Huang \etal~\cite{huang2015learning} develop two methods to generate adversarial examples and use them to train robust \ac{DNN}.
In this section we present the adversarial generation process and postpone the training procedure for Section~\ref{subsubsec:learning_strong_adversary}.

The search for a minimal perturbation is based on the \emph{linear approximation}\footnote{The linear approximation follows from Taylor's theorem when n=1.} of the \ac{DNN} output. 
If we denote the output of the softmax layer as $y = (\alpha_1, \alpha_2, \dots, \alpha_l)$, the linear approximation for a perturbation $\eta$ can be written as $ \hat{y}(\vx+\eta) = y(\vx) + \mathbf{H} \eta $; where $\mathbf{H}$ is the Jacobian matrix \wrt~$\vx$.
It follows that for a target $l' \neq l$ the necessary condition for $\eta$ is $	\mathbf{H}_{l'} \eta - \mathbf{H}_{l} \eta \leq \alpha_{l} - \alpha_{l'} $. 
Thus the norm for the optimal perturbation can be formally defined as:
\begin{equation}
	\label{eq:strong_adversary}
	\min_{\eta} \| \eta \|_p \quad \text{s.t.} \quad \mathbf{H}_{l'} \eta - \mathbf{H}_{l} \eta \leq \alpha_{l} - \alpha_{l'}.
\end{equation}
\noindent
The final perturbation is computed as $\eta = \epsilon \eta_{i} / \| \eta_{i}\|_p$ where $\eta_{i}$ is the output of \eqq~(\ref{eq:strong_adversary}) for a chosen $p$ norm.

The authors propose a second method to generate a perturbation that follows from using the linear approximation, introduced earlier, when maximising the loss function.
The objective is to find a perturbation smaller than $\epsilon$ that can maximise the loss function: $ \eta = \arg\max_{\|\eta_{i}\|\leq \epsilon} J\left( \vtheta, \vx + \eta_{i}, y_i \right) $.
Using the linear approximation detailed earlier, a perturbation can be formally defined as:
\begin{equation}
	\eta = \{\eta :\, \|\eta\|_{p} \le \epsilon;\, \langle H_{y_i},\,\eta_{i}\rangle = \epsilon \|H_{y_i}\|_p\}.
\end{equation}
\noindent
Depending of the chosen norm, the optimal solution takes different forms.
For example, if we use $L_\infty$, the attacks is identical to \ac{fgsm}.

The solely performance of the attacks is not discussed in the paper.
However, the attacks prove efficient for training a robust neural network (Section~\ref{subsubsec:learning_strong_adversary}).

\subsection{Attacks based on Sensitive Features (SA)}
\label{subsec:sensitivity}
Until now we reviewed attacks that use optimisation methods in order to search for adversarial perturbations in the input space, project an input on a discriminant surface or optimise alternatives of \eqq~(\ref{eq:adversarial_generic}).
In this section we introduce attacks that search for sensitive features of one input and modify them in order to generate adversarial examples.
Although these attacks also use optimisation methods, their objective is to determine sensitive features or directions of perturbations, and later use them to build an adversarial example.

\subsubsection{The \ac{fgsm} Attack}
\label{subsubsec:fgsm}

The \ac{fgsm}~\cite{goodfellow2014explaining} attack is as a consequence of the linearity conjecture, presented in Section~\ref{sec:causes}.
It assumes that summing small perturbations in the direction of the gradient taken \wrt~one input can lead to adversarial examples.
In order to overcome the speed limitations of L-BFGS, Goodfellow \etal~\cite{goodfellow2014explaining} propose a fast, upper-bounded, method to generate an example, formally defined as:
\begin{equation}
	\label{eq:fgsm}
	\veta = \eps \sign \left( \nabla_\vx J(\vtheta, \vx, y) \right),
\end{equation}
\noindent
where $\epsilon$ is a hyper-parameter that controls the size of a perturbation.
The final adversarial example results from $\vx' = \vx + \veta$.

The gradient is a vector of partial derivatives, where each partial derivative gives the local rate of change of the output \wrt~the corresponding input, holding the other inputs fixed.
Analysing the gradient of the loss function \wrt~the input is often referred to as \emph{sensitivity} or \emph{saliency} analysis~\cite{yeung2010sensitivity} and can reveal the importance of one feature in the decision process.
Moreover, the gradient points the direction of the maximum of a function.
By taking a small step in this direction, the \ac{fgsm} attack takes a small step in increasing the loss function \wrt~each input feature.

The value of $\epsilon$ controls the size of a perturbation and impacts the sensitivity of a human observer.
It represents an upper bound for the robustness of a model.
In practice, values as low as $.1$ generate error rate of over $85\%$ on small convolutional max-out networks.
\subsubsection{The \ac{jsma}}
\label{subsubsec:jsma}

Papernot \etal~\cite{papernot2016limitations} introduce an attack based on saliency analysis \cite{yeung2010sensitivity}.
In order to discover the importance of each pixel in the decision process, a \emph{saliency map}
\footnote{In computer vision a saliency map is an image which shows each pixel's unique quality.}
is generated by computing the forward derivative (Jacobian) of the function learned by a \ac{DNN}, $f(\cdot)$.
This method contrasts with gradient based methods, which take the backward gradient of the loss function \wrt~network parameters or the input vector (as in the case of \ac{fgsm}). 
The forward derivative allows to find input features that lead to significant changes in the \ac{DNN} output.

For image classification, the input features are image pixels. 
In order to construct the saliency map, one can analyse the forward derivative \wrt~each pixel and see how it influences the newly selected target.
A formal definition for saliency map is introduced~\cite{papernot2016limitations} as:
\begin{equation}
	\label{eq:saliency-map-increasing-features}
	 S(\vx, l')[i] = \left\lbrace
\begin{array}{c}
0  \mbox{ if }   \frac{\partial f_{l'}(\vx)}{\partial \vx_i}<0  \mbox{ or } \sum_{j\neq l'} \frac{\partial f_{j}(\vx)}{\partial \vx_i}>0,\\
\left(  \frac{\partial f_{l'}(\vx)}{\partial \vx_i}\right)  \left| \sum_{j\neq l'} \frac{\partial f_{j}(\vx)}{\partial \vx_i}\right| \mbox{ otherwise },
\end{array}\right.
\end{equation}
\noindent
where $i$ is an input feature and $l'$ is the desired label. 

The first condition rejects input features with a negative target derivative or an overall positive derivative for other classes.
Similarly, $ \sum_{j\neq l'} \frac{\partial f_{j}(\vx)}{\partial \vx_i}$ needs to be negative to decrease or stay constant when feature $\vx_i$ is increased. 
The product on the second line makes it possible to consider all other forward derivative such that we can compare all $S(\vx,l')[i]$.

In summary, high values of $S(\vx,l')[i]$ correspond to input features that will either increase the target class, decrease other classes significantly, or both. By increasing these features, the adversary eventually misclassifies the sample into the target class.
The algorithm selects, at each iteration, relevant input features using the saliency map and increases or decreases their intensity.
An upper bound parameter limits the overall distortion that can be applied.
 
An important limitation of the \ac{jsma} attack, as noted in~\cite{papernot2016towards}, is its inherent computational cost for big images (\eg~ImageNet~\cite{deng2009imagenet}).

\subsubsection{\ac{rssa}}
\label{subsubsec:radomised_attack}

Sharp curvatures near a data point can mask the true direction of steepest ascent and burden the discovery of adversarial examples with single-shot gradient methods.
In order to escape this phenomenon, Tramer \etal~\cite{tramer2017ensemble} introduce an attack that precedes single-shot attacks with a randomisation step.
\noindent
Formally, for hyper-parameters $\epsilon$ and $\alpha$ (where $\alpha < \epsilon$) the attack can be described as follows:
\begin{equation}
	\begin{aligned}
		\vx' & = \vx_{rand} + (\epsilon - \alpha) \cdot \sign \nabla_{\vx_{rand}} J(\vtheta, \vx_{rand}, y), \\
		\text{where} \quad \vx_{rand} & = \vx + \alpha \cdot \sign(\mathcal{N}(\mathbf{0}^d, \mathbf{I}^d)),
	\end{aligned}
\end{equation}
\noindent
and $ \mathcal{N}(\mathbf{0}^d, \mathbf{I}^d) $ is a normal distribution with mean 0 and variance 1. 
The proposed attack can be seen as a single-step alternative to the complete first order adversary, computed with \ac{pgd} (Section~\ref{subsubsec:madry}).

\ac{rssa} searches for an adversarial example starting every time from a random vicinity of the input data point, thus avoiding gradient masking.
The extra random step yields a stronger attack for all models under evaluation, even those not trained on adversarial examples and suggests that a model's loss function is generally less smooth near the data points.

\subsubsection{\ac{bpda}}
\label{subsubsec:backward_pass}

In order to avoid cases when the gradient can not be well approximated - a phenomenon called gradient obfuscation and presented in Section~\ref{sec:defences} - the authors of \cite{athalye2018obfuscated} introduced a new type of attack, called \ac{bpda}.


To approximate the gradient of a non-differentiable layer of a neural network, $f^{i}(\cdot)$, one can search for a differentiable approximation s.t. $g(x) \approx f^{i}(x)$.
Then, the gradient of $f(x)$ can be approximating by performing the forward pass through the whole network, but on the backward pass $f^{i}(x)$ is replaced with $g(x)$.
As long as the two functions are similar, the slightly inaccurate gradients still prove useful in constructing an adversarial example and avoid gradient masking and obfuscation.

\subsubsection{Elastic-Net Regularisation}
\label{subsubsec:elastic_attack}

Chen \etal~\cite{chen2017ead} explore the use of $L_1$ norm in creating adversarial examples.
They propose an attack based on elastic-net regularisation - a mixture of penalty functions used for high-dimensional feature selection~\cite{zou2005regularization}.
Elastic-net regularisation uses $L_1$ and $L_2$ norms as penalty functions. 
Formally, this regularisation technique is defined as:

\begin{equation}
    \min_{\vz \in \sZ} f(\vz) + \lambda_1 \| \vz \|_1 + \lambda_2 \| \vz \|_2^2,
\end{equation}
\noindent
where $\vz$ is a vector of optimisation variables, $f(\vz)$ is a loss function and $\lambda_1$, $\lambda_2$ are regularisation parameters.
\noindent
In practice, the authors extrapolate from the Carlini attack \eqq~(\ref{eq:adversarial_carlini}) using the elastic-net regularisation:

\begin{equation}
    \begin{aligned}
        \label{eq:ead-attack}
            &\min_{\vx'} & & c f(\vx', t) + \beta \| \vx' - \vx \|_1 + \| \vx' - \vx \|_2^2 
            &s.t. & & \vx' \in [0,1]^n,
        \end{aligned}
\end{equation}
\noindent
where $t$ is the target class.

Experimental results show $L_1$-based adversarial examples crafted with this method can be as succesful as $L_2$ and $L_\infty$ attacks.

\subsubsection{The \ac{bim} Attack}
\label{subsubsec:bim}

Kurakin \etal~\cite{kurakin2016adversarial} are one of the first groups to test adversarial examples in the physical world.
Towards this end, they introduce two new methods to generate adversarial examples based on the \ac{fgsm} attack (Section~\ref{subsubsec:fgsm}).
The first one, \ac{bim}, uses the \ac{fgsm} attack iteratively in order to generate adversarial examples, but limits the amount of perturbation on a pixel basis.
The limit allows a more versatile attack, where the perturbation is smoothened among pixels.

Similar to gradient \emph{clipping}~\cite{pascanu2013difficulty}, the authors define a clipping function which limits the value of a pixel by an upper bound ($255$) and forces the adversarial example to stay in the $L_p$-neighbourhood of the original image $\vx$. 
The clipping function is formally defined as follows. 
Given an adversarial example generated through \ac{fgsm}, $\vx' = \vx + \eps \sign \left( \nabla_\vx J(\vtheta, \vx, y) \right) $, the limit is:
\begin{equation}
	Clip_{\vx, \epsilon} \left\{ \vx' \right\} (x, y, z) = \min \Bigl\{ 255, \vx(x,y,z) + \epsilon, \max \bigl\{ 0, \vx(x,y,z) - \epsilon, \vx'(x,y,z) \bigr\} \Bigr\},
\end{equation}
\noindent
where $\vx(x,y,z)$ is the value of channel $z$ from the image $\vx$, at coordinates $(x, y)$ and $\epsilon$ is an upper bound for the model's robustness.

The \ac{bim} attack extends \ac{fgsm} by applying it iteratively, with a small step size, and clipping the pixel values of the intermediate results on each step.
Formally, the \ac{bim} attack is defined as:
\begin{equation}
	\label{eq:bim}
	\vx'_{0} = \vx, \quad
	\vx'_{N+1} = Clip_{\vx, \epsilon}\Bigl\{ \vx'_{N} + \alpha \sign \bigl( \nabla_\vx J(\vx'_{N}, y_{true})  \bigr) \Bigr\},
\end{equation}
\noindent
where $\alpha$ represents the step size. 
In practice, the authors use $\alpha = 1$ and select the number of iterations using the formula $min(\epsilon+4, 1,25\epsilon)$, determined heuristically.


\subsubsection{The \ac{illcm} Attack}
\label{subsubsec:illcm}

By using the \ac{fgsm} or the \ac{bim} attack one can generate an un-targeted misclassification.
However, for classification problems with a large number of classes, but little difference between them, an un-targeted attack might have insignificant impact.
For example, mistaking one breed of dog for another can have a low impact for the classification task, while mistaking a person for a bird can take an interesting turn.
In order to create more interesting attacks, Kurakin \etal~\cite{kurakin2016adversarial} propose he \ac{illcm} attack.
This iterative method uses the \ac{bim} attack (Section~\ref{subsubsec:bim}) in a targeted fashion.
However, instead of manually choosing the target class, \ac{illcm} prefers the \emph{least likely} class according to the model's prediction for an input:
\begin{equation}
	y_{LL} = \argmin_{y} \bigl\{ p( y | \vx ) \bigr\}.	
\end{equation}
\noindent
Thus \eqq~(\ref{eq:bim}) can be re-written, for the \ac{illcm} attack, as follows:
\begin{equation}
	\vx'_{0} = \vx, \quad
	\vx'_{N+1} = Clip_{\vx, \epsilon}\Bigl\{ \vx'_{N} - \epsilon \sign \bigl( \nabla_\vx J(\vx'_{N}, y_{LL})  \bigr) \Bigr\},
\end{equation}
\noindent
because we want to maximise the likelihood for class $y_{LL}$, which is equivalent to minimising the loss function for this class - take a step in the opposite direction of its gradient.
\subsubsection{Momentum Based Gradient Attacks}
\label{subsubsec:attack_momentum}

Dong \etal~\cite{dong2017boosting} propose a class of momentum-based, iterative, adversarial attacks.
As in the case of gradient descent, momentum can stabilise the update directions and help escape poor local minimum/maxima by accumulating a velocity vector in the gradient direction.
Formally, the iterative \ac{fgsm} with momentum is defined as:

\begin{equation}
    \begin{aligned}
        \vg_{t+1} = \mu \vg_t + \frac{\nabla_\vx' J(\vtheta, \vx, y) }{\| \nabla_\vx' J(\vtheta, \vx, y) \|_{1}} \\
        \vx'_{t+1} = \vx'_{t} + \epsilon \text{sign}(\vg_{t+1})
    \end{aligned}
\end{equation}
\noindent
where $ \mu $ is a decay factor.
When the decay factor is equal to $0$, the method is equivalent to \ac{fgsm} (Section~\ref{subsubsec:fgsm}).

\subsection{Attacks which Exploit Geometric Transformations (GT)}
\label{subsec:geometric_attacks}

The attacks introduced by now are based on special crafted perturbations.
It was not yet explored if such perturbations are common in real-life scenarios \eg~due to scratches on cameras or sensor wear.
In this section, however, we review attacks that explore natural and common perturbations.

At first, we present an attack based on very simple geometric transformations of images, followed by an algorithm that measures the robustness of \ac{DNN} to geometric transformations.
Towards the end we introduce an approach that changes the geometry of a scene, without altering the pixels.

\subsubsection{Adversarial Examples Using Rotations and Translations}
\label{subsubsec:rotation_translation}

Engstrom \etal~\cite{engstrom2017rotation} show that only simple transformations - rotations and translations - are sufficient to fool \ac{DNN}.
These transformations are easy to craft and realistic in real-life scenarios.
The authors introduce three methods to search for the best transformations: at first, they propose a gradient-based method, similar to the attacks in Section~\ref{subsec:sensitivity}, which optimises pixel positions in order to increase the cost function.
Secondly, they propose a grid search method that can be ran in black-box mode and, lastly, they propose to randomly sample $k$ different transformations and keep the one on which the model performs worse.

The drop in accuracy ranges from $34-90\%$, depending on the model and dataset used, even though the models were trained with data augmentation techniques (which include affine transformations).
This result leads us to think that adversarial robustness is a concern which surpasses strict adversarial settings.

\subsubsection{ManiFool}
\label{subsubsec:geometric_robustness}

ManiFool~\cite{kanbak2017geometric} is an algorithm for finding small worst-case geometrical transformations of images.
Moreover, the authors define a quantitative measure of robustness to geometrical transformations based on the geodesic distance between two transformations.

The idea behind ManiFool is simply to iteratively move from an image sample towards the decision boundary of the classifier where the classification decision changes, while staying on the transformation manifold.
Each iteration is composed of two steps: choosing the movement direction and mapping this movement onto the manifold. 
The iterations continue until the algorithm reaches the decision boundary and finds a fooling transformation example~\cite{kanbak2017geometric}.


\subsubsection{Spatially Transformed Adversarial Examples}
\label{subsubsec:spatially_transformed}

Xiao \etal~\cite{xiao2018spatially} propose to change the geometry of the scene, while keeping the original appearance.
Instead of imposing norm constraints on the pixel space, the authors introduce a new regularisation loss on the local geometric distortion, thus producing higher perceptual quality for adversarial examples.
Surprisingly, spatially transformed adversarial examples prove harder to defend with defences such as adversarial training.

\subsection{Attacks based on Generative Models (GM)}
\label{subsec:gan_attacks}

In this section we cover adversarial attacks based on generative models - a class of machine learning algorithms that learn to estimate a probability distribution by looking at samples drawn from it.
The  model is used to produce artificial examples belonging to the same distribution.
For example, one could start with a database with raw pictures of plants and generate more examples in order to design a forest.
The goal is to generate examples that are alike the training samples, but not exactly the same.

In particular, we study two generative models: (1) \ac{vae}~\cite{kingma2013variational, danilo2014stochastich} and (2) \ac{gan}~\cite{goodfellow2014generative}.

\ac{vae} search for an optimal function that returns entries very close to those in the input distribution $P(\vx)$, given a vector of latent variables $\vz$.
Formally, we wish to optimise $\vtheta$ s.t. we can sample $\vz$ from $P(\vz)$ and get a high probability that $f(\vtheta, \vz)$ is close to a sample in the input dataset.
This translates as maximising the probability of each sample $\vx$ in the training dataset, under the entire generative process:
\begin{equation}
	\label{eq:vae_1}
	P(\vx) = \int P(\vx|\vz; \vtheta)P(\vz)d\vz,
\end{equation}
\noindent
where $P(\vx |\vz; \vtheta)$ replaces $f(\vtheta, \vz)$. 
This allows the problem to be formulated in a maximum likelihood fashion.
In order to solve \eqq~(\ref{eq:vae_1}), generative models choose the latent variables $\vz$ s.t. task specific information is captured.
In practice, \ac{vae} assume that samples of $\vz$ can be drawn from a simple distribution, \ie~$P(\vz) = \mathcal{N}(\vz | 0, I)$ and use powerful approximators (\eg~\ac{DNN}) to automatically learn a mapping between the independent, normally-distributed $\vz$ values and the required latent variables~\cite{doersch2016tutorial}.
Afterwards, the latent variables are mapped to $\vx$.


\bigskip
\ac{gan} is a generative technique that simultaneously trains two models: a generator - which captures the data distribution - and a discriminator - which estimates the probability that a sample comes from the training data, rather than from the generator.
In order to train \ac{gan}, both models play a min-max game in the following setting: (1) the discriminator, $D$, trains in order to maximise the probability to assign a correct label to training data or samples coming from the generator and (2) the generator, $G$, trains in order to minimise the output of $D$.

Formally, in order to learn the generator's distribution $P_g$ we define a prior on input noise variables $P_\vz(\vz)$, then represent a mapping to data space as $G(\vz, \vtheta_g)$. 
$G$ is a differentiable function represented by a multi-layer perceptron with parameters $\vtheta_g$.
We also define a second multilayer perceptron $D(\vx, \vtheta_d)$ that outputs a single scalar representing the probability that $\vx$ came from the data rather than $p_g$~\cite{goodfellow2014generative}.
The min-max game with value function $V(G,D)$ can be described as follows:
\begin{equation}
\label{eq:gan}
	\min_G \max_D V(D,G) = V(D, G) = \mathbb{E}_{\vx \sim p_{\text{data}}(\vx)}[\log D(\vx)] + \mathbb{E}_{\vz \sim p_{\vz}(\vz)}[\log (1 - D(G(\vz)))].	
\end{equation}
\noindent
Informally, the generative model can be thought of as analogous to a team of counterfeiters, trying to produce fake currency and use it without detection, while the discriminative model is analogous to the police, trying to detect the counterfeit currency~\cite{goodfellow2014generative}.
The same setting can be applied to generate adversarial examples - the generative model aims to generate examples that cause misclassifications and appear normal to human observers and the discriminant enforces such constraints.

\subsubsection{\ac{atn}}
\label{subsubsec:atn}

Baluja and Fischer~\cite{baluja2017adversarial} train a \ac{DNN} that transforms an input into an adversarial example.
The transformation network is trained to fool a target network or to generate examples transferable to a range of networks.
Although \ac{atn} can be potentially used in black-box scenarios, the authors demonstrate its use in targeted, white-box settings.
Formally, \ac{atn} can be defined as:
\begin{equation}
	g_{f, \vtheta_g}(\vx) = \vx \in X \rightarrow \vx',
\end{equation}
\noindent
where $f$ is the target network.
In order to train $g_{f, \vtheta_g}$, the following optimisation problem is solved:
\begin{equation}
	\arg \min_{\vtheta_g} \sum_{\vx_i \in X} \beta L_{X} (g_{f, \vtheta_g}(\vx_i), \vx_i) + L_y (f(g_{f, \vtheta_g}(\vx_i)), f(\vx_i)),
\end{equation}
\noindent
where $L_{X}$ is a loss function in the input space ($L_p$ norm for images), $L_{y}$ is a custom loss function and $\beta$ is a weight that balances the loss.
The $L_{y}$ function attempts to change just one probability in the softmax layer, corresponding to the target class, while keeping all others the same.

The authors use two approaches to generate adversarial examples with an \ac{atn}: (1) use a residual network~\cite{he2016deep} to generate a perturbation and (2) use auto-encoders with $L_{y}$ regulariser.
In practice, using auto-encoders yields the best results and successfully scales to large datasets.
\ac{atn} are efficient to train, fast to execute, and produces diverse  adversarial examples.
Because the network is pre-trained, the generation of adversarial examples only takes one step, suggesting the use of \ac{atn} in adversarial training.
However, this approach was not investigated.

\subsubsection{\ac{nae}}
\label{subsubsec:gan_natural}

The intuition behind generating natural adversarial examples~\cite{zhao2017adversarial} is to perform the search for adversarial examples in a \emph{deep representation} of the input data, instead of searching in the input data space directly.

For this, a generator $G$ is trained to map random noise vectors $\vz$ to samples $\vx$ from the input distribution $P(X)$ (from noise to input domain).
A second generative model called \emph{inverter}, $I$, is trained to map data instances to corresponding dense representations (from input domain to $G$).
This is equivalent to finding an adversary $\vz'$ in an underlying vector space which defines the distribution $P(\vx)$ and then map it back to $\vx'$ with the help of a generative model.

Both the generator and the inverter are trained with Wasserstein \ac{gan}~\cite{arjovsky2017wgan} - a \ac{gan} model that uses the Wasserstein distance in the objective function.
Using the learned functions, a natural adversarial example can be described as follows:
\begin{equation}
	\vx' = G(\vtheta_g, \vz') \quad\text{where}\quad \vz' = \arg \min_{\tilde{\vz}} \| \tilde{\vz} - I(\vtheta_i, \vx) \| \quad\text{s.t.}\quad f(G(\vtheta_g, \tilde{\vz})) \neq f(x).
\end{equation}
\noindent
Instead of $\vx$, its dense representation $\vz' = I(\vtheta_i, \vx)$ is perturbed.
The generator is used to test whether a perturbation $\tilde{\vz}$ fools the classifier by querying $f$ with $\tilde{\vx}=G(\vtheta, \tilde{\vz})$.
An intermediary step minimises the reconstruction error of $\vx$ and the divergence between sample $\vz$ and the inversion $I(\vtheta_i, G(\vtheta_g, \vz))$ in order to encourage the latent space to be normally distributed.

The authors propose two approaches to search for an adversary: (1) by incrementally increasing the search space in the vicinity of $\vz'$ until an adversary $\vx'$ is found and (2) searches for adversaries in a wide search range, then recursively tighten the upper bound of the search range with denser sampling in bisections.
The examples are tested on both computer vision and natural language processing tasks and prove that natural adversaries can help evaluating the accuracy of black-box classifiers even in absence of labeled training data.

\subsection{Grey and Black-box Attacks}
\label{subsec:black_box}

The fact that adversarial examples generated for a model can transfer to others, regardless of architecture, was first observed by Szegedy \etal~\cite{szegedy2013intriguing}.
However, this property was only later explored by Papernot \etal~\cite{papernot2016transferability} in an attempt to evaluate grey and black-box attacks. 
The authors examine how adversarial examples crafted on one model can transfer between several \ac{ML} techniques such as linear regression, \ac{SVM} or \ac{DNN}.
In such cases, an attacker has partial or full knowledge of the training data and can train a substitute model with it.

Papernot \etal~\cite{papernot2017practical} develop and evaluate \emph{practical} black-box attacks against \ac{DNN}.
In order to construct adversarial examples for a target model without any information available, the authors train a substitute model. 
The training data for the substitute model is generated by an adversary and label by querying the target model.
Afterwards, adversarial examples are crafted using attacks based on sensitivity analysis on the substitute model and later transferred to the target model.
In order to properly evaluate their technique, the authors attack various \ac{DNN} models hosted online by large companies and are able to generate misclassifications of over 80\%.

Liu \etal~\cite{liu2016delving} generate adversarial examples for an ensemble of methods, hypothesising that examples that transfer across several substitute models are more likely to transfer across a large array of black-box models.
Their work is also the first attempt to scale black-box attacks on large datasets.
Experimental results show that, in a targeted fashion, the precision of back-box attacks is quite low \ie~the adversarial examples do not maintain their intended class.
However, whenever adversarial examples are used for un-targeted black-box attacks, the success rate increases significantly.

Chen \etal~\cite{chen2017zoo} present an attack that removes the need to train substitute models.
Their approach, called \ac{zoo} uses a derivative free optimisation model where only the output of the classifier is needed for optimisation.
This method can estimate the gradient across the perturbation's direction taking into consideration the value of the objective function at two neighbour points (corresponding to adding or subtracting a small perturbation).
The experimental results show \ac{zoo} outperforms black box attacks that artificially train a substitute model~\cite{papernot2017practical, papernot2016transferability}.
A similar approach was introduced in~\cite{bhagoji2017exploring}.

Another method that avoids training a substitute models is proposed in~\cite{narodytska2017simple}.
The authors use an iterative search procedure to explore the local neighbourhood of a datapoint and refine an adversarial input.
This information provides an approximation of the gradient of the loss function \wrt~to the input and, thus, provides valuable information about the sensitive pixels.
Experimental results show good accuracy even for very deep networks.

Other publications use genetic algorithms to synthesise adversarial examples~\cite{alzantot2018genattack} or as a black-box estimating technique~\cite{ilyas2018black} in order to generate adversarial examples in limited domains (such as limited queries or information).
Outside image recognition, black-box adversarial examples have been used against mall-ware detection systems~\cite{rosenberg2017generic}


\subsection{Other Attacks}

In this section we explore peculiar attacks - which do not fit any of the classes presented before.
In particular, we present two attacks that use evolutionary algorithms: one which evolves a population of unrecognisable images which still fool \ac{DNN} and one which only modifies one pixel of an image.
The last section (Section~\ref{subsec:real_world}) is reserved to adversarial examples used for different \ac{ML} tasks than object recognition.

\subsubsection{Unrecognisable Images}
\label{subsubsec:unrecognisable}

Although most publications consider adversarial examples in close resemblance to inputs drawn from the training set, the authors of~\cite{nguyen2015deep} show that it is easy to produce images that are completely unrecognisable by human which are able to fool \ac{DNN} with over $90\%$ accuracy.
The authors use an evolutionary algorithm called multi-dimensional archive of phenotypic elites~\cite{cullyrobots} which allows to simultaneously evolve a population that contains individuals which score well on many classes.
Fitness is determined by sending the image to a \ac{DNN}. 
If the image generates a higher prediction score for any class than has been seen before, the newly generated individual becomes the champion for that class.

The algorithm finds images that are completely unrecognisable, but get classified with very high confidence by \ac{DNN}.
It is believed that synthetic images are far from the decision boundary and deep into a classification region, therefore the images generate high confidence scores even though they are far from the natural images in the class.
\subsubsection{The \ac{opa}}
\label{subsec:one_pixel}

The One Pixel attack~\cite{su2017one} is based on the $L_0$ norm, described in Section~\ref{subsec:norm_balls}. 
However, instead of using gradient based optimisation techniques - which require access to the underlying model - the authors use differential evolution~\cite{storn1997differential} - an evolutionary optimisation method that ensures high population diversity.
This technique only requires access to the softmax layer output.

Formally, the perturbation is encoded into an array optimised by differential evolution. 
One candidate solution contains a fixed number of perturbations and each perturbation is a tuple holding five elements: x-y coordinates and RGB values of one pixel. 
At each iteration candidate solutions are produced using the following formula:
\begin{equation}
	\begin{aligned}
		\eta_i(g + 1)   &=  \eta_{r1}(g) + F(\eta_{r2}(g) + \eta_{r3}(g)), \\
		r1 &\neq  r2 \neq r3, 	
	\end{aligned}
\end{equation}
\noindent
where $\eta_i$ is an element of the candidate solution, $r1, r2, r3$ are random numbers, $F$ is the scale parameter set to be 0.5 and $g$ is the current index of generation.

In order to define a fitness function, only the output of the softmax function is required.
For targeted attacks, the fitness function will aim to increase the probability of a target class; while for un-targeted attacks it will aim to decrease the probability of the true class.

While it requires less information about the model under attack, the \ac{opa} attack performs poorly when compared to gradient-based methods.

\subsubsection{Adversarial Examples for Other \ac{ML} Tasks}
\label{subsec:real_world}

The threat to adversarial example was also explored for applications other than object recognition.
Of particular interest is the task of malware detection~\cite{grosse2016adversarial, hu2017generating, laskov2014practical, xu2016automatically, sahs2012machine, kreuk2018adversarial}.
Other tasks explored come from the field of reinforcement learning~\cite{behzadan2017vulnerability, huang2017adversarial, lin2017tactics},
speech recognition~\cite{carlini2016hidden, carlini2018audio} or facial recognition~\cite{sharif2016accessorize}.
However, it is not common to present a accurate threat model or consider the economics of carrying an attack using adversarial examples as opposed to other methods.
Therefore, it is not clear if these threats are really important.

Other publications explore the impact of adversarial examples in the physical world~-~by printing corrupted images~\cite{evtimov2017robust, kurakin2016adversarial} or altering the image acquisition device (\eg~phone camera, digital camera, \etc)~\cite{moosavi2017universal}.

\else

\paragraph{Taxonomy of attacks} At a high level, we identify two classes of algorithms used to craft adversarial examples: (1) attacks which \emph{modify} a real input - \eg~an image - and (2) attacks which \emph{generate} adversarial inputs from noise. 
Following the hypothesis presented in Section~\ref{sec:causes}, of the first class we distinguish between attacks that efficiently search the input space for adversarial examples using \emph{optimisation} techniques (Section~\ref{subsec:optimisation_attacks}) and attacks that exploit \emph{sensitive features} (Section~\ref{subsec:sensitivity}).
In some cases, adversarial examples can be generated only using \emph{geometric transformations} (Section~\ref{subsec:geometric_attacks}).
The second class of methods, however, aims to learn the  adversarial or input data generation distribution and use it in order to convert inputs to adversarial examples or generate adversarial examples from noise (Section~\ref{subsec:gan_attacks}).

One level deeper, we distinguish between attacks that use \emph{iterative} or \emph{single-shot} methods~-~thus illustrating a tradeoff between precision and speed.
Some algorithms approximate the minimum perturbation, at the cost of speed, while others use very fast methods, trading precision.
Fast algorithms are used to enhance the training set with adversarial examples - a procedure called adversarial training and presented in Section~\ref{subsec:proactive} - but are weaker and easier to protect against.
Iterative algorithms require more processing time, but are very strong and hard to protect against.

Further on, when classifying adversarial attacks we have to consider the adversarial goals and capabilities, introduced in Section~\ref{sec:attack_models}.
At first, given the adversarial goals, one could aim to generate a non-targeted, random, misclassification or a targeted misclassification.
This goal often shapes the algorithms used to generate adversarial examples because simpler methods (such as single-shot sensitivity attacks) are fitted for non-targeted attacks, but harder to adapt to targeted attacks.
Similarly, iterative attacks are sometimes harder to implement and require more processing time, therefore they are more qualified for targeted attacks.
Once again, the economics of carrying an attack matters. 
Without clear security requirements, it is impossible to draw a line between the efficiency of any of the methods.

Secondly, given the knowledge available to an attacker, we distinguish between white, grey and black-box attacks.
Most attacks assume full-knowledge of the model under attack and, therefore, fall in the white-box category.
Therefore, we will present grey and black-box attacks in a separate section (Section~\ref{subsec:black_box}), although they sometimes use optimisation methods.
The rather small number of black-box attacks, in contrast with white-box attacks, is another way of judging the security implications of adversarial examples.

We note that most attack traits are orthogonal and the existence of one does not imply the absence of others.
For example, one may generate an iterative, targeted, white-box attack as well as an iterative, not-targeted, grey-box attack.
Selecting a method is, therefore, context dependent and requires a clear threat model.
Unfortunately, most publications do not present any threat model, therefore, we have to accept some overlap in the classification.


Before delving into the details of each attack, we present an overview in Table~\ref{tbl:attacks} where we enumerate all attacks in relation to their features.

In the following sections we introduce, in details, each category attacks.
We try to extensively cover all attacks in the current literature, however, new attacks might be rolled out after writing this paper.
The classification follows the criteria introduced earlier.

\subsection{Attacks based on Optimisation Methods}
\label{subsec:optimisation_attacks}

All the attacks are somehow based on optimisation methods. 
For example, in order to discover sensitive features, one has to perform one step of gradient ascent. 
However, in this section we review the attacks that use optimisation methods in order to search for adversarial examples in the input space or optimise alternatives of \eqq~(\ref{eq:adversarial_generic}).

Szegedy \etal~\cite{szegedy2013intriguing} were the firsts to discover the \ac{DNN} sensitivity to adversarial examples and coin the term adversarial examples.
The authors used limited memory box constrained optimisation (L-BFGS) in order to approximate the minimum perturbation needed to change the label of an input.
The initial experimental results showed that the minimum average distortion drops as low as $0.062$.

Carlini and Wagner~\cite{carlini2017towards} proposed an alternative to \eqq~(\ref{eq:adversarial_generic}) in which the classification function is replaced due to it's non-linear character.
The alternative function is chosen \stt~$f(\vx + \eta) = l'$ if and only if $f(\vx + \eta) \le 0$.
This formulation allows the use of powerful optimisation methods in order to search for very small perturbations and remains, at the  time of writing this paper, one of the state-of-the-art attacks.

Madry \etal~\cite{madry2017towards} proposed to use \ac{pgd}~\cite{boyd2004convex} in order to search for the perturbation in a pre-known set that will maximise the loss function.
The use of \ac{pgd} suggests this problem is tractable and a large part of the loss landscape can be explored through it.
This methods suggests that, if we can find the maximum perturbation from a given set and protect against it, we can protect against the whole set.
Huang \etal~\cite{huang2015learning} also search for a minimum perturbation in a given set, but use the linear approximation of the \ac{DNN} output.

The DeepFool attack~\cite{moosavi2016deepfool} assumes \ac{DNN} behave linear around an input point and projects the point to a separation plane between two classes.
In the case of multi-class classifiers, the separation plane represents the face of a polyhedron whose faces are the all discriminants to other classes.
The attack assumes the surface that separates two classes is defined by an implicit equation \stt~the geometrical normal is equal to its gradient vector.
Finding the normal requires an optimisation procedure equivalent to sensitivity analysis, however, the purpose is not to select sensitive features.
Therefore, we mention the DeepFool attack in this section.

Moosavi-Dezfooli \etal~\cite{moosavi2017universal} use the DeepFool attack~\cite{moosavi2016deepfool} iteratively for all images in the training dataset in order to find an universal perturbation~-~which can fool a large part of the training set.
Several passes over the training dataset are required in order to improve the quality of the perturbation.
At the end of the procedure, however, the method succeeds in finding \emph{image-agnostic} perturbations which cause misclassifications with high confidence.

\if 0\mode \bigskip \else\fi
The attacks based on optimisation methods are, in general, precise in finding a minimum perturbation.
However, they are disadvantaged by processing time. 
Searching for the minimum perturbation is not always desired.
In use-cases where one wants to train \ac{DNN} with adversarial example, the speed of finding adversarial examples is crucial because it directly impacts training time.
This need led to the development of faster algorithms, which exploit different characteristics of \ac{DNN}.

\subsection{Attacks based on Sensitivity Analysis}
\label{subsec:sensitivity}

In order to overcome the speed drawbacks of optimisation-based attacks and following the linearity hypothesis introduced in Section~\ref{sec:causes}, Goodfellow \etal~\cite{goodfellow2014explaining} proposed to sum small perturbations in the direction of the loss gradient taken \wrt~one input. 
Since the gradient gives the loss rate of change \wrt~an input, moving a small step in the direction of the gradient results in taking a step towards maximising the loss function.
Analysing the gradient of the loss function \wrt~the input is often referred to as \emph{sensitivity} or \emph{saliency} analysis~\cite{yeung2010sensitivity} and reveals the importance of one feature in the decision process.

Formally, the perturbation resulting from this simple attack (called \ac{fgsm}) is defined as: $\veta = \eps \sign \left( \nabla_\vx J(\vtheta, \vx, y) \right)$.
The value of $\epsilon$ controls the size of a perturbation and impacts the sensitivity of a human observer.
Because this attack only requires the computation of the gradient vector, it is fast to apply and can be used to quickly generate new training data.
However, this method trades precision for speed.

In order to increase the precision of \ac{fgsm} Kurakin \etal~\cite{kurakin2016aadversarial} propose to apply this method iteratively and, similar to the gradient clipping procedure, limit the the value of a pixel by an upper bound.
Moreover, the authors propose to use this method in a targeted fashion, by maximising the likelihood for a chosen class.

Sharp curvatures near a data point can mask the true direction of steepest ascent and burden the discovery of adversarial examples with single-shot gradient methods.
In order to escape this phenomenon, Tramer \etal~\cite{tramer2017ensemble} introduce an attack that precedes single-shot attacks with a randomisation step.
\ac{rssa}~\cite{tramer2017ensemble} searches for an adversarial example starting every time from a random vicinity of the input data point, thus avoiding any gradient obfuscations.

Dong \etal~\cite{dong2017boosting} proposed to boost the iterative version of \ac{fgsm} using gradient momentum.
As in the case of gradient descent, momentum can stabilise the update directions and help escape poor local minimum/maxima by accumulating a velocity vector in the gradient direction.
Setting the velocity vector to $0$, is equivalent to the normal \ac{fgsm} attack.

Papernot \etal~\cite{papernot2016limitations} introduced an attack based solely on saliency analysis~\cite{yeung2010sensitivity}.
In order to discover the importance of each pixel in the decision process, a \emph{saliency map} is generated by computing the forward derivative of the function learned by \ac{DNN}.
This method contrasts with early methods introduced in this section, which use the backward gradient of the loss function.
The forward derivative allows to find better input feature, which ultimately lead to significant changes in the \ac{DNN} output.
However, its inherent computational costs for big images limits the impact of this method.
Khrulkov and Oseledets~\cite{khrulkov2017art} construct universal adversarial perturbations using singular value vectors of the Jacobian matrix of the feature maps.

For cases when the gradient can not be well approximated~-~a phenomena called gradient obfuscation and presented in Section~\ref{sec:defences}~-~the authors of~\cite{athalye2018obfuscated} introduce an attack which replaces the gradient of a non-differentiable layer with a differentiable approximation. 
Thus, the gradient of a \ac{DNN} can be approximated by performing the forward pass through the whole network, but on the backward pass each layer is replaced by its approximation.
As long as the two functions are similar, the slightly inaccurate gradients prove useful in constructing adversarial examples.

Chen \etal~\cite{chen2017ead} extrapolate the Carlini\&Wagner attack~\cite{carlini2017towards} from elastic-net regularisation~-~a mixture of penalty functions used for high-dimensional feature selection~\cite{zou2005regularization}.
This algorithm is a bridge between optimisation and sensitivity analysis methods because it uses different minimisation techniques (in the form of regularisation) to discover sensitive features and, thus, adversarial examples.

%
%
%

\subsection{Attacks which Exploit Geometric Transformations}
\label{subsec:geometric_attacks}

By now we have introduced algorithms that craft special perturbations~-~which are not common in real-life scenarios.
In this section we review some attacks that use natural and common geometric transformations such as rotation or translation in order to fool \ac{DNN}.

Engstrom \etal~\cite{engstrom2017rotation} show that only simple transformations - rotations and translations - are sufficient to fool \ac{DNN}.
These transformations are easy to craft and realistic in operational scenarios.
The authors propose several methods ranging from randomly sampling different transformations to grid search or gradient approaches.
Depending on the chosen method, the drop in accuracy ranges from $34-90\%$ on models trained with data augmentation techniques (which already include affine transformations).

ManiFool~\cite{kanbak2017geometric} searches for the smallest, worst-case, transformation that can fool \ac{DNN}.
Similar to the perturbations generated by optimisation or sensitivity based methods, these perturbations are imperceptible by human observers.
The main idea behind ManiFool is simply to iteratively move from an image sample towards the decision boundary where the classification decision changes, while staying on the data transformation manifold.

Xiao \etal~\cite{xiao2018spatially} propose to change the geometry of the scene, while keeping the original appearance.
Instead of imposing norm constraints on the pixel space, the authors introduce a new regularisation loss on the local geometric distortion.
The perceptual quality of the adversarial examples remains high, while most defences fail against this attack.


\subsection{Attacks based on Generative Models}
\label{subsec:gan_attacks}

In this section we cover adversarial attacks based on generative models - a class of machine learning algorithms that learn to estimate a probability distribution by looking at samples drawn from it.
The  model is used to produce artificial examples belonging to the same distribution.
For example, one could start with a database with raw pictures of plants and generate more examples in order to design a forest.
The goal is to generate examples that are alike the training samples, but not exactly the same.
In particular, two generative models are heavily used: (1) \ac{vae}~\cite{kingma2013variational, danilo2014stochastich} and (2) \ac{gan}~\cite{goodfellow2014generative}.

Baluja and Fischer~\cite{baluja2017adversarial} trained a \ac{DNN} that transforms an input into an adversarial example.
The transformation network is trained to fool a target network or to generate examples transferable to a range of networks.
The authors use two approaches to generate adversarial examples: (1) use a residual network~\cite{he2016deep} to generate a perturbation and (2) use auto-encoders.
In practice, using auto-encoders yields the best results and successfully scales to large datasets.
This model is efficient to train, fast to execute, and produces diverse  adversarial examples.
Because the network is pre-trained, the generation of adversarial examples only takes one step, suggesting its efficiency in adversarial training.
However, this approach was not investigated.

Zhao, Dua and Singh~\cite{zhao2017adversarial} propose to search for adversarial examples in a \emph{deep representation} of the input data~-~instead of searching directly in the input data space.
For this, a generator $G$ is trained to map random noise vectors $\vz$ to samples $\vx$ from the input distribution $P(X)$ (from noise to input domain).
A second generative model called \emph{inverter}, $I$, is trained to map data instances to corresponding dense representations (from input domain to $G$).
This is equivalent to finding an adversary $\vz'$ in an underlying vector space which defines the distribution $P(\vx)$ and then map it back to $\vx'$ with the help of a generative model.
Both the generator and the inverter are trained with Wasserstein \ac{gan}~\cite{arjovsky2017wgan} - a \ac{gan} model that uses the Wasserstein distance in the objective function.


\subsection{Other types of attacks}
\label{subsec:others}


Although most publications consider adversarial examples in close resemblance to inputs drawn from the training set, the authors of~\cite{nguyen2015deep} 
	craft special images that can not be understood by human observers, but can generate a targeted classification with over $90\%$ accuracy.
The authors leverage evolutionary algorithms~\cite{cullyrobots} in order to evolve candidate solutions that can fool \ac{DNN}.
The fitness function evaluates candidates by sending the image to a target \ac{DNN}.

Instead of using gradient based optimisation techniques in order to change only one pixel of an image, the authors of~\cite{su2017one} use differential evolution~\cite{storn1997differential} - an evolutionary optimisation method that ensures high population diversity.
In order to define a fitness function, only the output of the softmax function is required.
For targeted attacks, the fitness function aims to increase the probability of a target class; while for un-targeted attacks it aims to decrease the probability of the true class.
While it requires less information about the model under attack, this attack performs poorly when compared to gradient-based methods.

\if 0\mode \bigskip \else  \newpara  \fi

\subsection{Attacks using Adversarial Examples on other \ac{ML} Tasks}
\label{subsec:real_world}

The threat to adversarial example was also explored for tasks other than the object recognition task.
Of particular interest is the task of malware detection~\cite{grosse2016adversarial, hu2017generating, laskov2014practical, xu2016automatically, sahs2012machine, kreuk2018adversarial} because it poses high security risks.
Other tasks explored come from the fields of reinforcement learning~\cite{behzadan2017vulnerability, huang2017adversarial, lin2017tactics},
speech recognition~\cite{carlini2016hidden, carlini2018audio} or facial recognition~\cite{sharif2016accessorize}.
However, none of these publications present an accurate threat model or judge the economics of carrying an attack using adversarial examples as opposed to other methods.
Other publications explore the impact of adversarial examples in the physical world - by printing corrupted images~\cite{evtimov2017robust, kurakin2016adversarial} or altering the image acquisition device (\eg~phone camera, digital camera, \etc)~\cite{moosavi2017universal}.

\fi
    \section{Defences}
\label{sec:defences}

\if 0\mode

	\paragraph{Taxonomy of defences.} Following the defender's capabilities, presented in Section~\ref{subsec:defender_goals}, we identify three major classes of defences:
\begin{enumerate}
	\item \emph{Reactive} defences. In this case defenders leverages pre-processing techniques in order to alleviate the impact of adversarial examples, or employ detection mechanisms in order to discard them completely.
	\item \emph{Obfuscation} defences. In this case a defender aims to hide or obfuscate sensitive traits of a model (\eg~gradients) in order to alleviate the impact of adversarial examples. 
	\item \emph{Proactive} defences. In this case defenders aim to build models that are natively robust to small adversarial perturbations.
\end{enumerate}

\if -\mode \paragraph{Overview of reactive defences (Section~\ref{subsec:reactive_defences}).} \else \fi%
In the following sections we will review two classes of \emph{reactive defences}, namely, \emph{detection of adversarial examples} \if 0\mode (Section~\ref{subsubsec:detection} \else\fi and \emph{input transformations} \if 0\mode (Section~\ref{subsec:input_transform}) \else\fi.
Detection techniques assume that adversarial examples have special characteristics or come from a different data generation distribution than normal inputs.
Therefore, it was hypothesised that a detector can learn to distinguish between the two~\cite{song2017pixeldefend, meng2017magnet, ghosh2018resisting, lee2017generative}.
In some cases, the learning procedure is embedded in the main model, by defining a special class corresponding to adversarial inputs.
In others, a detector is trained separately, using specially crafted features.
\if 0\mode 
For example, using the output of hidden layers~\cite{metzen2017detecting}, using statistics on the output of convolutional layers~\cite{li2017adversarial} or measuring the distance between a point and the data manifold, at the output layer of \ac{DNN}~\cite{feinman2017detecting}.
An interesting approach, besides detection of adversarial examples, is the attempt to recover the original samples and correctly classify them.
In spite of the fact that promising results are presented in all papers, most detectors can be by-passed with iterative attacks (Section~\ref{subsec:overall_defences}).
\else \fi

Defences belonging to the \emph{input-transformation} class use pre-processing techniques such as compression or bit-depth reduction~\cite{guo2017countering} in order to remove the effect of adversarial perturbations.
The field sometimes overlaps with adversarial detection methods, where input transformations are used in order to build adversarial detectors.
However, in the former case, the purpose is not to discard an input, but to pre-process it in such a way that adversarial perturbations become useless.
A promising feature of input transformation methods is their speed of adoption and adaptation. 
In most cases, they can be applied at test time and do not require any training time.
However, input transformations are domain specific and hard to transfer between different machine learning tasks.
For example, one has to develop different pre-processing techniques for speech and image recognition.

\if 0\mode \bigskip \else \fi
\if 0\mode \paragraph{Overview of defences by obfuscation.} \else\fi%
A natural answer to attacks that use sensitive features or special traits of models, is to diminish or \emph{obfuscate} such features, in order to restrict the attacker's capabilities.
This phenomenon was first described as \emph{gradient masking} - the use of models that do not have useful gradients~\cite{papernot2017practical}.
However, the range of defences in this category extends from the use of models that do not have useful gradients to modifying the models in order to lower the gradient amplitudes or to remove features that do not contribute to classification.

A peculiar instantiation of the gradient masking phenomenon, that affects the evaluation of many defences, was named \emph{obfuscated gradients}~\cite{athalye2018obfuscated}.
This corresponds to instances where (1) the gradients are non-existent or incorrect, caused intentionally through non-differentiable operations or un-intentionally through numerical instability, (2) the gradients depend on test-time randomness or (3) the model is subject to vanishing or exploding gradients~\cite{pascanu2013difficulty}.
Obfuscated gradients is a phenomenon prevalent across \ac{ML} models and leads to an amplified effect of defences. 
The authors of~\cite{athalye2018obfuscated} develop an offensive technique able to overcome these limitations and exploit the gradient informations even in such scenario.
This result shows that defence by obscurity is not a strong method to employ against adversarial examples.
Defence obfuscation can be achieved through different techniques that heavily overlap with adversarial training or new architectures~-~which also serve other purposes. 
Because of this overlap, we choose to not discuss these techniques in a separate section and only signal the gradient obfuscation effect whenever the case.

\if 0\mode \bigskip \else \fi
\if 0\mode \paragraph{Overview of proactive defences (Section~\ref{subsec:proactive_defences}).} \else\fi%
Defences that fall in the last class~-~\emph{proactive}~-~aim to build models that are natively robust to small perturbations.
An initial attempt to actively increase the \ac{DNN} robustness to adversarial examples, suggested in the initial paper~\cite{szegedy2013intriguing}, is to include adversarial examples in the training set.
This procedure was scalable with the advent of fast methods to generate adversarial examples, which increased the training speed and lead to scalable \emph{adversarial training}\if 0\mode~(Section~\ref{subsec:adv_training})\else\fi.
Within this category we can distinguish \if 0\mode and separate \else between \fi a set of defences that formulate the learning problem as a \emph{min-max} (or saddle) problem and employ optimisation techniques\if 0\mode(Section~\ref{subsec:minmax})\else\fi.
\if 0\mode
The reason behind this choice is that min-max learning uses only perturbed samples during the training process; as opposed to adversarial training where both perturbed and normal inputs are used.
\else 
The distinction can be made because min-max learning uses only perturbed samples, as opposed to adversarial training where both perturbed and normal inputs are used.
\fi

As discussed in Section~\ref{sec:attack_models}, most defences act against small perturbations.
Some techniques try to completely remove the sensitivity to adversarial examples by designing new models and \ac{DNN} architectures.
\if 0\mode 
We gather all defences that alter the learning process in any way - \eg~modify any of the layers or imposing other restrictions than gradient obfuscation~-~into one category called \emph{architectural defences} \if 0\mode (Section~\ref{subsec:defences_arch}) \else\fi.
\else 
We can, thus, separate the defences that alter the learning process by modifying any of the layers or imposing other restrictions than gradient obfuscation under the tag \emph{architectural defences}.
\fi

A normal follow up of the linear hypothesis, introduced in Section~\ref{sec:causes}, is to increase the non-linearity of \ac{DNN}.
Other authors have suggested to increase the capacity of a neural network in order to increase robustness.
Both methods affect training time.
Although, at the beginning, these approaches seem similar to architectural defences, they only change the \emph{hyper-parameters} of a network (number of layers or activation function). 
\if 0\mode Therefore, they are reviewed in a different section~(Section~\ref{subsec:hyperparam}).\else \fi 

An interesting approach to solving the adversarial examples issue is to search for the smallest space at $L_p$ distance from an input where the classifier is robust.
Defences falling in this category aim to certify that a classifier can work under some levels of uncertainty \ie~is robust within some bounds.
\if 0\mode
Therefore, they are reviewed separately, as \emph{provable adversarial defences}, in Section~\ref{subsubsec:provable}.
Towards the end we review a set of defences that use \emph{generative models} (Section~\ref{subsec:defences_gans}).
\else 

\fi
Before delving into each individual class, we provide a complete list of the defences in Table~\ref{tbl:defences}.

	\if 0\mode
\begin{table}
	\centering
	\begin{tabular}{|l|l|l|l|l|}
		\toprule%
		Defence & \specialcell{Type} &%
		  		 \specialcell{Method} &%
		  		 \specialcell{Threat Model} &%
		  		 \specialcell{Defeated} \\ 
		\midrule
		Statistical Detection~\cite{grosse2017statistical} & Reactive & Detection & Yes  & Yes \\ \hline
		Binary Classification~\cite{gong2017adversarial} & Reactive & Detection & No &  Yes  \\ \hline
		In-Layer Detection~\cite{metzen2017detecting} & Reactive & Detection & No & Yes  \\ \hline
		Detecting from Artifacts~\cite{feinman2017detecting} & Reactive & Detection & No & Yes \\ \hline
		SafetyNet~\cite{lu2017safetynet} & Reactive & Detection & No & Yes  \\ \hline
		Saliency Data Detector~\cite{zhang2018detecting} & Reactive & Detection & No & Yes  \\ \hline
		Linear Transformations Detector~\cite{bhagoji2018enhancing} & Reactive & Detection & No & Yes \\ \hline
		Key-based Networks~\cite{zhao2018detecting} & Reactive & Detection & No & Yes  \\ \hline
		Ensemble Detectors~\cite{abbasi2017robustness} & Reactive & Detection & No & Yes  \\ \hline
		Generative  Detector~\cite{lee2017generative} & Reactive & Detection & No & -  \\ \hline
		Convolutional Statistics Detector~\cite{li2017adversarial} & Reactive & Detection & No & Yes  \\ \hline
		Feature Squeezing~\cite{xu2017feature} & Reactive & Detection & Yes & Yes  \\ \hline
		PixelDefend~\cite{song2017pixeldefend}  & Reactive & Detection & No & Yes \\ \hline
		MagNet~\cite{meng2017magnet} & Reactive & Detection & Yes & Yes  \\ \hline
		VAE Detector~\cite{ghosh2018resisting}  & Reactive & Detection & No & Yes  \\ \hline
		Bit-Depth~\cite{guo2017countering} & Reactive & Input Transformation & Yes & Yes  \\ \hline
		Basis Transformations~\cite{shaham2018defending} & Reactive & Input Transformation & Yes &  Yes \\ \hline
		Randomised Transformations~\cite{xie2017mitigating} & Reactive & Input Transformation & Yes & No \\ \hline
		Thermometer Encoding~\cite{buckman2018thermometer} & Reactive & Input Transformation & No & Yes \\ \hline
		Blind Pre-Processing~\cite{rakin2018blind} & Reactive & Input Transformation & No & Yes   \\ \hline
		Data Discretisation~\cite{chen2018improving} & Reactive & Input Transformation & No & Yes  \\ \hline
		Adaptive Noise~\cite{liang2017detecting} & Reactive & Input Transformation & Yes & Yes  \\ \hline
		FGSM Training~\cite{goodfellow2014explaining} & Proactive & Training & No & Yes \\ \hline
		Gradient Training~\cite{sinha2018gradient} & Proactive & Training & No &  Yes \\ \hline
		Gradient Regularisation~\cite{lyu2015unified} & Proactive & Training & No & Yes \\ \hline	
		Structured Regularisation~\cite{roth2018adversarially} & Proactive & Training & Yes & Yes \\ \hline
		Robust Training~\cite{shaham2015understanding} & Proactive & Robust Training & No & Yes \\ \hline
		Strong Adversary Training~\cite{huang2015learning} & Proactive & Robust Training & No & Yes \\ \hline
		CFOA Training~\cite{madry2017towards} & Proactive & Robust Training & Yes & Yes \\ \hline
		Ensemble Training~\cite{tramer2017ensemble} & Proactive & Robust Training & Yes & Yes \\ \hline
		Stochastic Pruning~\cite{dhillon2018stochastic} & Proactive & Robust Training & No &  Yes \\ \hline
		Distillation~\cite{hinton2015distilling} & Proactive & Architecture & No & Yes \\ \hline
		Parseval Networks~\cite{cisse2017parseval} & Proactive & Architecture & No &  - \\ \hline
		Deep Contractive Networks~\cite{gu2014towards} & Proactive & Architecture & No & Yes \\ \hline
		Biological Networks~\cite{nayebi2017biologically}& Proactive & Architecture & No & Yes \\ \hline
		DeepCloak~\cite{gao2017deepcloak} & Proactive & Architecture & No & Yes \\ \hline
		Fortified Networks~\cite{lamb2018fortified} & Proactive & Architecture & Yes & No \\ \hline
		Rotation-Equivariant Networks~\cite{dumont2018robustness} & Proactive & Architecture & No & Yes \\ \hline
		HyperNetworks~\cite{sun2017hypernetworks} & Proactive & Architecture & No & Yes \\ \hline
		Bidirectional Networks~\cite{pontes2018bidirectional} & Proactive & Architecture & No & Yes \\ \hline
		DAM~\cite{krotov2017dense} & Proactive & Architecture & No & - \\ \hline
		Certified Defences~\cite{raghunathan2018certified} & Proactive & Certified & - & - \\ \hline
		Formal Tools~\cite{katz2017reluplex, ehlers2017formal, huang2017safety, ruan2018reachability}  & Proactive & Certified & - & - \\ \hline
		Distributional Robustness~\cite{sinha2018certifying}  & Proactive & Certified &  - & - \\ \hline
		Convex Outer Polytope~\cite{kolter2017provable}  & Proactive & Certified & - & - \\ \hline
		Lischitz Margin~\cite{tsuzuku2018lipschitz}  & Proactive & Certified & - & - \\ \hline
		Defence Gan~\cite{samangouei2018defense}  & Proactive & Generative & Yes & Yes \\ \hline
		FB-GAN~\cite{bao2018featurized}  & Proactive & Genearative & No  & - \\ 
	\bottomrule
	\end{tabular}	
	\caption{Catalog of defences against adversarial examples.}	
	\label{tbl:defences}
\end{table}

\else \fi
\begin{table}
	\centering
	\begin{tabular}{|l|l|l|l|l|}
		\toprule%
		Defence & \specialcell{Type} &%
		  		 \specialcell{Method} &%
		  		 \specialcell{Threat Model} &%
		  		 \specialcell{Defeated} \\ 
		\midrule
		Statistical Detection~\cite{grosse2017statistical} & Reactive & Detection & Yes  & Yes \\ \hline
		Binary Classification~\cite{gong2017adversarial} & Reactive & Detection & No &  Yes  \\ \hline
		In-Layer Detection~\cite{metzen2017detecting} & Reactive & Detection & No & Yes  \\ \hline
		Detecting from Artifacts~\cite{feinman2017detecting} & Reactive & Detection & No & Yes \\ \hline
		SafetyNet~\cite{lu2017safetynet} & Reactive & Detection & No & Yes  \\ \hline
		Saliency Data Detector~\cite{zhang2018detecting} & Reactive & Detection & No & Yes  \\ \hline
		Linear Transformations Detector~\cite{bhagoji2018enhancing} & Reactive & Detection & No & Yes \\ \hline
		Key-based Networks~\cite{zhao2018detecting} & Reactive & Detection & No & Yes  \\ \hline
		Ensemble Detectors~\cite{abbasi2017robustness} & Reactive & Detection & No & Yes  \\ \hline
		Generative  Detector~\cite{lee2017generative} & Reactive & Detection & No & -  \\ \hline
		Convolutional Statistics Detector~\cite{li2017adversarial} & Reactive & Detection & No & Yes  \\ \hline
		Feature Squeezing~\cite{xu2017feature} & Reactive & Detection & Yes & Yes  \\ \hline
		PixelDefend~\cite{song2017pixeldefend}  & Reactive & Detection & No & Yes \\ \hline
		MagNet~\cite{meng2017magnet} & Reactive & Detection & Yes & Yes  \\ \hline
		VAE Detector~\cite{ghosh2018resisting}  & Reactive & Detection & No & Yes  \\ \hline
		Bit-Depth~\cite{guo2017countering} & Reactive & Input Transformation & Yes & Yes  \\ \hline
		Basis Transformations~\cite{shaham2018defending} & Reactive & Input Transformation & Yes &  Yes \\ \hline
		Randomised Transformations~\cite{xie2017mitigating} & Reactive & Input Transformation & Yes & No \\ \hline
		Thermometer Encoding~\cite{buckman2018thermometer} & Reactive & Input Transformation & No & Yes \\ \hline
		Blind Pre-Processing~\cite{rakin2018blind} & Reactive & Input Transformation & No & Yes   \\ \hline
		Data Discretisation~\cite{chen2018improving} & Reactive & Input Transformation & No & Yes  \\ \hline
		Adaptive Noise~\cite{liang2017detecting} & Reactive & Input Transformation & Yes & Yes  \\ \hline
		FGSM Training~\cite{goodfellow2014explaining} & Proactive & Training & No & Yes \\ \hline
		Gradient Training~\cite{sinha2018gradient} & Proactive & Training & No &  Yes \\ \hline
		Gradient Regularisation~\cite{lyu2015unified} & Proactive & Training & No & Yes \\ \hline	
		Structured Regularisation~\cite{roth2018adversarially} & Proactive & Training & Yes & Yes \\ \hline
		Robust Training~\cite{shaham2015understanding} & Proactive & Robust Training & No & Yes \\ \hline
		Strong Adversary Training~\cite{huang2015learning} & Proactive & Robust Training & No & Yes \\ \hline
		CFOA Training~\cite{madry2017towards} & Proactive & Robust Training & Yes & Yes \\ \hline
		DiffAI~\cite{mirman2018differentiable} & Proactive & Robust Training & Yes & - \\ \hline		
		Ensemble Training~\cite{tramer2017ensemble} & Proactive & Robust Training & Yes & Yes \\ \hline
		Stochastic Pruning~\cite{dhillon2018stochastic} & Proactive & Robust Training & No &  Yes \\ \hline
		Distillation~\cite{hinton2015distilling} & Proactive & Architecture & No & Yes \\ \hline
		Parseval Networks~\cite{cisse2017parseval} & Proactive & Architecture & No &  - \\ \hline
		Deep Contractive Networks~\cite{gu2014towards} & Proactive & Architecture & No & Yes \\ \hline
		Biological Networks~\cite{nayebi2017biologically}& Proactive & Architecture & No & Yes \\ \hline
		DeepCloak~\cite{gao2017deepcloak} & Proactive & Architecture & No & Yes \\ \hline
		Fortified Networks~\cite{lamb2018fortified} & Proactive & Architecture & Yes & No \\ \hline
		Rotation-Equivariant Networks~\cite{dumont2018robustness} & Proactive & Architecture & No & Yes \\ \hline
		HyperNetworks~\cite{sun2017hypernetworks} & Proactive & Architecture & No & Yes \\ \hline
		Bidirectional Networks~\cite{pontes2018bidirectional} & Proactive & Architecture & No & Yes \\ \hline
		DAM~\cite{krotov2017dense} & Proactive & Architecture & No & - \\ \hline
		Certified Defences~\cite{raghunathan2018certified} & Proactive & Certified & - & - \\ \hline
		Formal Tools~\cite{katz2017reluplex, ehlers2017formal, huang2017safety, ruan2018reachability}  & Proactive & Certified & - & - \\ \hline
%
	\end{tabular}	
\end{table}

\begin{table}
	\begin{tabular}{|l|l|l|l|l|}
		\toprule%
		Defence & \specialcell{Type} &%
		  		 \specialcell{Method} &%
		  		 \specialcell{Threat Model} &%
		  		 \specialcell{Defeated} \\ 
		\midrule
		Distributional Robustness~\cite{sinha2018certifying}  & Proactive & Certified &  - & - \\ \hline
		Convex Outer Polytope~\cite{kolter2017provable}  & Proactive & Certified & - & - \\ \hline		
		Lischitz Margin~\cite{tsuzuku2018lipschitz}  & Proactive & Certified & - & - \\ \hline
		Defence Gan~\cite{samangouei2018defense}  & Proactive & Generative & Yes & Yes \\ \hline
		FB-GAN~\cite{bao2018featurized}  & Proactive & Genearative & No  & - \\ 
	\bottomrule
	\end{tabular}
	\caption{Catalog of defences against adversarial examples.}	
	\label{tbl:defences}		
\end{table}

	\subsection{Reactive Defences}
\label{subsec:reactive_defences}

Reactive defences do not change the model under attack.
These methods act early in the processing pipeline; before an input reaches a model.
Within this class, we distinguish between (1) defences that \emph{detect} adversarial examples and thus process them differently and (2) defences that apply \emph{input transformations} to all inputs, before they reach the classifier.
Since the latter do not distinguish between adversarial and normal inputs, a first requirement is for classifiers to maintain accuracy on normal, transformed, inputs.

\subsubsection{Detection of adversarial examples}
\label{subsubsec:detection}

The algorithms presented in this section assume that adversarial examples are sampled from a different distribution than benign data.
Thus, a classifier can be trained to distinguish between normal and adversarial inputs.
One can either add a new class, corresponding to adversarial examples, to the main model or train a separate detector; with a different architecture an features.
As we shall see, adversarial detection is strongly correlated with the attack types and the hyper-parameters.
Thus, an universal detector is hard to develop.

While most papers propose to discard adversarial examples, towards the end we introduce some attempts to recover the original input and correctly classify it.
This is an important step for safety critical applications where no input can be discarded.
Final remarks regarding the efficacy of adversarial examples are given in Section~\ref{subsec:overall_defences}

\subsubsection*{Statistical Detection of Adversarial Examples}
\label{subsubsec:statistical_detection}

Gross \etal~\cite{grosse2017statistical} use statistical testing to check if adversarial examples are outside the training distribution.
In particular, they use a statistical test designed for high dimensional data, formalised as the biased estimator of the maximum mean discrepancy~\cite{gretton2012kernel}.
The null hypothesis states that an adversarial and a normal input are drawn from the same distribution.

In order to reject the null hypothesis, the test needs at least $50-100$ examples from each attack and class.
This requirement limits the applicability of statistical testing. 
In order to overcome this limitation, the authors propose to add a new class to the model and train it to detect adversarial examples.
This augmentation mechanism allows the model to classify a sample as benign or adversarial and, in the former case, to return its true label.
Experimental results show the approach is feasible, however, the results are tested on models with very low capacity and a low number of classes.
It is not clear if this approach scales to bigger models or datasets.
Moreover, the procedure requires data about an attack before a detector can be trained.
Therefore, it remains vulnerable to new attacks.
\subsubsection*{Binary Adversarial Classification}
\label{subsubsec:adversarial_twins}

Gong, Wang and Ku~\cite{gong2017adversarial} suggest that training a binary classifier to detect between benign and adversarial examples is possible.
The authors train a separate neural network with adversarial examples developed using \ac{fgsm}, \ac{jsma} and \ac{illcm} (Section~\ref{subsec:sensitivity}).
While achieving almost maximum accuracy for one type of attack, their proposal does not generalise to different attacks or different hyper-parameters.
This suggests two directions: that each attack type generates a unique distribution of samples or their model heavily over-fits.
\subsubsection*{In Layer Detection}
\label{subsubsec:on_detecting_adv}

Metzen \etal~\cite{metzen2017detecting} train a separate neural network to detect adversarial examples.
However, instead of using the perturbed input for training, their approach trains a detector with the output of hidden layers.
In particular, they use a ResNet~\cite{he2016deep} architecture and 'plug-in' the detector at the output of various residual blocks.

The authors study two cases, corresponding to different attack models: (1) a static adversary where the attacker has access to the classification model and (2) a dynamic adversary where the attacker has access to both the classification model and the detector.
While the experimental results prove the feasibility of this approach, there are some peculiar properties which limit its applicability.
At first, the location for the detector varies with an attack type \ie~by plug-ing in the detector at different locations they obtain different results for different attacks.
Some detectors generalise well for a range of attacks, however, no detector can be considered universal.
Secondly, the precision of a detector is correlated with the attack's hyper-parameters.
These limitations suggest several detectors must be trained, accounting for different attacks or hyper-parameters.
\subsubsection*{Detecting Adversarial Examples from Artifacts}
\label{subsubsec:detecting_artifacts}

Feinman \etal~\cite{feinman2017detecting} train a linear adversarial detector using two features: (1) kernel density estimates in the subspace of the last layer and (2) bayesian uncertainty estimates extracted from the drop-out layers.
Both features measure if a point belongs to the probability distribution of a class.

The first one, kernel density estimates, is a distance-based metric that evaluates how far a point is from a manifold corresponding to a class.
The second feature, bayesian uncertainty, makes use of drop-out variance to estimate the uncertainty distribution of each input.

Experimental results show that a simple linear model using these two feature can be trained with good performance.
However, the defence is dependent on drop-out architectures, a requirement which drastically limits its impact.
Newer developments in batch normalisation~\cite{ioffe2015batch} achieve better regularisation results than drop-out.
\subsubsection*{SafetyNet}
\label{subsubsec:safetynet}

SafetyNet~\cite{lu2017safetynet} enforces an attacker to solve a discrete optimisation problem.
For a layer of ReLU, each activation is quantised in order to generate a discrete code, later used to train an RBF-SVM adversarial detector.
This enforces an attacker to solve an optimisation problem in order  to find optimal values for the ReLU thresholds.
The approach is interesting and scales well to models with high capacity and to larger datasets.


\subsubsection*{Detecting Perturbations with Saliency Data}
\label{subsubsec:detection_saliency}

Zhang \etal~\cite{zhang2018detecting} train a binary classifier with benign and saliency data.
Saliency data is used to identify significant pixels, later used to predict adversary examples.
The approach scales well on models with large capacity and on large datasets.
However, the model is tested only against attacks based on gradient methods, which by design exploit salience information.
\subsubsection*{Detection using Linear Transformations}
\label{subsubsec:detector_linear_transf}

Bhagoji \etal~\cite{bhagoji2018enhancing} use PCA~\cite{shlens2014tutorial} transformations to train adversarial detectors.
The intuition behind using PCA is its ability to identify the directions in which the input data has maximum variance.
By projecting the data data along these axes one can reduce the input dimensions, while preserving variance.

The authors train a separate detector using the same dataset as the model under attack, transformed with PCA.
Experimental results prove the feasibility of this approach, however, fail to generalise to a multitude of attacks.
Moreover, the defence is tested for models with low capacity.
\subsubsection*{Detection with Key-based Networks}
\label{subsubsec:key_networks}

Zhao \etal~\cite{zhao2018detecting} propose a new detection mechanism that aims to hide the input label.
This will prevent an attacker from maximising an input given a label and, thus, from creating an adversarial example.
The authors define a ono-to-one encoding scheme from true labels to code vectors.

In order to detect adversarial examples, one can verify if the code vector computed from an input matches the signature of a class with certain precision.
If the output is negative, the input is treated as an adversarial example.
An interesting characteristic of this approach is that knowledge of adversarial examples is not necessary for the detector, thus enhancing generalisation across attacks.
Experimental results show good performance on iterative and adaptive attacks.
However, the approach is tested on networks with low capacity and small datasets.

\subsubsection*{Ensemble Detectors}
\label{subsubsec:ensemble_detection}

Ababsi and Gagne~\cite{abbasi2017robustness} developed an ensemble of detectors based on the confusion matrix of a classifier.
The underlying idea is that adversarial instances originating from a given class tend to fall into a small subset of incorrect classes.
Therefore, developing an ensemble of detectors, which can distinguish between confusion classes, can more easily spot adversarial examples.

While experimental results show an improvement in robustness for simple attacks, it is not clear if this approach will work against adaptive and iterative attacks.
\subsubsection*{Generative Adversarial Trainer}
\label{subsubsec:generative_trainer}

Lee \etal~\cite{lee2017generative} propose a generative training method (Section~\ref{subsec:gan_attacks}) in which two networks are trained alternatively.
The first network generates adversarial examples, while the second tries to correctly classify benign and adversarial examples.
This training procedure outperforms adversarial training with \ac{fgsm}, while maintaining a positive regularisation effect.

\subsubsection*{Detection using Convolution Filter Statistics}
\label{subsubsec:convolution_statistics}

Li and Li~\cite{li2017adversarial} develop an adversarial detector based on features extracted at every layer of a convolutional neural network.
They treat an image as a distribution of pixels that can be used to collect statistics and later used the statistics to train an adversarial detector.
Experimental results show it is not necessary to use statistics coming from all convolutional layers of a \ac{DNN}.
Selecting a number of initial layers is enough for good accuracy.

The selected features are non-differentiable in order to exclude adversaries that use gradient-based attacks.
In particular, the authors collect for each pixel the normalised PCA~\cite{shlens2014tutorial} coefficients, the minimal and maximal values and the 25-th, 50-th and 75-th percentile values of the distribution.
The rationale behind using PCA is that, at every layer, a linear transformation is applied before the non-linear one. 
Therefore, a significant part of the learning process lies within the linear transformation, for which PCA is a standard tool to analyse.
In order to include the features from all layers, the authors define a cascade classifier.

The detector shows good results when evaluated against L-BFGS attacks. 
The authors are also among the firsts to suggest that simple transformations applied to the first layer of a network can reduce the impact of adversarial examples and help recover the original input.
They show that, a simple $3x3$ average filter applied on adversarial examples improves the accuracy with over $73\%$.
\subsubsection*{Feature Squeezing}
\label{subsubsec:feature_squeezing}

Similar to the approach suggested in Section~\ref{subsubsec:convolution_statistics}, Xu and Qi~\cite{xu2017feature} apply small filters to the input image before processing.
This simple technique significantly alleviates the impact of adversarial examples.
In particular, the authors investigate reducing the colour depth and applying a local smoothing (blur) filter.

Using complex input representations can misguide \ac{DNN} into choosing irrelevant features.
Therefore, reducing the input space can restrict the adversarial examples space.
A simple technique for images is to reduce the number of bits used to indicate the colour of a pixel, thus decreasing quality and resolutions.
While human observers are sensitive to these changes, machine learning algorithms are not.
Another technique proposed by Xu and Qi is to reduce the amount of noise in an input.
In the image space, this technique is called spatial smoothing or blur.

The authors develop independent detectors for each feature and compare their output with a normal trained classifier.
Initial experimental results show these cheap techniques do reduce the adversarial space and deserve future investigations.
\subsubsection*{Pixel Defend}
\label{subsubsec:pixeldefend}

Song \etal~\cite{song2017pixeldefend} propose to not only detect adversarial examples, but try to \emph{purify} them and search for the true labels.
They leverage a generative adversarial network (Section~\ref{subsec:gan_attacks}), called PixelCNN~\cite{van2016conditional, salimans2017pixelcnn}, to learn the probability distribution of the training set and use statistical tests to detect if a new input belongs to this probability distribution.
If the input does not belong, it is probably an adversarial example.

However, instead of dis-regarding the input, the authors propose to \emph{purify} it and find its true class by searching for a training sample in the vicinity of the input.
Formally, given the input distribution $p(\vx)$ learned by PixelCNN and a new input $\vx'$, the algorithm searches for a training sample $\vx$ that maximises $p(\vx')$ such that $\vx$ is within the $\epsilon_{\text{defend}}$-ball of $\vx'$:
\begin{equation}
	\begin{aligned}
	\max_{\vx} p(\vx), \\
	\text{s.t.} \quad  \| \vx-\vx' \|_\infty \le \epsilon_{\text{defend}}.
	\end{aligned}
\end{equation}
\noindent
Large values of $\epsilon_{\text{defend}}$ might change the meaning of $\vx'$ while small values might be insufficient to return $\vx'$ to the correct distribution.
In practice, $\epsilon_{\text{defend}}$ is chosen adaptively.

\noindent
Pixel Defend shows good performance when PixelCNN is used to estimate the probability distribution of a training set containing adversarial examples (as in adversarial training).
A similar approach for reconstruction using \ac{gan} was proposed by Santhanam and Grnarova~\cite{santhanam2018defending}.

\subsubsection*{MagNet}
\label{subsubsec:magnet}

MagNet~\cite{meng2017magnet} trains a model to detect how different a test example is from normal examples.
The detector approximates the distance between one example and the data manifold. 
If the data is bigger greater than a threshold, then the detector rejects the input sample.
The approach is model independent and can be applied to any neural network architecture.

When the adversary is close to the data manifold (bellow threshold), MagNet tries to recover it using auto-encoders (Section~\ref{subsec:gan_attacks}).
This approach shows high accuracy even when tested against iterative and adaptive attacks. 
A similar approach that uses sparse encoding and gaussian mixture models is presented in the next section. 
\subsubsection*{Gaussian Mixture Variational Autoencoders}
\label{subsubsec:gaussian_resisting}

The authors of~\cite{ghosh2018resisting} designed a generative model that finds a latent random variable such that the input and its label become conditionally independent given the latent variable.
The latent space is chosen as a mixture of Gaussians, such that each mixture component represents one of the classes in the data.
Inferring the label given the latent encoding is done by computing the contribution of the mixture components.
Adversarial samples are rejected based on thresholding the encoder and decoder outputs.
For example, if the distance between the sample encoding and the encoding of the predicted class in the latent space is bellow a threshold.

Similar to other defences using \ac{vae}, the authors propose a method for reconstruction and correct classification.
The experimental results show an increase on robustness for the COIL-100 dataset.
However, it is not clear if this method can scale to ImageNet dataset and can face adaptive attacks.

\subsubsection{Defences Based on Input Transformations}
\label{subsec:input_transform}


As suggested in Section~\ref{subsubsec:detection}, input transformation can help alleviate the impact of adversarial examples.
However, instead of using them to build detectors, the following publications propose the use of input transformations in a pre-processing step.
Although most publications focus on empirical proofs, some recent works~\cite{gopalakrishnan2018combating} introduced a theoretical view on how input sparsity can attenuate the distortion in the softmax layer.
%
A big disadvantage of input transformation techniques is their context dependence - each transformations is specific to a machine learning task.
Thus a method can not transfer from task to task (\eg~object to speech recognition).

\subsubsection*{Countering Adversarial Examples with Bit-depth Reduction and Compression}
\label{subsubsec:countering_transformations}

The authors of~\cite{guo2017countering} suggest the use of bit-depth reduction, JPEG compression, total variance minimisation and image quilting as a pre-processing step of a convolutional classifier.
The idea of using JPEG compression was also explored in~\cite{dziugaite2016study, das2017keeping, shaham2018defending}. 
Variance minimisation and image quilting prove, in practice, more effective.
This result is emphasised when the models are trained with modified images - a technique similar to adversarial training.
The transformations scale to large datasets, however, they are not strong enough to withstand adaptive attacks.
\subsubsection*{Basis Functions Transformations}
\label{subsubsec:basis_transformations}

Shaham \etal~\cite{shaham2018defending} experiment with different input transformations: low-pass filters, PCA, JPEG compression, low resolution wavelet approximations and soft-thresholding.
However, neither of those techniques scales to large models and large datasets.
In some settings the model performs better without any employed defences.
\subsubsection*{Randomised transformation}
\label{subsubsec:defence_ransomisation}

Xie \etal~\cite{xie2017mitigating} propose to use two randomisation operations: (1) random resizing of input images and (2) random padding with zeros around the input images.
The approach performs and generalise surprisingly well even on large networks and datasets, without training the classifier with modified inputs. 
However, the authors do not discuss accuracy-robustness trade-offs, making it unclear if the approach decreases the performance on clean examples or not.

\subsubsection*{Thermometer One Hot Encoding}
\label{subsubsec:thermometer}

Thermometer one hot encoding~\cite{buckman2018thermometer} breaks the linear extrapolation behaviour of \ac{DNN} by pre-processing the input with an extremely non-linear function.
However, instead of replacing a real number with its counterpart transformation, the authors replace each real number with a binary vector.
Multiplying the input vector with the network's weight enables different input values to use different parameters of the network.

The authors use pixel-wise one-hot encodings and thermometer encodings for discretising the inputs.
Experimental results show that thermometer encoding in combination with adversarial training is a powerful defence capable to withstand \ac{pgd} attacks in both white-box and black-box settings.
\subsubsection*{Blind Pre-Processing}
\label{subsubsec:blind_preprocessing}

Inspired by thermometer encoding~\cite{buckman2018thermometer}, Rakin \etal~\cite{rakin2018blind} propose to process the input data using an ensemble of methods which includes $tanh$ function, batch normalisation, thermometer encoding and one hot encoding.
The authors slightly change the attack model, giving the attacker access to all hyper-parameters except for input transformations.
The experiments, however, are only carried against the \ac{fgsm} and Carlini attacks, while thermometer encoding is tested against \ac{pgd}.
\subsubsection*{Data Specific Discretisation}
\label{subsubsec:data_discretisation}

Chen \etal~\cite{chen2018improving} propose a pre-processing technique that can successfully mask the gradients even for adaptive and iterative attackers.
Their proposal is based on encoding the whole input space using a small set of separable codewords and training a classifier on the encoded information.
The only requirement for finding the codewords is to have large pairwise distances under an appropriate metric.

In practice, the authors use density estimators in order to construct separable codes.
Experimental results demonstrate an increase in performance when tested on large datasets and against adaptive attacks.
However, the technique works only for small perturbations. 
\subsubsection*{Adaptive Noise Reduction}
\label{subsubsec:adaptive_noise_detection}

Lian \etal~\cite{liang2017detecting} treat perturbations as noise and leverage noise reduction techniques to reduce their adversarial effects.
If the effect is downgraded properly, the de-noised adversarial example will be classified as a new class, distinct from the adversarial target.
The authors use two image processing techniques: scalar quantisation and smoothing spatial filter.
In order to improve the generality of this method, an adaptive noise reduction is used in the following way: for images with low entropy, an aggressive noise reduction strategy is adopted, while for images with high entropy light strategies are employed.
Upon receiving a sample to a classifier, the algorithm will first apply adaptive noise reduction in order to create a clean sample.
Later, both samples are sent to the original classifier and, if they get different labels, the initial input is treated as an adversarial example.

The experimental results show good performance of detecting \ac{fgsm} attacks.
However, a big requirement of this attack is that adversarial examples are generated with very small perturbations.
Thus, this approach is limited to use-cases where robustness against small requirements is needed.
Moreover, more testing against iterative attacks is needed.
	\subsection{Proactive Defences}
\label{subsec:proactive_defences}

This class of defences alter the model or its learning procedure in order to increase \ac{ML} robustness.
As opposed to reactive defences, where some degree of robustness is achieved  earlier in the processing pipeline, proactive defences aim to design better architectures or training procedures.
Thus, they eliminate any pre-processing steps.

Within this class of defences we distinguish between defences that use adversarial examples in the training set with (Section~\ref{subsec:adv_training}) or without normal inputs (Section~\ref{subsec:minmax}).
Moreover, we distinguish between defences that re-design the network's architecture (Section~\ref{subsec:defences_arch}) or just increase their non-linearity and capacity (Section~\ref{subsec:hyperparam}).
Lastly, we introduce a class of defences that aims to certify a degree of robustness (Section~\ref{subsubsec:provable}) and a class of defences that use generative models (Section~\ref{subsec:defences_gans}).

\subsubsection{Adversarial Training}
\label{subsec:adv_training}

Training with adversarial examples is a form of regularisation~\cite{girosi1995regularization} \ie~a strategy designed to reduce the test error, possibly affecting the training error. 
A common way to make a \ac{ML} algorithm better generalise on a task is to train it with more data. 
In practice, however, the amount of available data is limited. 
In order to overcome this barrier, one can create fake data (by applying noise, translations, rotations, \etc) and \emph{augment} the training set with the new examples.

Unlike common data augmentation schemes, generating adversarial examples is different because these inputs are not expected to occur naturally.
Adversarial examples expose flaws in the ways a model conceptualises its decision functions~\cite{goodfellow2014explaining} and adversarial training helps in overcoming them.

\subsubsection*{Adversarial Training with \emph{FGSM}}
\label{subsubsec:fgsm_training}

Many regularisation schemes limit the capacity of a model by adding a penalty term to the cost function. 
Goodfellow \etal~found that training with an adversarial error function based on the fast gradient sign method (Section~\ref{subsubsec:fgsm}) was an effective regulariser~\cite{goodfellow2014explaining}.
Formally, the adversarial error function is defined as:
\begin{equation}
	 \tilde{J}(\vtheta, \vx, y) = \alpha J(\vtheta, \vx, y) + (1-\alpha) J(\vtheta, \vx + \eps \sign \left( \nabla_\vx J(\vtheta, \vx, y) \right),	
\end{equation}
\noindent
where $\alpha$ is a hyper-parameter that weights the relative contribution of the adversarial penalty.
In practice, Goodfellow \etal~used a value equal to $\alpha = 0.5$.


An important take-away from experimenting with \ac{fgsm} adversarial training is that it is generally better to perturb the input, rather than a hidden layer of a neural network.
When applied to hidden units whose activations are unbounded, \ac{DNN} respond by making their hidden unit activations very large.

The results show that adversarial training is able to significantly improve the robustness of a model.
Moreover, adversarial training helps to better generalise and improve the original accuracy of a model.

\subsubsection*{Gradient Adversarial Training}
\label{subsubsec:gradient_adversarial}

Sinha \etal~\cite{sinha2018gradient} propose a new training framework which assumes that simultaneous gradient updates should be statistically indistinguishable from each other.
Thus the gradient can be regularised in order to remove salient information that can lead to adversarial examples.
Further on, the gradient tensor is processed in an auxiliary network and then passed to the main network via a gradient reversal procedure.
This training procedure adapts the cross-entropy loss function and adds weight to negative classes whose gradient tensors are similar to those of the primary class.
The weight weight is evaluated using the auxiliary network which indicates the gradient tensors similarities between the primary and negative classes.

Experimental results show better regularisation during training and an increase in robustness when tested against gradient adversarial attacks.
However, robustness is not always preserved when testing is done against iterative gradient methods (Section~\ref{subsubsec:illcm}).

\subsubsection*{Training with Gradient Regularisation}
\label{subsec:gradient_regularisation}

Lyu, Huang and Liang~\cite{lyu2015unified} propose a family of gradient based perturbations that can be used as a regularisation technique.
Their work can be seen as a generalisation of the \ac{fgsm} for all possible norms.
The regularisation follows from the Taylor expansion of the min-max problem (maximising the perturbation while minimising the loss - Section~\ref{subsec:minmax}) and yields:

 \begin{equation}
     \min_{\vtheta} J(\vtheta, \vx, y) + \epsilon \| \nabla_{x} J(\vtheta, \vx, y) \|_p.
 \end{equation}
\noindent
In the case of $p_{\infty}$ the perturbation is equivalent to \ac{fgsm} - Section~\ref{subsubsec:fgsm}.
However, the authors propose the use of $p_{2}$, which resembles marginalised Gaussian noise and propose an heuristic way to optimise the objective.
Using this regularisation technique increases performance against single step attacks, on small datasets.
However, the impact of iterative attacks is not empirically evaluated.


\subsubsection*{Training with Structured Gradient Regularisation}
\label{subsubsec:structured_regularisation}

Roth \etal~\cite{roth2018adversarially} propose to use a generative model (Section~\ref{subsec:gan_attacks}) in order to learn the distribution of adversarial corruptions from examples and use it as for regularisation.
The authors propose to define an objective function that includes a mixing distribution between un-corrupted and corrupted data.
Formally, the loss function can be defined as:
\begin{equation}
    L(\vtheta) = (1-\lambda) \mathbb{E}_p{\hat{P}} [L(\vtheta, \vx, y)] + \lambda \mathbb{E}_p{\hat{P}_Q} [L(\vtheta, \vx, y)],
\end{equation}
\noindent
where $\hat{P}$ denotes the empirical distribution of the input and $\hat{P}_Q$ the noise corrupted distribution.



\subsubsection{Learning in a Min-Max Setting}
\label{subsec:minmax}
Another way of training with adversarial examples is to exclude the clean examples from the training procedure.
This is equivalent to training on the 'worst' inputs possible, generated by adversarial attack methods.
The procedure draws from the field of \emph{robust optimisation}~\cite{ben2009robust}; an area of optimisation theory that aims to obtain solutions stable under some level of uncertainty.

Robust optimisation problems have a min-max formulation in which the objective function is being minimised with respect to maximum worst-case perturbations.
The assumption is that the perturbations can be drawn from specific sets called \emph{uncertainty sets} $\mathcal{U}$.
There is a number of cases that one can consider for the uncertainty sets.
One example is the $\mathcal{U}_i = B_p(\vx_i, r)$ norm ball centred at $\vx_i$, with radius $r$ and norm $p$. 
Searching for a maximum perturbation, however, increases the training time considerably and might not always be a tractable problem.

In this section we introduce several attempts to use robust optimisation for training \ac{DNN}.
This procedure is also called \emph{strong} learning because we aim to find the worst adversarial example for each input in the training data set; as opposed to \emph{weak} learning which searches for an adversarial example within some bounds.

\subsubsection*{Robust Adversarial Training}
\label{subsubsec_understanding_adv}

Shaham, Yamada and Negahban~\cite{shaham2015understanding} propose a first training framework based on robust optimisation.
Formally, the minimisation-maximisation approach is stated as:
\begin{equation}
\label{eq:adv_robust_training}
\min_\vtheta \tilde{J}(\vtheta,\vx,y) = \min_\vtheta \sum_{i=1}^m \max_{\tilde{\vx}_i \in \mathcal{U}_i} J(\vtheta, \tilde{\vx}_i,y_i),
\end{equation}
\noindent
where $\mathcal{U}_i$ is the uncertainty set corresponding to input $\vx_i$.
The inner maximisation problem seems intractable for the authors (later, Madry \etal~\cite{madry2017towards} show this is in fact tractable). 
Therefore, they propose to minimise a surrogate of \eqq~(\ref{eq:adv_robust_training}) in which each sub-procedure is reduced to a single ascent/descent step.
The formal expression for finding the worst-case perturbation for an input $\vx_i$ is based on the first-order Taylor expansion of the loss around the example, which yields:
\begin{equation}
  \label{eq:adv_robust_max}
  \eta_{i} = \arg \max_{\eta: \vx_i + \eta \in \mathcal{U}_i} J(\vtheta, \vx_i + \eta, y_i).
\end{equation}
\noindent
The outer minimisation problem is solved using a single descent step with respect to the perturbed data $\vx_i + \eta_i$.
The experimental results show that robust optimisation works best when using the $L_\infty$ norm and helps improving robustness of \ac{DNN} against adversarial examples on both MNIST an CIFAR-10 datasets. 

\subsubsection*{Learning with a Strong Adversary}
\label{subsubsec:learning_strong_adversary}

Huang \etal~\cite{huang2015learning} also formulate the learning procedure as a min-max game.
However, the maximisation problem is solved using the approaches described in Section~\ref{subsubsec:strong_adversary} for upper-bounded uncertainty sets.
Formally, the objective function is stated as:
\begin{equation}
\label{eq:LWA}
	\min_\vtheta \sum_i \max_{\|\eta_{i}\|\leq \epsilon} J(\vtheta,  x_i + \eta_i, y_i).
\end{equation}
\noindent
Another interesting approach proposed in~\cite{huang2015learning} is to perturb only the representation learned by a neural network.
One can regard \ac{DNN} as consisting of two parts: (1) lower layers that can learn a representation of the input data (e.g. convolution layers) and (2) higher layers that learn a classification model on top of the representation (fully connected and softmax layers).
Instead of perturbing the raw input data, the authors propose to use the representation learned by the lower layers in order to train a robust neural network.
Formally, the proposed objective function is:
\begin{equation}
\label{eq:LWA2}
\min_{f{rep}, f{ cla}} \sum_i \max_{\|\eta_{i}\|\leq \epsilon} J \left(f_{cla}\left(f_{ rep}(\vx_i) + \eta_{i}\right), y_i\right),
\end{equation}
\noindent
where $f_{rep}$ are the lower layers in a neural network and $f_{cla}$ are the higher ones.
The perturbation is computed using one of the methods described in Section~\ref{subsubsec:strong_adversary} and
the experimental results show that training using the first method from Section~\ref{subsubsec:strong_adversary} achieves better results than adversarial training with \ac{fgsm} - Section~\ref{subsubsec:fgsm_training}.
\subsubsection*{Learning with a Complete First Order Adversary}
\label{subsubsec:madry_learning}

Madry \etal~\cite{madry2017towards} solve the inner maximisation problem using projected gradient descent (pgd).
The initial results presented in Section~\ref{subsubsec:madry} suggest that training a network with \ac{cfoa} will assure robustness against all first order methods.
Formally, the training procedure is defined as a min-max game:
\begin{equation}
	\label{eq:mandry}
	\min_{\vtheta} \rho(\vtheta), \quad \text{where} \quad \rho(\vtheta) = \mathbb{E}_{(\vx, y) \sim p_{\text{data}}} \left[\max_{\eta \in \mathcal{S}} J(\vtheta, \vx + \eta, y) \right],
\end{equation}
\noindent
The inner maximisation problem is tackled through projected gradient descent starting from a random perturbation around the natural example while the minimisation problem is tackled using stochastic gradient descent.
The results show state of the art performance at the moment of writing this paper.


\subsubsection*{\ac{eat}}
\label{subsubsec:ensemble}

\ac{eat}~\cite{tramer2017ensemble} is a technique that augments training data with adversarial examples crafted on other static, pre-trained, models.
The work is inspired by domain adaptation\footnote{In domain adaptation a model is trained on one or more input distributions and is evaluated on samples from a different, but related, distribution.}~\cite{zhang2012generalization} and tries to decouple the adversarial generation process from the parameters of a trained model.

Training with ensemble methods increases the diversity of perturbations seen during training.
This is equivalent to enhancing the uncertainty sets, $\mathcal{U}$, with examples crafted on other models.
Because adversarial examples are transferable between models, the intuition is that adversarial examples crafted for other models should help in cross-training robust models too.

Ensemble training uses the same procedure as learning with a complete first order adversary \eqq~(\ref{eq:mandry}), but applies a variety of attacks on pre-trained models in order to approximate the inner maximisation problem (e.g. \ac{fgsm}, \ac{rssa}). 
In a white-box setting, models trained through \ac{eat} are slightly less accurate when compared with standard adversarial training. 
However, \ac{eat} significantly boosts robustness to black-box attacks transferred from pre-trained models. 
This seems to be the expected behaviour, because the transfer models are similar to the tested ones (e.g. inceptionV3 vs InceptionV4 or IncRes v2 vs ResNet v2)


\subsubsection*{\ac{sap}}
\label{subsubsec:stochastic_pruning}

Dhillon \etal~\cite{dhillon2018stochastic} define a minimax zero-sum game between an adversary and \ac{DNN}.
The strategy for this game is to prune a random subset of activations (\eg~those with smaller magnitude) and scale-up the survivors in order to compensate.
This approach is similar to the dropout technique, where the activations with high absolute values have a higher chance of being sampled.

\ac{sap} can be applied to pre-trained networks and presents small improvements in accuracy when combined with adversarial training.


\subsubsection{Robust Deep Architectures}
\label{subsec:defences_arch}

This section introduces publications that alter the training method or the network's architecture in order to increase robustness to adversarial examples.
Some works draw inspiration from knowledge transfer, while others take inspiration from regularisations specific to de-noising auto-encoders - a technique used to recover original inputs from recover partially corrupted data.

\subsubsection*{Distillation Defence}
\label{subsubsection:distillation}

Distillation is a training procedure initially designed to train a \ac{DNN} using knowledge transferred from a different \ac{DNN} \cite{hinton2015distilling}.
The motivation behind distillation is to reduce the computational complexity of \ac{DNN} architectures by transferring knowledge from larger architectures to smaller ones.
A distilled network can be trained with \emph{soft} labels \ie~a probability distribution over all classes given by a softmax function, instead of \emph{hard} labels \ie~a discrete value.
The soft labels are the output of a large \ac{DNN} and used as knowledge transfer for smaller networks. 
An important parameter is the \emph{temperature} of the softmax function because a higher temperature produces a smoother probability distribution over classes, thus influencing the knowledge transfer.

Papernot \etal~\cite{papernot2016distillation} define a new distillation mechanism that provides defence training \ie~instead of transferring knowledge between architectures, the knowledge extracted from a \ac{DNN} is used to improve its own resilience to adversarial examples.
This means that, as opposed to the original distillation mechanism, the same network architecture is used for training and distillation.

The knowledge extracted from defensive distillation is used to reduce the amplitude of network gradients.
In general, if the gradients are high, crafting adversarial examples becomes easier because small perturbations will induce high output variations.
The authors show that using distillation with a high temperature reduces the model's sensitivity to small variations of its inputs and increases the overall robustness.
However, Carlini \etal~\cite{carlini2017towards} introduced a series of attacks that break defence distillation. They are discussed in Section~\ref{subsubsec:carlini}. 



\subsubsection*{Parseval Networks}
\label{subsubsection:parseval}

Parseval networks~\cite{cisse2017parseval} are a regularisation scheme that works by constraining the Lipschitz constant of each hidden layer of a \ac{DNN} to be smaller than a threshold.
Through this constraint any exponential growth of the Lipschitz constant is avoided.
In this setting a regularisation scheme (such as weight decay) at the last layer controls the overall Lipschitz constant of the network.

Szegedy \etal~\cite{szegedy2013intriguing} first suggested the idea that robustness of  \ac{DNN} can be proved inspecting the Lipschitz constant at each layer.
As mentioned in Section~\ref{sec:robustness_eval}, Weng \etal~\cite{weng2018evaluating} showed this bound might be loose.
In contrast, the experimental results in ~\cite{cisse2017parseval} show that a network's sensitivity to adversarial examples can be controlled using the Lipschitz constant.

\subsubsection*{Deep Contractive Networks}
\label{subsubsec:deep_contractive}


Gu and Rigazio~\cite{gu2014towards} propose a new network architecture, called deep contractive networks, which imposes a layer-wise penalty in a neural network.
This penalty minimises the network output variance \wrt~perturbations in the input s.t. a trained model can achieve robustness for perturbations around training data points.

The authors use contractive auto-encoders - a variation of auto-encoders with an additional penalty for minimising the squared norm of the Jacobian of the hidden representation with respect to input data~\cite{rifai2011higher}.
When applied to feed-forward networks, this penalty helps to explicitly learn invariant features at each layer.
Formally, the network optimises the following loss function:
\begin{equation}
	J_{DCN}(\vtheta)  = \sum_{i=1}^{m}  (L( \vx, \vtheta, y) + 
	\sum_{j=1}^{h+1} \lambda_j \| \frac{\partial y^{(i)}}{\partial(\vx)^{(i)}} \|_2 ),	
\end{equation}
\noindent
where h is a hidden layer and $\lambda$ is a scaling factor.

Experimental results show deep contractive networks help increase the average perturbation size needed to build an adversarial example.
However, the architecture trades accuracy on clean examples for these results.

\subsubsection*{Biologically Inspired Robust Neural Networks}
\label{subsubsec:biologically}

%

Nayebi and Ganguli~\cite{nayebi2017biologically} develop a new training scheme inspired by biophysical principles in neural circuits.
Following the idea suggested by Goodfellow \etal~\cite{goodfellow2014explaining} that adversarial examples are due to the linear summing of high dimensional input with small weights, the authors propose to force neural networks to operate in a non-linear, saturated, regime.

In order to enforce this constraint, the authors \emph{saturate} the network \ie~ensure that each element of the Jacobian matrix of the model is sufficiently small s.t. the model becomes insensitive to perturbations.
This approach is similar to deep contractive networks - Section~\ref{subsubsec:deep_contractive}.
However, the contractive penalty degrades test accuracy on clean examples and is difficult to compute for large networks.

In order to overcome these limitations, the authors propose the use of penalties specific to saturating auto-encoders~\cite{goroshin2013saturating} in order to explicitly encourage activations to be in the saturating regime of the nonlinearity.
Formally, for a given activation $h = W\vx + b$ and $\lambda \in \sR$, the saturation penalty is:
\begin{equation}
	\lambda \sum_{i=1}^{h} \phi_c(h_i),
\end{equation}
\noindent
where the complementary function is defined as:
\begin{equation}
	\phi_c{\vz} = \inf_{\vz' \in \sS} \| \vz - \vz' \|, \sS = \{\vz | \phi'(\vz) = 0\},
\end{equation}
\noindent
and reflects the distance of any individual activation to be the nearest saturation region.
\noindent
The experimental results show improvements on both first-order methods and adaptive, iterative, attacks.

\subsubsection*{DeepCloak Defence}
\label{subsubsec:deep_cloak}

DeepCloak Defence~\cite{gao2017deepcloak} propose to remove features not use in classification in order to increase robustness to adversarial examples.
To identify unnecessary features a pairs of adversarial samples are tested against the clean example.
In order to remove the features, a mask layer is introduced before the logit layer. 
The mask serves as a selector, keeping the necessary features and setting the unnecessary to 0.

DeepCloak is model independent and easy to implement.
However, the robustness increase is not substantial.
And, as in most architectures, the defence trades accuracy from the original model.

\subsubsection*{Fortified Networks}
\label{subsubsec:fortified}

Lamb \etal~\cite{lamb2018fortified} identify which hidden states are off the data manifold and map these states back to parts of the data manifold where the network performs better.
The fortification consists of inserting de-noising auto-encoders at crucial points between layers of the original network in order to clean up the transformed data points which may lie outside the data manifold.

The experimental results show good results even against the \ac{phd} attack (Section~\ref{subsubsec:madry}).
Similar techniques have been suggested in~\cite{xi2018manifold}.
However, both methods prove inefficient when faced with adaptive attacks~\cite{carlini2017magnet}.
\subsubsection*{Rotation-Equivariant Networks}
\label{subsubsec:rotation_equivariant}

Dumont, Maggio and Montalvo~\cite{dumont2018robustness} investigate the resistance to adversarial attacks of three rotation-equivariant network architectures~\cite{cohen2016group, worrall2017harmonic, zhou2017oriented}.
They discover that rotation equivariant networks are significantly more robust to attacks based on small translations and rotations, and marginally more robust to attacks based on local geometric distortions.
\subsubsection*{HyperNetworks}
\label{subsubsec:hypernetworks}

Sun \etal~\cite{sun2017hypernetworks} propose to use data dependent weights in the hidden layers of a neural network. 
The main idea is to adaptively filter convolution weights using HyperNetworks~\cite{ha2016hypernetworks} - a technqiue that uses a network to generate the weights for another network.

The experimental results show an increase in robustness, without trading off accuracy of the initial model. 
However, although HyperNetworks proved to be powerful, this architecture is not used in large scale vision tasks due to the high dimension of weights in the recent state-of-the-art convolutional neural network architecture.
\subsubsection*{Bidirectional Learning}
\label{subsubsec:bidirectional_learning}

The authors of \cite{pontes2018bidirectional} propose to use bidirectional learning in order to increase robustness to adversarial examples.
Bidirectional learning trains two models - the first on the input and the second on a reversed copy of the input.
However, in \cite{pontes2018bidirectional}, a single model is trained to behave both as a discriminative and generative model.
Therefore, the same model can be a classifier and a generator in the same time.
This behaviour is achieved using an undirected neural network that back-propagates the errors in both directions - each direction of the network has its own biases and the weights are shared.

The positive weights of the last layer of a generator become the first layers of a classifier.
The authors also introduce a hybrid method in which two models are trained (instead of an undirected network) and the weights are shared.
The experimental results show an increase in robustness, due to more robust weights.
Moreover, adversarial examples are harder to generate for bi-directional networks.
However, the results fail to scale to larger datasets.

\subsubsection*{\ac{dam} Models}
\label{subsubsec:dense_memory}

\ac{dam} models~\cite{krotov2017dense} store a set of memory vectors, corresponding to the learnt patterns.
At query time, the network is presented with an incomplete pattern resembling, but not identical to, one of the stored memories and the task is to recover the full memory.
For example, pixel intensities can be combined with the label of the image into one vector, which will serve as a memory for the associative memory.
One limitation of using associative memory models is the limited memory; the standard model of associative memory works fine in the limit when then umber of stored patterns is much smaller than the number of neurones, or equivalent the number of pixels in an image~\cite{krotov2016dense}.

Krotov and Hopfield suggest that \ac{dam} with higher order energy functions are closer to human visual perception that \ac{DNN} with RELUs.
Thus, \ac{dam} models are more robust to adversarial examples.
However, the statements are not empirically supported.

\subsubsection{Hyper-parameter Tuning}
\label{subsec:hyperparam}
\subsubsection*{Non-linearity}
\label{subsec:non_linearity}

Given the linear conjecture, introduced in Section~\ref{sec:causes}, which states that adversarial examples exploit the linear behaviour of \ac{DNN}, a normal attempt to increase robustness is to use more non-linear activation functions. 

Goodfellow \etal~\cite{goodfellow2014explaining} suggest that RBF networks - which behave in a non-linear fashion - are naturally immune to adversarial examples and have low confidence when they are fooled.
However, RBF units can not generalise very well and do not achieve the same performance as ReLU \ac{DNN}.
Krotov and Hopfield~\cite{krotov2016dense} tried to replace ReLU activations with higher rectified polynomials and observed an increase in robustness.
However, the increase is not sufficient to completely remove the adversarial phenomenon.
Moreover, the training procedure is slower.
Some architectural designs also exploit non-linearities and are presented in~\cite{nayebi2017biologically, krotov2016dense} - Section~\ref{subsubsec:biologically} and~\ref{subsubsec:dense_memory}.




\subsubsection*{Increased Capacity}
\label{subsec:capacity}

Several publications suggested that a high capacity of \ac{DNN} increases their robustness to adversarial examples\cite{madry2017towards, kurakin2016aadversarial, rozsa2016accuracy}.
However, no publication investigated the correlation in isolated experiments.
Increasing the capacity comes with additional costs in terms of processing power and training time.
It is interesting to follow if capacity and robustness are, indeed, correlated and which are the trade-offs between the two.
\subsubsection{Provable defences}
\label{subsubsec:provable}

In this section we introduce works that \emph{guarantee} a threshold for the size of a perturbation.
Following the discussion in Section~\ref{sec:robustness_eval}, the thresholds are chosen to guarantee either lower-bound or upper-bound robustness to adversarial examples.



\subsubsection*{Certified Defenses}
\label{subsubsec:certified}

The authors of~\cite{raghunathan2018certified} focus on computing and upper bound on the worst-case loss of linear classifiers and \ac{DNN} with two hidden layer.
This upper bound serves as a \emph{certificate of robustness} against all attacks for a given network and input.

They use integration to obtain an exact expression for upper bound loss using all gradients in a p-ball around an example and use a semidefinite relaxation in order to convert this to a convex, tractable, optimisation problem.
The loss is later used in training a model.

There are, however, some disadvantages with this method. 
At first, it is limited to linear models, therefore, has limited impact in the image recognition task.
Secondly, increasing the number of layers (from 2) also increases the complexity of the method and might result in intractable optimisation problems.


\subsubsection*{Formal Tools for DNN Robustness}
\label{subsubsec:reluplex}

Katz \etal~\cite{katz2017reluplex} propose the use of SMT solvers to prove upper bound robustness of \ac{DNN}.
This procedure encodes the neural network and the constraints regarding upper bound robustness (max $\epsilon$) as a set of linear equations.
The authors extend the simplex algorithm to handle ReLU activation functions - therefore the name Reluplex.
Reluplex uses SAT solving techniques to check if an adversarial example exists in the given constraint space.
If the constraint is not satisfied, the algorithm returns a counter-example which constitutes a valid adversarial example.
Some recent work extends Reluplex to \cite{carlini2018provably} to verify $L_1$ and $L_\infty$ norm, by encoding absolute values using ReLUs.

On the same path, Huang \etal~\cite{huang2017safety} use model checkers in order to guarantee robustness against \emph{known} adversarial perturbations,
Ehlers~\cite{ehlers2017formal} uses SMT solvers for neural networks with linear activation

An inherent disadvantage of using SMT solvers is processing time. 
However, as opposed to other provable defences, this approach is not constrained by network capacity.

In \cite{ruan2018reachability}, the authors design  and  implement  a  reachability  analysis tool  for  deep  neural  networks,  which  has  provable  guarantees and can be applied to \ac{DNN} with deep layers and nonlinear activation functions.  
Their work has the ability to work with larger networks lower  computational  complexity,  i.e.,  NP- completeness  with  respect  to  the  input  dimensions  to  be changed, instead of the number of hidden neurones.
\subsubsection*{Certifying Some Distributional Robustness}
\label{subsubsec:certified_distributional}

The authors of~\cite{sinha2018certifying} provide a training procedure that augments parameter update with worst-case perturbations of training data - similar to min-max games presented in Section~\ref{subsec:minmax}.
However, the uncertainty sets are chosen to be distributions at close Wasserstein-distance from the training distribution.

The main idea behind providing certified distributional robustness is to optimise a loss surrogate that allows adversarial perturbations within a certain range.
The adversarial range is regulated by controlling the distribution drift, through the Wasserstein-distance.
The experimental results show this method provides better robustness than most networks trained with adversarial examples (Section~\ref{subsec:adv_training}) and even for networks trained with \ac{pgd} (Section~\ref{subsec:minmax}).
\subsubsection*{Convex Outer Adversarial Polytope}
\label{subsubsec:convex_outer}

Kolter and Wong~\cite{kolter2017provable} present a method for training provably robust deep ReLU classifiers.
The approach provides robustness against any norm-bounded adversarial perturbations on the training set.
While no previously unseen example can by-pass this defence, it might sometimes flag non-adversarial examples as adversarial.
The main idea is to construct an outer bound on the set of all final-layer activations that can be achieved by applying a norm-bounded perturbation to an input.
If one can guarantee that the class prediction of an example does not change within this outer bound, it is equivalent to a proof that the example can not be adversarial.

The algorithm shows that the feasible set of the dual problem can be expressed as a neural network.
Therefore, because finding a feasible dual solution provides a guaranteed lower bound on the solution of the primal, using a single backward pass through the second network can prove the lower bound for the primal network under analysis.
Some extensions that help with scalability are presented in~\cite{wong2018scaling}.

\subsubsection*{Lower bounds on the robustness to adversarial perturbations}
\label{subsubsec:lower_bounds}

Peck \etal~\cite{peck2017lower} derive lower-bounds on the magnitude of perturbations necessary to change the classification for various \ac{DNN} architectures.
The lower-bounds are expressed directly in terms of the neural network's parameters.
Although not used as a defence, this methodology allows comparisons between different classifiers, therefore, helping to evaluate defences.

\subsubsection*{Lipschitz-Margin Training}
\label{subsubsec:lipschitz_margin_training}
In~\cite{tsuzuku2018lipschitz}, the authors propose a method to enlarge the provable robust norm-ball during training.
At first, the robust norm ball is defined in terms of Lipschitz constant and the gap between the correct label and other labels, for each input.
This concept, called \emph{margin}~\cite{bartlett2017spectrally} is converted into a loss function in order to ensure non-trivial norm-balls during training. 
The main idea behind this process is to maximises the number of training data points that have guarded areas larger than a hyper-parameter, as long as the original training procedure maximises the number of inputs that are correctly classified.

The authors also introduce a fast method to calculate the upper bounds of the Lipschitz constant, making Lipschitz-margin training computational feasible.
Experimental results show an increase in the provable norm-ball for large networks and improved robustness against adaptive attacks.

\subsubsection{Defences based on \ac{gan}}
\label{subsec:defences_gans}

In this section we present defences based on generative models, first introduced in Section~\ref{subsec:gan_attacks}.

\subsubsection*{Defence-GAN}
\label{subsubsec:defence_gan}

Defence-GAN~\cite{samangouei2018defense} is trained to model the distribution of unperturbed images.
At inference time, it finds a close output to a given image, which does not contain the adversarial change.
This output is then fed to the classifier.
Since it does not alter the classifier's structure or training procedure, it can be used with any classifier architecture.
The experimental results, however, suggests Defence-GAN trades a lot of accuracy for robustness.
\subsubsection*{\ac{fbgan}}
\label{subsubsec:bidirectional_gan}

\ac{fbgan}~\cite{bao2018featurized} first captures the semantic information of any input, original or adversarial, and then retrieves the unperturbed input from this information.
This process is meant to remove the perturbation.
The authors use a bi-directional \ac{gan}~\cite{donahue2016adversarial, dumoulin2016adversarially} to learn the feature mapping and add mutual information to the latent space in order to reduce the dimension of latent codes and ensure they capture the semantical variation.

The defence shows good accuracy, however, it is only tested on small datasets.
It is unclear if the method will scale to large datasets and models with high capacity.

	\subsection{On the Overall Efficacy of Adversarial Defences}
\label{subsec:overall_defences}

One limitation of most publications presented in the previous section is the lack of rigorous evaluation.
With the exception of defences that certify some robustness to adversarial examples, most publications test their defences on simple datasets (such as MNIST~\cite{lecun1998mnist}), on models with small capacity or against single-shot attacks.
Therefore, the efficacy of these defences is often over-emphasised.
\if 0\mode  We illustrate this behaviour in Appendix~\ref{sec:benchmarks}. \else\fi
In this section we review some publications that evaluated some defences and showed they are, in fact, not efficient at all. 
Moreover, we comment on future evaluation technique.

Carlini and Wagner~\cite{carlini2017adversarial} evaluated the efficacy of adversarial detection methods and proved that most detectors perform poorly when faced with powerful, iterative, attackers.
In particular, the authors use their attack~\cite{carlini2017towards} (presented in Section~\ref{subsec:optimisation_attacks} to show that none of the detectors~-~\cite{bhagoji2018enhancing, feinman2017detecting, gong2017adversarial, grosse2017statistical, hendrycks2016early, metzen2017detecting, li2017adversarial}~-~efficiently detect adversarial examples and that the reported results do not reflect the reality.
In another paper, Carlini and Wagner~\cite{carlini2017magnet} show that MagNet~\cite{meng2017magnet} is also easily defeated by adaptive attacks.

A recent result~\cite{athalye2018obfuscated} shows that 7 out of 9 defences published at ICLR 2018 as non-certified defences suffer from gradient obfuscation\if and are not efficient against the0\mode \ac{bpda} attack (Section~\ref{subsubsec:backward_pass})\else.\fi.
He \etal~\cite{he2017adversarial} show that combining adversarial defences in an ensemble defence is not effective, while Sharma and Chen~\cite{sharma2017breaking} broke the complete first order adversary training method~\cite{madry2017towards} \if 0\mode (Section~\ref{subsubsec:madry_learning})~-~\else\fi considered state-of-the-art.
Tramer \etal~\cite{tramer2017ensemble} note  that the MNIST dataset is a simple baseline for assessing the potential of a defence, but the obtained results do not always generalise to harder tasks. 

\if 0\mode \bigskip \else\fi
\paragraph{A protocol for defence evaluation.} Without a common evaluation strategy, it is impossible to compare the strength of multiple defences.
Moreover, without any insight about the effect of a defence on the protected model, combining two or more defences becomes a game of trial and error.
A good protocol to evaluate defences against adversarial examples would be to test against state-of-the-art models~-~\eg~Residual Networks~\cite{he2015delving}~-~trained on large datasets~-~\eg~ImageNet~\cite{ILSVRC15}~-~and under clear security or safety threat models.

\else

	\subsection{Reactive Defences}
\label{subsec:reactive}

Reactive defences add a pre-processing step to the classification pipeline; in which adversarial examples are either detected or their impact is diminished through input transformations.
Because they do not alter the target model, reactive defences do not trade accuracy for robustness.
Moreover, input transformation are fast and can be easily deployed. 
Unfortunately, neither adversarial detection or input transformations are powerful enough to protect \ac{DNN} against adversarial examples (as discussed in Section~\ref{subsec:overall_defences}).

\paragraph{Detection of Adversarial Examples.} Various adversarial detection mechanisms have been developed. 
Gross \etal~\cite{grosse2017statistical} use statistical testing to check if adversarial examples are outside the training data distribution. 
Gong, Wang and Ku~\cite{gong2017adversarial} train a separate \ac{DNN} only with adversarial examples.
Similarly, Metzen \etal~\cite{metzen2017detecting} train a separate \ac{DNN} to detect adversarial examples.
However, instead of using the perturbed input for training, the authors use the output of the hidden layers.
Li and Li~\cite{li2017adversarial} develop an adversarial detector based on features extracted at every layer of a convolutional neural network.
They treat an image as a distribution of pixels that can be used to collect statistics and later used the statistics to train an adversarial detector.
Xu and Qi~\cite{xu2017feature} train two independent detectors using inputs processed with bit reduction or blur.

Feinman \etal~\cite{feinman2017detecting} train a linear adversarial detector using two features: (1) the kernel density estimates in the subspace of the last layer and (2) the bayesian uncertainty estimates extracted from the drop-out layers.
SafetyNet~\cite{lu2017safetynet} enforces an attacker to solve a discrete optimisation problem.
Each activation of a ReLU layer is quantised in order to generate a discrete code, later used to train an RBF-SVM adversarial detector.
Zhang \etal~\cite{zhang2018detecting} train a binary classifier with benign and saliency data.
The saliency data is used to identify significant pixels, later used to predict adversary examples.

Bhagoji \etal~\cite{bhagoji2018enhancing} use PCA~\cite{shlens2014tutorial} transformations to train adversarial detectors.
The intuition behind using PCA is its ability to identify the directions in which the input data has maximum variance.
By projecting the data data along these axes one can reduce the input dimensions, while preserving variance.
Zhao \etal~\cite{zhao2018detecting} propose a new detection mechanism that aims to hide the input label.
This will prevent an attacker from maximising an input given a label and, thus, from creating an adversarial example.
The authors define a ono-to-one encoding scheme from true labels to code vectors.
In order to detect adversarial examples, one can verify if the code vector computed from an input matches the signature of a class with certain precision.
If the output is negative, the input is treated as an adversarial example.

Ababsi and Gagne~\cite{abbasi2017robustness} developed an ensemble of detectors based on the confusion matrix of a classifier.
The underlying idea is that adversarial instances originating from a given class tend to fall into a small subset of incorrect classes.
Therefore, developing an ensemble of detectors, which can distinguish between confusion classes, can more easily spot adversarial examples.
Lee \etal~\cite{lee2017generative} propose a generative training method in which two \ac{DNN} are trained alternatively.
The first network generates adversarial examples, while the second tries to correctly classify benign and adversarial examples.
Song \etal~\cite{song2017pixeldefend} propose to not only detect adversarial examples, but try to \emph{purify} them and search for the true labels.
They leverage a generative adversarial network, called PixelCNN~\cite{van2016conditional, salimans2017pixelcnn}, to learn the probability distribution of the training set and use statistical tests to detect if a new input belongs to this probability distribution.
If the input does not belong, it is probably an adversarial example.

MagNet~\cite{meng2017magnet} trains a model which distinguishes between a test and training input using the distance between an input and the data manifold. 
A thresholding function later decides if an input is normal or adversarial.
The approach is model independent and can be applied to any neural network architecture.
Ghosh, Losalka and Black~\cite{ghosh2018resisting} designed a generative model that finds a latent random variable such that the input and its label become conditionally independent given the latent variable.
The latent space is chosen as a mixture of Gaussians, such that each mixture component represents one of the classes in the data.
Inferring the label given the latent encoding is done by computing the contribution of the mixture components.
Adversarial samples are rejected based on thresholding the encoder and decoder outputs.
For example, if the distance between the sample encoding and the encoding of the predicted class in the latent space is bellow a threshold.

\paragraph{Input Transformation.} Input transformations are believed to restrict the space of adversarial examples, therefore diminishing their impact.
Guo \etal~\cite{guo2017countering} suggest the use of "bit-depth reduction, JPEG compression, total variance minimisation and image quilting"~\cite{guo2017countering} as a pre-processing step of a convolutional classifier.
The idea of using JPEG compression was also explored in~\cite{dziugaite2016study, das2017keeping, shaham2018defending}. 
Variance minimisation and image quilting prove, in practice, more effective.
Shaham \etal~\cite{shaham2018defending} experiment with different input transformations: low-pass filters, PCA, JPEG compression, low resolution wavelet approximations and soft-thresholding.
Xie \etal~\cite{xie2017mitigating} propose to use two randomisation operations: (1) random resizing of input images and (2) random padding with zeros around the input images.

Thermometer one hot encoding~\cite{buckman2018thermometer} breaks the linear extrapolation behaviour of \ac{DNN} by pre-processing the input with an extremely non-linear function.
However, instead of replacing a real number with its counterpart transformation, the authors replace each real number with a binary vector.
Multiplying the input vector with the network's weight enables different input values to use different parameters of the network.
Inspired by thermometer encoding~\cite{buckman2018thermometer}, Rakin \etal~\cite{rakin2018blind} propose to process the input data using an ensemble of methods which includes $tanh$ function, batch normalisation, thermometer encoding and one hot encoding.

Chen \etal~\cite{chen2018improving} propose a pre-processing technique that can successfully mask the gradients even for iterative attackers.
Their proposal is based on encoding the whole input space using a small set of separable codewords and training a classifier on the encoded information.
Lian \etal~\cite{liang2017detecting} treat perturbations as noise and use noise reduction methods in order to mitigate their threat.
	\subsection{Proactive Defences}
\label{subsec:proactive}
This class of defences alter the model or its learning procedure in order to increase \ac{ML} robustness.
As opposed to reactive defences, where some degree of robustness is achieved  earlier in the processing pipeline, proactive defences aim to design better architectures or training procedures.

\paragraph{Adversarial Training.} 
Training with adversarial examples is a form of regularisation~\cite{girosi1995regularization} \ie~a strategy designed to reduce the test error, possibly affecting the training error. 
A common way to make \ac{DNN} better generalise on a task is to train them with more data. 
In practice, however, the amount of available data is limited. 
In order to overcome this barrier, one can create fake data (by applying noise, translations, rotations, \etc) and \emph{augment} the training set with the new examples.

Goodfellow \etal~found that training with an adversarial error function based on the \ac{fgsm} attack is an effective regulariser~\cite{goodfellow2014explaining}.
Sinha \etal~\cite{sinha2018gradient} propose a new training framework which assumes that simultaneous gradient updates should be statistically indistinguishable from each other.
Thus the gradient can be regularised in order to remove salient information that can lead to adversarial examples.
Lyu, Huang and Liang~\cite{lyu2015unified} propose a family of gradient based perturbations that can be used as a regularisation technique.
Their work can be seen as a generalisation of the \ac{fgsm} for all possible norms.
Roth \etal~\cite{roth2018adversarially} propose to use a generative model in order to learn the distribution of adversarial corruptions from examples and use it as for regularisation.
The authors define an objective function that includes a mixing distribution between un-corrupted and corrupted data.

Another way of training with adversarial examples is to exclude the clean examples from the training procedure.
This method is equivalent to training on the \emph{worst} inputs possible, generated by adversarial attack methods and is similar to \emph{robust optimisation}~\cite{ben2009robust}.
Robust optimisation problems are formulated in a mix-max fashion where the loss function is being minimised \wrt~the most effective perturbation from a known set $S$: $\min_{\vtheta} \rho(\vtheta), \quad \text{where} \quad \rho(\vtheta) = \mathbb{E}_{(\vx, y) \sim p_{\text{data}}} \left[\max_{\eta \in \mathcal{S}} J(\vtheta, \vx + \eta, y) \right]$.
However, solving the inner maximisation problem increases the training time considerably and might not always be tractable.

Shaham, Yamada and Negahban~\cite{shaham2015understanding} first investigated the min-max training procedure, but found the inner maximisation problem intractable. 
Therefore, they propose to minimise a surrogate of the min-max function in which they only evaluate a single step of the gradient (ascent and descent, corresponding to min and max).
Huang \etal~\cite{huang2015learning} use linear approximation to discover the uncertainty sets. 
Madry \etal~\cite{madry2017towards} use \ac{pgd} for the inner maximisation problem, which suggests the problem is, in fact, tractable.
In order to explore a large part of the loss landscape, \ac{pgd} is restarted from many points in the ball around an input.
Surprisingly, although there are many local maxima spread widely apart within this space, they tend to have very well concentrated loss values.
This suggests that an adversarial example found by this method is representative for \emph{all} adversarial examples generated with first order methods.
Similarly, Mirman, Gehr and Vechev~\cite{mirman2018differentiable} use abstract interpretation in order to approximate the uncertainty sets and train \ac{DNN} in a min-max fashion.

Tramer \etal~\cite{tramer2017ensemble} developed a technique that augments training data with adversarial examples crafted on other static, pre-trained, models.
Training with ensemble methods increases the diversity of perturbations seen during training.
This is equivalent to enhancing the uncertainty sets with examples crafted on other models.
Dhillon \etal~\cite{dhillon2018stochastic} define a min-max zero-sum game between an adversary and \ac{DNN}.
The strategy for this game is to prune a random subset of activations (\eg~those with smaller magnitude) and scale-up the survivors in order to compensate.
This approach is similar to the dropout technique, where the activations with high absolute values have a higher chance of being sampled.

\paragraph{Architectural Defences.}
Some defences propose to change the overall architecture of \ac{DNN} by either imposing layer-wise constraints or by altering the final layer.
One of the first proposals uses distillation~-~a transfer learning method in which smaller \ac{DNN} are trained with knowledge extracted from larger \ac{DNN}~\cite{hinton2015distilling}.
Papernot \etal~\cite{papernot2016distillation} use distillation in order to increase the \ac{DNN} robustness.
However, instead of using multiple \ac{DNN} in the training process, distillation is used for a single \ac{DNN}.
As opposed to the original distillation mechanism, the same network architecture is used both for training and distillation.
Cisse \etal~\cite{cisse2017parseval} propose Parseval Networks~-~a regularisation scheme that constrains the Lipschitz constant layer-wise.
Through this constraint any exponential growth of the Lipschitz constant is avoided.
In this setting a regularisation scheme (such as weight decay) at the last layer controls the overall Lipschitz constant of the network.

Gu and Rigazio~\cite{gu2014towards} propose a new network architecture, called deep contractive networks, which also uses a layer-wise constraint.
This constraint minimises the network output variance \wrt~perturbations in the input s.t. a trained model can achieve robustness for perturbations around training data points.
Nayebi and Ganguli~\cite{nayebi2017biologically} develop a new training scheme inspired by biophysical principles in neural circuits.
Following the idea suggested by Goodfellow \etal~\cite{goodfellow2014explaining} that adversarial examples are due to the linear summing of high dimensional input with small weights, the authors propose to force neural networks to operate in a non-linear, saturated, regime.
In order to enforce this constraint, the authors \emph{saturate} the network \ie~ensure that each element of the Jacobian matrix of the model is sufficiently small s.t. the model becomes insensitive to perturbations.

DeepCloak~\cite{gao2017deepcloak} removes features not used in classification in order to increase robustness to adversarial examples.
To identify unnecessary features, adversarial samples are tested against the clean example.
In order to remove the features, a mask layer is introduced before the logit layer. 
The mask serves as a selector, keeping the necessary features and setting the unnecessary to 0.
Lamb \etal~\cite{lamb2018fortified} identify which hidden states are off the data manifold and map these states back to parts of the data manifold where the network performs better.
This process consists of inserting de-noising auto-encoders at crucial points between layers of the original network in order to clean up the transformed data points which may lie outside the data manifold.
Dumont, Maggio and Montalvo~\cite{dumont2018robustness} investigate the resistance to adversarial attacks of three rotation-equivariant network architectures~\cite{cohen2016group, worrall2017harmonic, zhou2017oriented}.
They discover that rotation equivariant networks are significantly more robust to attacks based on small translations and rotations, and marginally more robust to attacks based on local geometric distortions.

Sun \etal~\cite{sun2017hypernetworks} propose to use data dependent weights for each hidden layers of \ac{DNN}.
The weights are generated suing HyperNetworks~\cite{ha2016hypernetworks} - a training technique in which a \ac{DNN} generates the weights for another.

Sidney and Marcus~\cite{pontes2018bidirectional} propose to use bidirectional learning in order to increase robustness to adversarial examples.
Bidirectional learning trains two models - the first on the inputs and the second on a reversed copy of the inputs.
However, in \cite{pontes2018bidirectional}, a single model is trained to behave both as a discriminative and generative model.
Therefore, the same model can be a classifier and a generator in the same time.
This behaviour is achieved using an un-directed \ac{DNN} that back-propagates the errors in both directions~-~each direction of the network has its own biases and the weights are shared.

Dense associative models~\cite{krotov2017dense} store a set of vectors in memory, corresponding to the learnt patterns.
For example, pixel characteristics (\eg~intensity) can be stored together with the corresponding label.
At query time, the model tries to recover the memory sequences from incomplete data (corresponding to variations of an input, or perturbations).
If strong perturbations are applied to an input, the model will fail to recover the original label and consider the input adversarial.

\paragraph{Hyperparameters.}

Given the linear conjecture, introduced in Section~\ref{sec:causes}, which states that adversarial examples exploit the linear behaviour of \ac{DNN}, a normal attempt to increase robustness is to use more non-linear activation functions. 

Goodfellow \etal~\cite{goodfellow2014explaining} suggest that RBF networks - which behave in a non-linear fashion - are naturally immune to adversarial examples and have low confidence when they are fooled.
However, RBF units can not generalise very well and do not achieve the same performance as ReLU \ac{DNN}.
Krotov and Hopfield~\cite{krotov2016dense} tried to replace ReLU activations with higher rectified polynomials and observed an increase in robustness.
However, the increase is not sufficient to completely remove the adversarial phenomenon.

Several publications suggested that \ac{DNN} capacity increases their robustness to adversarial examples\cite{madry2017towards, kurakin2016aadversarial, rozsa2016accuracy}.
However, no publication investigated the correlation in isolated experiments.
Increasing the capacity comes with additional costs in terms of processing power and training time.
It is interesting to follow if capacity and robustness are, indeed, correlated and which are the trade-offs between the two.

\paragraph{Certified Defences.} The following defences \emph{guarantee} a lower or an upper bound for the robustness of \ac{DNN}.
Raghunathan, Steinhardt and Liang~\cite{raghunathan2018certified} focus on computing and upper bound on the worst-case loss of linear classifiers and \ac{DNN} with two hidden layer.
This upper bound serves as a \emph{certificate of robustness} against all attacks for a given network and input.
They use integration to obtain an exact expression for upper bound loss using all gradients in a p-ball around an example and use a semidefinite relaxation in order to convert this to a convex, tractable, optimisation problem.
The loss is later used in training a model.

Katz \etal~\cite{katz2017reluplex} propose the use of SMT solvers to prove upper bound robustness of \ac{DNN}.
This procedure encodes the neural network and the constraints regarding upper bound robustness (max $\epsilon$) as a set of linear equations.
The authors extend the simplex algorithm to handle ReLU activation functions - therefore the name Reluplex.
Reluplex uses SAT solving techniques to check if an adversarial example exists in the given constraint space.
If the constraint is not satisfied, the algorithm returns a counter-example which constitutes a valid adversarial example.
Some recent work extends Reluplex to \cite{carlini2018provably} to verify $L_1$ and $L_\infty$ norm, by encoding absolute values using ReLUs.

On the same path, Huang \etal~\cite{huang2017safety} use model checkers in order to guarantee robustness against \emph{known} adversarial perturbations and 
Ehlers~\cite{ehlers2017formal} uses SMT solvers for neural networks with linear activation 
An inherent disadvantage of using SMT solvers, however, is processing time. 

Sinha, Namkoong and Duchi~\cite{sinha2018certifying} provide a training procedure that augments parameter updates with worst-case perturbations of training data~-~similar to min-max training procedure.
However, the uncertainty sets are chosen to be distributions at close Wasserstein-distance from the training distribution.
The main idea behind providing certified distributional robustness is to optimise a loss surrogate that allows adversarial perturbations within a certain range.
The adversarial range is regulated by controlling the distribution drift, through the Wasserstein-distance.

Kolter and Wong~\cite{kolter2017provable} present a method for training provably robust deep ReLU classifiers.
The approach provides robustness against any norm-bounded adversarial perturbations on the training set.
The main idea is to construct an outer bound on the set of all final-layer activations that can be achieved by applying a norm-bounded perturbation to an input.
If one can guarantee that the class prediction of an example does not change within this outer bound, it is equivalent to a proof that the example can not be adversarial.

Tsuzuku and Sato~\cite{tsuzuku2018lipschitz} propose a method to enlarge the provable robust norm-ball during training.
At first, the robust norm ball is defined in terms of Lipschitz constant and the gap between the correct label and other labels, for each input.
This concept, called \emph{margin}~\cite{bartlett2017spectrally} is converted into a loss function in order to ensure non-trivial norm-balls during training. 
The main idea behind this process is to maximises the number of training data points that have guarded areas larger than a hyper-parameter, as long as the original training procedure maximises the number of inputs that are correctly classified.

\paragraph{Defences based on Generative Models.} Defence-GAN~\cite{samangouei2018defense} is trained to model the distribution of unperturbed images.
At inference time, it finds a close output to a given image, which does not contain the adversarial change.
This output is then fed to the classifier.
Since it does not alter the classifier's structure or training procedure, it can be used with any classifier architecture.

Bao Liang and Wang~\cite{bao2018featurized} first capture the semantic information of any input, original or adversarial, and then retrieve the unperturbed input from this information.
This process is meant to remove the perturbation.
The authors use a bi-directional \ac{gan}~\cite{donahue2016adversarial, dumoulin2016adversarially} to learn the feature mapping and add mutual information to the latent space in order to reduce the dimension of latent codes and ensure they capture the semantical variation.

\fi
    \section{Transferability}
\label{sec:transferability}

Besides the existence of adversarial examples, the first by Szegedy \etal~\cite{szegedy2013intriguing} showed that adversarial examples can also transfer between different \ac{DNN}.
However, this phenomenon was only later explored in depth by Papernot, McDaniel and Goodfellow~\cite{papernot2016transferability}.
They studied the ability to transfer adversarial examples not only between \ac{DNN}, but also across different \ac{ML} techniques.
The authors propose an initial taxonomy for transferability, where adversarial examples that transfer between "models trained with \emph{the same} machine learning technique, but different parameter initialisation or datasets"~\cite{papernot2016transferability} are called (1) \emph{intra-technique} and "adversarial examples that transfer between models trained using \emph{different} machine learning techniques"~\cite{papernot2016transferability} are called (2) \emph{cross-technique} examples.
It is worth to say that \ac{DNN} with different architectures are considered to belong to the first category, as the number of hidden layers or neurones in a layer can be considered a tunable hyper-parameter.

Another important property used to describe \ac{ML} techniques and of importance when judging the transferability property of adversarial examples is the trait of such algorithms to be differentiable or not.
\ac{DNN} and \ac{lm} learn by attempting to minimise a loss function and are, by nature, differentiable.
On contrast, \ac{SVM}, \ac{dt} or \ac{knn} are non-differentiable machine learning techniques.
In particular, \ac{knn} algorithms do not require a labeled dataset during training and learn in an \emph{un-supervised} fashion.

The authors of~\cite{papernot2016transferability} found that all \ac{ML} models are vulnerable to adversarial examples crafted using the same training technique (intra-technique).
The phenomenon shows stronger for differentiable models than for non-differentiable models.
In regard to cross-technique transferable adversarial examples, \ac{lm}, \ac{SVM}, \ac{dt} or an ensemble of models that collectively make predictions are more vulnerable than \ac{DNN} or \ac{knn}, which present some resilience.
The initial experiments presented in~\cite{papernot2016transferability} are performed on the MNIST~\cite{lecun1998mnist} dataset, which is of small dimensions and less challenging.

Liu \etal~\cite{liu2016delving} investigated the transferability phenomenon at larger scale using the ImageNet~\cite{krizhevsky2012imagenet} dataset and various deep architectures.
Moreover, the authors discuss transferability in close relation to the most import attacker goals (Section~\ref{sec:attack_models}): (1) to cause a random misclassification or (2) a targeted misclassification.
The experimental results show non-targeted attacks transfer easily between models.
In contrast, targeted attacks do not maintain the target labels when transferred.
This result casts doubt on the use of adversarial examples in many security contexts.
Moreover, when only the hyper-parameters vary and the same architecture is used, transferability is not consistent, revealing that it is heavily dependent on hyper-parameters.
The authors believe cross-technique targeted transferability performs poorly because the attacks are based on gradient methods (Section~\ref{subsec:sensitivity}), which search for adversarial examples in a small subspace which do not contain the target label for several methods.

Tramer \etal~\cite{tramer2017space} show empirical evidence that different models draw similar decision boundaries and propose a method to measure the space where adversarial examples can be found.
Following the hypothesise that adversarial examples inhabit large, continuous regions instead of small pockets~\cite{gu2014towards} (Section~\ref{sec:causes}), the authors measure the number of orthogonal perturbations that can lead to a misclassification.
Although these vectors are not sufficient to represent a basis of the adversarial space, they are a good indicator of its size.
Experimental results show this number heavily depends on the network's architecture.
However, even small adversarial spaces are sufficient to intersect across different models and give transferable example.

Tramer \etal~\cite{tramer2017space} also measure the distance in directions that lead to adversarial examples and compare them with distances between very similar inputs in the training set, that represent different classes.
As expected, since adversarial examples are very close to normal inputs, the  first distance is smaller.
Nevertheless, the measurement method remains important.

Another important conclusion of this study~\cite{tramer2017space} shows that transferability is inherent to models that "learn feature representations that preserve non-robust properties of the input space"~\cite{tramer2017space}.
If models were to learn (or designed to select) different features, it would have not been possible to transfer adversarial examples.
Therefore, transferability is a consequence of algorithm design.

\if 0\mode \bigskip \else\fi
The literature brings sufficient empirical evidence to prove different \ac{ML} algorithms learn a close representation of the input space.
Thus, evading a region corresponding to a correct class with sufficient distance from the boundaries would most certainly transfer between models that learn in a similar fashion.
However, while the representation of classes is similar, they are not distributed alike in the output space.
Therefore, it is much harder to transfer an input in a desired target region.

Not being able to perform targeted transfers limits the applicability of adversarial examples in security contexts. 
An important conclusion that strengthens this point is that transferability is dependent on hyper-parameters.
Thus, while non-targeted attacks are successful for intra-technique transfers, targeted attacks are heavily burdened by variations in hyper-parameters.
In real-world scenarios is common to use similar the same learning technique, but given the proliferation of \ac{DNN} architectures and training methods, it is less common to find the same hyper-parameters in different black-box systems.
\if 0\mode 
Nevertheless, this is, at the moment, an assumption that deserves future investigations.
\else \fi
\if 0\mode
We argue that transferability should be an important topic (often neglected in current publications) when designing attacks or defences against adversarial examples, but less significant when trying to give robustness guarantees.
This statement also follows from the conclusions in~\cite{tramer2017space}, that transferability is not an inherent property of non-robust models, but a consequence of algorithm design.
It is, therefore, compulsory to evaluate and discuss the importance of transferability, together with the threat model, when designing new attacks or defences.
\else\fi
    \section{Distilled Knowledge}
\label{sec:distilled_knowledge}


We focused on several points related to the adversarial examples phenomenon.
In this section we summarise and comment on the most interesting directions, that can also shape future research.

At first, as discussed in Section~\ref{sec:causes}, an unanimously accepted conjecture on the existence of adversarial examples is still missing.
Following the two main classes of attacks, we distinguish between two different perspectives on the existence of adversarial examples: (1) the case in which adversarial examples are drawn from the same distribution as normal inputs and lie on the same data manifold and (2) the case in which adversarial examples are part of a different distribution and lie off the data manifold.
To this moment there is not enough empirical data to refute any of these conjectures - although the adherents to the second hypothesis do not try to reject the first one.

Carlini and Wagner~\cite{carlini2017adversarial} questioned the second conjecture by defeating all the defences built upon it.
However, this evidence is not sufficient to completely refute it.
Regarding the first conjecture, some interesting insights are presented in~\cite{gilmer2018adversarial}, on artificial tasks where the data manifold can be explored.
We argue that similar inquiries that can reveal information about models with higher capacity are needed.
Moreover, more fundamental research on the causes and effects of perturbations can lead to models with increased robustness.
For example, limiting (or saturating) the gradients~\cite{nayebi2017biologically} as a response to sensitive features is not enough to completely remove the effect of adversarial examples.
This reveals that sensitive features are not the only cause for adversarial examples and highlights the need for further investigations.

Secondly, although a consistent body of publications claims security consequences of adversarial examples, very few present concrete (and practical) threat models or scenarios.
Close to none take into consideration the economics of using adversarial examples as opposed to other attacks.
From this perspective most security claims of adversarial examples remain invalid.
\if 0\mode We argue, in agreement with~\cite{gilmer2018adversarial}, that whenever authors claim security implications, a clear threat model and use cases must be presented.\else\fi
This angle opens new research questions: when (and why) attacks using adversarial examples are better?
And which defences act better in these use-cases?

Thirdly, the lack of standardised evaluation methods makes it difficult to compare  adversarial defences.
New publications should evaluate and present their results on models with high capacity (\eg~\cite{he2016deep}) and trained on large datasets (\eg~ImageNet~\cite{ILSVRC15}).
Moreover, the authors should present the results obtained against iterative (adaptive) attacks~-~considered state of the art.
Large scale evaluations, like the one suggested, will lead to better defences and, ultimately, to more robust models.

Several definitions of robustness are presented in the literature (and outlined in Section~\ref{sec:robustness_eval}).
They focus on the area around an input where no adversarial examples can be found or on the maximum perturbation for which an adversarial examples, specific to an input, can not be found.
Both definitions gravitate around a specific input.
In these circumstances, even if a certified region around some inputs can be guaranteed to be adversarial-free, it is not clear if this is enough to declare a model robust.
Since the training dataset is only an approximation of the data generation distribution, one can argue that a certainty obtained on the training dataset is only an approximation of a certainty for the data generation distribution.
Judging the notion of certified robustness from this angle raises the question if certification around inputs are sufficient to certify classes of inputs.

Lastly, the phenomenon of adversarial examples sheds light on a more important topic, often neglected~-~deep learning models learn differently than human beings do~-~which opens a number of fundamental questions.
It is not clear (from the literature) why the sensitivity to adversarial examples must be removed.
Is this problem relevant when the objective is to design a computer vision system that resembles the human perceptual system and can be used in a complex system; able to achieve some form of intelligence?
Further on, what is the next step if the phenomenon of adversarial examples is intrinsic to \ac{DNN} models which have even the smallest error~-~as suggested in~\cite{gilmer2018adversarial}?
\ac{DNN} show impressive results on the object recognition task and the ability to distinguish patterns similar to the way humans do. 
It is not clear how adversarial examples influence these patterns, however, when the perturbations are small the patterns are still differentiable.
Can \ac{DNN} make better use of such patterns?
Moreover, when is the problem of adversarial examples considered 'solved'? 
When one has to change an image so much that it resembles another object?
All these questions are still to be answered in future literature.


%


    \small
    \bibliographystyle{abbrv}
    \bibliography{clean_bib}

    \normalsize
    \appendix
\section{Benchmarks}
\label{sec:benchmarks}
In this Appendix we give references to the tools, datasets and \ac{DNN} architectures used in the literature about adversarial examples.
Moreover, for each defence we present the dataset and \ac{DNN} architecture used for evaluation.

\subsection{Tools and Libraries}
Open source tools used to generate attacks or defences against adversarial examples.

\begin{table}[h]
	\centering
	\begin{tabular}{|l|l|l|}
		\toprule
		Name & Link & Publication \\ 
		\midrule
		CleverHans & https://cleverhans.readthedocs.io & \cite{papernot2016cleverhans} \\ \hline
		Foolbox & https://foolbox.readthedocs.io & \cite{rauber2017foolbox} \\ \hline
		Adversarial Robustness Toolbox & https://github.com/IBM/adversarial-robustness-toolbox & \cite{nicolae2018adversarial} \\ 
		\bottomrule
	\end{tabular}
	\caption{List of libraries for adversarial examples.}
\end{table}

\subsection{Datasets and \ac{DNN} architectures.}
Common datasets and \ac{DNN} architectures used in the literature of adversarial examples.

\begin{table}[h]
	\centering
	\begin{minipage}{0.4\linewidth}
		\begin{tabular}{|l|l|}
		\toprule
		Dataset & Reference \\ 
		\midrule
		MNIST & \cite{lecun1998mnist} \\ \hline
		F-MNIST & \cite{xiao2017fashion} \\ \hline
		DEBRIN & \cite{arp2014drebin} \\ \hline
		Micro-RNA & \cite{shimomura2016novel} \\ \hline
		CIFAR-10/100 & \cite{krizhevsky2009learning, krizhevsky2014cifar} \\ \hline
		ImageNet & \cite{krizhevsky2012imagenet}	 \\ \hline
		ImageNet-1000 & \cite{deng2009imagenet} \\ \hline		
		SVHN & \cite{netzer2011reading} \\ \hline
		HAR & \cite{anguita2013public} \\ \hline
		COIL-100 & \cite{nayar1996columbia} \\ 
		\bottomrule
	\end{tabular}
	\caption{List of common datasets used \\ in the literature of adversarial examples.}
	\end{minipage}%
	\begin{minipage}{.4\linewidth}
	\begin{tabular}{|l|l|}
		\toprule
		Model & Reference \\ 
		\midrule
		LeNet & \cite{lecun1999object} \\ \hline
		Maxout & \cite{goodfellow2013maxout} \\ \hline
		MxNet & \cite{chen2015mxnet} \\ \hline
		AlexNet & \cite{krizhevsky2012imagenet} \\ \hline
		ResNet & \cite{he2016deep} \\ \hline
		Wide ResNet & \cite{zagoruyko2016wide} \\ \hline
		VGG19 & \cite{simonyan2014very} \\ \hline
		DenseNet & \cite{huang2017densely} \\ \hline
		MobileNet & \cite{howard2017mobilenets} \\ \hline
		InceptionResNet-v2 & \cite{szegedy2017inception} \\ \hline
		Inception-v3 & \cite{szegedy2016rethinking} \\ \hline
		Inception-v4 & \cite{szegedy2017inception} \\ \hline
		DAM & \cite{krotov2017dense} \\ \hline
		Defence-GAN & \cite{samangouei2018defense} \\ 
		\bottomrule
	\end{tabular}
	\caption{List of common \ac{ML} models \\ used for adversarial examples on the \\ object recognition task.}
	\end{minipage}
\end{table}

\subsection{Defences Benchmark}
\begin{table}[h]
	\centering
	\begin{tabular}{|l|l|l|}
		\toprule
		Defence &%
		Datasets &%
		Models \\ 
		\midrule
		Statistical Detection~\cite{grosse2017statistical} & MNIST, DREBIN, MicroRNA  &  \ac{dt}, \ac{SVM}, 2 layers-CNN \\ \hline
		Binary Classification~\cite{gong2017adversarial} & MNIST, CIFAR-10, SVHN  &  AlexNet \\ \hline
		In-Layer Detection~\cite{metzen2017detecting} & CIFAR-10, 10-class ImageNet  &  ResNet  \\ \hline
		Detecting from Artifacts~\cite{feinman2017detecting} & MNIST, CIFAR-10, SVHN & LeNet, 12-layer CNN  \\ \hline
		SafetyNet~\cite{lu2017safetynet} & CIFAR-10, ImageNet-1000 & ResNet, VGG19  \\ \hline
		Saliency Data Detector~\cite{zhang2018detecting} & MNIST, CIFAR-10, ImageNet &  AlexNet, AlexNet, VGG19 \\ \hline
		Linear Transformations Detector~\cite{bhagoji2018enhancing} &  MNIST, HAR & SVM  \\ \hline
		Key-based Networks~\cite{zhao2018detecting} & MNIST & 2/3-layers CNN  \\ \hline
		Ensemble Detectors~\cite{abbasi2017robustness} & MNIST, CIFAR-10 &  3-layers CNN \\ \hline
		Generative  Detector~\cite{lee2017generative} & CIFAR-10, CIFAR-100 & 6-layers CNN  \\ \hline
		Convolutional Statistics Detector~\cite{li2017adversarial} & ImageNet & VGG-16  \\ \hline
		Feature Squeezing~\cite{xu2017feature} & MNIST, CIFAR-10, ImageNet & \specialcell{7-layers CNN, DenseNet \\ MobileNet}  \\ \hline
		PixelDefend~\cite{song2017pixeldefend}  & ImageNet & ResNet, VGG \\ \hline
		MagNet~\cite{meng2017magnet} & MNIST, CIFAR-10 & 4/9-layers CNN \\ \hline
		VAE Detector~\cite{ghosh2018resisting}  & MNIST, SVNH, COIL-100 &  - \\ \hline
		Bit-Depth~\cite{guo2017countering} & ImageNet  & ResNet, DenseNet, Inception-v4  \\ \hline
		Basis Transformations~\cite{shaham2018defending}  & ImageNet & Inception-v3, Inception-v4  \\ \hline
		Randomised Transformations~\cite{xie2017mitigating} & ImageNet & Inception-v3, ResNet \\ \hline
		Thermometer Encoding~\cite{buckman2018thermometer} & MNIST, CIFAR-10, CIFAR-100, SVHN & 30-layers CNN, Wide ResNet  \\ \hline
		Blind Pre-Processing~\cite{rakin2018blind}  & MNIST, CIFAR-10, SVHN & LeNet, ResNet-50, ResNet-18  \\ \hline
		Data Discretisation~\cite{chen2018improving} & MNIST, CIFAR-10, ImageNET & InceptionResnet-V2   \\ \hline
		Adaptive Noise~\cite{liang2017detecting} & MNIST, ImageNet &  -  \\ \hline
		FGSM Training~\cite{goodfellow2014explaining} & MNIST &   Maxout \\ \hline
		Gradient Training~\cite{sinha2018gradient} & CIFAR-10, SVHN & ResNet-18 \\ \hline
		Gradient Regularisation~\cite{lyu2015unified} & MNIST, CIFAR-10 & Maxout \\ \hline	
		Structured Gradient Regularisation~\cite{roth2018adversarially} & MNIST, CIFAR-10 & 9-layers CNN \\ \hline
		Robust Training~\cite{shaham2015understanding} & MNIST, CIFAR-10 &  2-layers CNN, VGG \\ \hline
		Strong Adversary Training~\cite{huang2015learning} & MNIST, CIFAR-10 & MxNet  \\ \hline
		CFOA Training~\cite{madry2017towards} & MNIST, CIFAR-10 & 2/4/6-layers CNN, Wide ResNet  \\ \hline
		Ensemble Training~\cite{tramer2017ensemble} & ImageNet & ResNet, InceptionResNet-v2 \\ \hline
		Stochastic Pruning~\cite{dhillon2018stochastic} & CIFAR-10 & Resnet-20 \\ \hline
		Distillation~\cite{hinton2015distilling} & MNIST, CIFAR-10 & 4-layers CNN \\ \hline
		Parseval Networks~\cite{cisse2017parseval} & MNIST, CIFAR-10, CIFAR-100, SVHN & ResNet, Wide Resnet \\ \hline
		Deep Contractive Networks~\cite{gu2014towards} & MNIST & LeNet, AlexNet \\ \hline
		Biological Networks~\cite{nayebi2017biologically} & MNIST & 3-layers CNN \\ \hline
		DeepCloak~\cite{gao2017deepcloak} & CIFAR-10 & ResNet-164 \\  \hline
		Fortified Networks~\cite{lamb2018fortified} & MNIST & 2-layers CNN \\ \hline
		Rotation-Equivariant Networks~\cite{dumont2018robustness} & CIFAR-10, ImageNet & 	ResNet \\ \hline
		HyperNetworks~\cite{sun2017hypernetworks} & ImageNet & ResNet \\ \hline
		Bidirectional Networks~\cite{pontes2018bidirectional} & MNIST, CIFAR-10 & 3-layers CNN \\ \hline
		DAM~\cite{krotov2017dense} & MNIST & DAM \\ \hline
		Certified Defences~\cite{raghunathan2018certified} & MNIST & 2-layers FC \\ \hline
		Formal Tools~\cite{katz2017reluplex, ehlers2017formal, huang2017safety, ruan2018reachability}  & - & - \\ \hline
		Distributional Robustness~\cite{sinha2018certifying}  & MNIST  & 3-layers CNN \\ \hline
		Convex Outer Polytope~\cite{kolter2017provable}  & MNIST, F-MNIST & 2-layers CNN \\ \hline
		Lischitz Margin~\cite{tsuzuku2018lipschitz}  & SVHN & Wide ResNet  \\ \hline
		Defence Gan~\cite{samangouei2018defense}  & MNIST, F-MNIST  & Defene-GAN \\ \hline
		FB-GAN~\cite{bao2018featurized}  & MNIST, F-MNIST & 8-layers CNN \\ 
	
	\bottomrule
	\end{tabular}	
	\caption{Datasets and \ac{ML} models used to benchmark defences against adversarial examples.}	
	\label{tbl:defences_benchmark}
\end{table}

    \end{document}